\documentclass{llncs}

\pdfoutput=1

\usepackage{graphicx}

\usepackage{tikz}

\usepackage{color}

\usepackage{url}

\usepackage{subfigure} 

\usepackage{amsmath}   

\usepackage{amssymb}

\usepackage{float}

\usepackage{ifpdf}

\expandafter\def\expandafter\UrlBreaks\expandafter{\UrlBreaks
  \do\a\do\b\do\c\do\d\do\e\do\f\do\g\do\h\do\i\do\j%
  \do\k\do\l\do\m\do\n\do\o\do\p\do\q\do\r\do\s\do\t%
  \do\u\do\v\do\w\do\x\do\y\do\z\do\A\do\B\do\C\do\D%
  \do\E\do\F\do\G\do\H\do\I\do\J\do\K\do\L\do\M\do\N%
  \do\O\do\P\do\Q\do\R\do\S\do\T\do\U\do\V\do\W\do\X%
  \do\Y\do\Z}

\newcommand{\Cauchy}{L}

\begin{document}

\frontmatter          

\title{ LDDMM meets GANs: Generative
Adversarial Networks for diffeomorphic
registration
 }

\titlerunning{LDDMM meets GANs}     
%

\author{Ubaldo Ramon \and Monica Hernandez \and Elvira Mayordomo}

\institute{Computer Sciences Department \\ Aragon Institute on Engineering Research \\ University of Zaragoza \\ \{uramon,mhg,elvira\}@unizar.es}

\maketitle

\begin{abstract}

The purpose of this work is to contribute to the state of the art of deep-learning methods for diffeomorphic registration. 
We propose an adversarial learning LDDMM method for pairs of 3D mono-modal images based on Generative Adversarial Networks. 
The method is inspired by the recent literature for deformable image registration with adversarial learning. 
We combine the best performing generative, discriminative, and adversarial ingredients from the state of the art 
within the LDDMM paradigm. 
We have successfully implemented two models with the stationary and the EPDiff-constrained non-stationary parameterizations
of diffeomorphisms. 
Our unsupervised and data-hungry approach has shown a competitive performance with respect to a benchmark supervised and 
rich-data approach.
In addition, our method has shown similar results to model-based methods with a computational time under one second.

\end{abstract}

\keywords{Large Deformation Diffeomorphic Metric Mapping  , Generative Adversarial Networks , geodesic shooting , stationary velocity fields.}

\section{Introduction}


Since the 80s, deformable image registration has  become a fundamental problem in medical image analysis~\cite{Sotiras_13}. 
A vast literature on deformable image registration methods exist, providing solutions to important clinical problems and applications.
Up to the ubiquitous success of methods based on Convolutional Neural Networks (CNNs) in computer vision and medical image analysis, 
the great majority of deformable image registration methods were based on energy minimization models~\cite{Modersitzki_09_book}. 
The problem of computing the deformation that {\it best} warps the source into the target image was solved from the minimization of a 
variational problem 
involving different ingredients such as the deformation parameterization, 
the regularization and image similarity metrics, and the optimization method used in the minimization of the energy. 
This traditional approach is model-based, in contrast with recent deep-learning approaches that are known as data-based.



Diffeomorphic registration constitutes the inception point in Computational Anatomy studies for modeling and understanding population 
trends and longitudinal variations, and for establishing relationships between imaging phenotypes and genotypes in Imaging Genetics~\cite{Hua_08,Liu_19}.
Model-based diffeomorphic image registration is computationally costly.
In fact, the huge computational complexity of large deformation diffeomorphic metric mapping (LDDMM)~\cite{Beg_05} is considered 
the curse of diffeomorphic registration, where very original solutions such as the stationary parameterization~\cite{Ashburner_07,Vercauteren_09,Hernandez_14}, 
the EPDiff constraint on the initial velocity field~\cite{Vialard_11}, or the band-limited parameterization~\cite{Zhang_18} 
have been proposed to alleviate the problem.


Since the advances that made it possible to learn the optical flow using CNNs (FlowNet~\cite{Dosovitskiy_15}),
dozens of deep-learning data-based methods have been proposed to approach the problem of deformable image registration in 
different clinical applications~\cite{Boveiri_20}. The trend is augmenting considerably in the last three years.
From them, some interesting proposals have been performed for diffeomorphic registration~\cite{Rohe_17,Yang_17,Dalca_18,Balakrishnan_19,Krebs_19,Fan_19a,Wang_20}. 
These proposals still use most of the ingredients of traditional model-based methods
such as
the stationary parameterization~\cite{Rohe_17,Dalca_18,Krebs_19,Balakrishnan_19}, 
the non-stationary parameterization, 
the parameterization with EPDiff-constrained velocity fields~\cite{Yang_17,Wang_20}, 
the LDDMM energy regularization, and traditional image similarity metrics.
The network architectures are based on ensembles of fully convolutional (FC) layers~\cite{Rohe_17,Yang_17,Dalca_18,Wang_20} or 
modified U-Net based architectures~\cite{Fan_19a}, 
where the input data varies between image patches~\cite{Yang_17} and the whole image~\cite{Fan_19a}.

All the proposals to diffeomorphic registration can be classified into supervised~\cite{Rohe_17,Yang_17,Wang_20} 
or unsupervised learning methods~\cite{Dalca_18,Krebs_19,Fan_19a}. 
Unsupervised methods are usually preferred over supervised ones since the transformations can be learned directly from
image pairs, avoiding the overhead to compute ground truth transformations for training, which is usually approached going back to model-based methods.
Overall, all data-based methods yield fast inference algorithms for diffeomorphism computation once the difficulties 
with training have been overcome.

Generative Adversarial Networks (GANs) is an interesting unsupervised approach and, to our knowledge, it has not been yet proposed 
in the framework of diffeomorphic registration. 
In fact, GANs for deformable registration can be considered at its infancy with few but interesting proposals like~\cite{Mahapatra_18} (2D) 
and~\cite{Duan_18,Fan_19b} (3D).
A GAN combines the interaction of two different networks during training: a generative network and a discrimination
network. 
The generative network itself can be regarded as an unsupervised method that, once included in the GAN system, 
is trained with the feedback of the discrimination network.
It is expected that the generator converges faster and more precisely since the discriminator urges to produce pairs of images 
indistinguishable from the real distribution of target and warped source image pairs.


The purpose of this work is to contribute to the state of the art of data-based methods for diffeomorphic registration
and propose an adversarial learning LDDMM method for pairs of 3D mono-modal images.
The method is inspired in the recent literature for deformable image registration with adversarial learning~\cite{Duan_18,Fan_19b}.
Indeed, it combines the best performing generative, discriminative, and adversarial ingredients from these works within 
the LDDMM paradigm.
We have successfully implemented two models for the stationary and the EPDiff-constrained non-stationary parameterizations. 
We demonstrate the effectiveness of our models in both 2D simulated and 3D real brain MRI data.


In the following, Section~\ref{sec:LDDMM} reviews the foundations of LDDMM underpinning in this work. 
Section~\ref{sec:GAN} describes the proposed method. 
Section~\ref{sec:Results} describes the datasets used for the training and evaluation of our method 
and shows the quantitative and qualitative evaluation results. 
Finally, Section~\ref{sec:Conclusions} 
derives some interesting conclusions of our work.

\section{Background on LDDMM}
\label{sec:LDDMM}

Let $\Omega \subseteq \mathbb{R}^d$ be the image domain.
Let $Diff(\Omega)$ be the LDDMM Riemannian manifold of diffeomorphisms and $V$ the tangent space at the identity element.
$Diff(\Omega)$ is a Lie group, and $V$ is the corresponding Lie algebra~\cite{Beg_05}.
The Riemannian metric of $Diff(\Omega)$ is defined from the scalar product in $V$, 
$\langle v, w \rangle_V = \langle \Cauchy v, w \rangle_{L^2}$,
where $\Cauchy$ is the invertible self-adjoint differential operator associated with the differential structure 
of $Diff(\Omega)$. 
In traditional LDDMM methods, $\Cauchy = (Id - \alpha \Delta)^s, \alpha >0, s \in \mathbb{R}$~\cite{Beg_05}.
We will denote with $K$ the inverse of operator $\Cauchy$.

Let $I_0$ and $I_1$ be the source and the target images.
LDDMM is formulated from the minimization of the variational problem
\begin{equation}
\label{eq:LDDMM}
E(v) = \frac{1}{2} \int_0^1 \langle \Cauchy v_t, v_t \rangle_{L^2} dt + \frac{1}{\sigma^2} \Vert I_0 \circ (\phi^{v}_{1})^{-1} - I_1\Vert_{L^2}^2.
\end{equation}
\noindent The LDDMM variational problem was originally posed in the space of time-varying smooth flows of velocity fields, 
${v} \in L^2([0,1],V)$.
Given the smooth flow ${v}:[0,1] \rightarrow V$, $v_t:\Omega \rightarrow \mathbb{R}^{d}$, 
the solution at time $t=1$ to the evolution equation
\begin{equation}
\label{eq:TransportEquation}
\partial_t (\phi_t^{v})^{-1} = -v_t \circ (\phi_t^{v})^{-1}  
\end{equation}
\noindent with initial condition $(\phi_0^{v})^{-1} = id$ is a diffeomorphism, $(\phi^{v}_{1})^{-1} \in Diff(\Omega)$.
The transformation $(\phi^{v}_{1})^{-1}$, computed from the minimum of $E({v})$,  
is the diffeomorphism that solves the LDDMM registration problem between $I_0$ and $I_1$.


The most significant limitation of LDDMM is its large computational complexity.
In order to circumvent this problem, the original LDDMM variational problem is parameterized on the space of 
initial velocity fields
\begin{equation}
\label{eq:MCCLDDMM}
E(v_0) = \frac{1}{2} \langle \Cauchy v_0, v_0 \rangle_{L^2} + \frac{1}{\sigma^2} \Vert I_0 \circ (\phi^{v}_{1})^{-1} - I_1\Vert_{L^2}^2.
\end{equation}
\noindent where the time-varying flow of velocity fields $v$ is obtained from the EPDiff equation
\begin{equation}
\label{eq:EPDiff}
 \partial_t v_t + K[ (Dv_t)^T \cdot L v_t + DL v_t \cdot v_t + L v_t \cdot \nabla \cdot v_t] = 0
\end{equation}
\noindent with initial condition $v_0$ (geodesic shooting).
The diffeomorphism $(\phi^{v}_{1})^{-1}$, computed from the minimum of $E({v_0})$ via Equations \ref{eq:EPDiff}
and \ref{eq:TransportEquation}, verifies the momentum conservation constraint (MCC)~\cite{Younes_07}, and, therefore,
it belongs to a geodesic path on $Diff(\Omega)$. 

Simultaneously to the MCC parameterization, a family of methods was proposed to further circumvent
the large computational complexity of original LDDMM~\cite{Ashburner_07,Vercauteren_09,Hernandez_14}. 
In all these methods, the time-varying flow of velocity fields $v$ is restricted to be steady or 
stationary~\cite{Arsigny_06}. 

\section{Generative Adversarial Networks for LDDMM}
\label{sec:GAN}

\subsection{GAN-based unsupervised deep-learning networks for diffeomorphic registration}

Similarly to model-driven approaches for estimating LDDMM diffeomorphic registration, 
data-driven approaches for learning LDDMM diffeomorphic registration 
aim at the inference of a diffeomorphism $(\phi^{v}_{1})^{-1}$ such that the LDDMM energy 
is minimized for a given $(I_0, I_1)$ pair.  
In particular, data-driven approaches compute an approximation of the functional
\begin{equation}
\label{eq:Aproximation}
\mathcal{S}(\arg \min_{v \in V} E(v, I_0, I_1))
\end{equation}
\noindent where $\mathcal{S}$ represents the operations needed to compute $(\phi^{v}_{1})^{-1}$
from $v$, and the energy $E$ is either given by Equations~\ref{eq:LDDMM} or~\ref{eq:MCCLDDMM}.
The functional approximation is obtained via a neural network representation with parameters
learned from a representative sample of image pairs.
Unsupervised approaches assume that the LDDMM parameterization in combination with the minimization 
of the energy $E$ considered as a loss function are enough for the inference of suitable diffeomorphic 
transformations after training.
Therefore, there is no need for ground truth deformations. 

GAN-based approaches depart from unsupervised approaches by the definition of two different networks: 
the generative network (G) and the discrimination network (D).
These networks are trained in an alternating way in an adversarial fashion.
The generative network in this context is the diffeomorphic registration network. 
G is aimed at the approximation of the functional given in Equation~\ref{eq:Aproximation} similarly 
to unsupervised approaches for the inference of $(\phi^{v}_{1})^{-1}$.
The discrimination network D outputs the probability $p \in [0,1]$ that for a pair $(I_0^w, I_1)$ 
the image $I_0^w$ comes from a warped source {\it not} being generated by G.

The discrimination network D learns to distinguish between a warped source image 
$I_0 \circ (\phi^{v}_{1})^{-1}$ 
generated by G and a plausible warped source image. 
The learnable parameters of the network G are trained to minimize traditional LDDMM cost functions 
between the warped source image and the target image while trying to fool the discriminator D.
In contrast to other unsupervised approaches, the loss function in G is determined from the combination 
of the LDDMM and the adversarial costs.

%
%
%
%



\subsection{Adversarial training}

As stated above, the registration architecture is composed of two neural networks, 
a generator G and a discriminator D, 
which are trained alternatively as follows.

The discriminator network D is trained using the loss function
\begin{equation}
L_D = \left\{ \begin{array}{lc}
             -\log(p) & c \in P^+   \\
             -\log(1-p) & c \in P^-   \\
             \end{array}
   \right.
\end{equation}
\noindent where $c$ indicates the input case, $P^+$ and $P^-$ indicate positive or negative 
cases for the GAN, and $p$ is the probability computed by D for the input case.

In the first place, D is trained on a positive case $c \in P^+$ representing a target image $I_1$ and 
a warped source image $I_0^w$ plausibly registered to $I_1$ with a diffeomorphic transformation.
The warped source image is modeled from a strictly convex linear combination of $I_0$ and $I_1$
$I_0^w = \beta I_0 + (1- \beta) I_1$. 
It should be noticed that, although the warped source image would ideally be $I_1$, 
the selection of $I_0^w = I_1$ (e.g. $\beta = 0$) empiricaly leads 
to the discriminator rapidly outperforming the generator.
The parameter $\beta$ is the relative $MSE$ error obtained after registration for $I_0^w$ and $I_1$ 
since
\begin{eqnarray}
MSE_{rel} =  \frac{\Vert I_0 \circ (\phi^{v}_{1})^{-1} - I_1\Vert_{L^2}^2}{\Vert I_0 - I_1 \Vert_{L^2}^2}
& \textnormal{and} &
\Vert I_0^w - I_1 \Vert_{L^2}^2 = \beta \Vert I_0 - I_1 \Vert_{L^2}^2.
\end{eqnarray}
\noindent Therefore, this model for $I_0^w$ can be regarded a good candidate for warped sources
after deformable registration for small $\beta$s. 
It has been successfully used in adversarial learning methods for deformable registration~\cite{Fan_19b}.

In the second place, D is trained on a negative case $c \in P^-$ representing a target image $I_1$ and 
a warped source image $I_0^w$ obtained from the generator network G.

In third place, the generator network G is trained using the combined loss function
\begin{equation}
 L_G = L_\textnormal{adv} + \lambda E(v, I_0, I_1). 
\end{equation}
\noindent In this loss function, $L_\textnormal{adv}$ is the adversarial loss function, 
defined from $L_\textnormal{adv} = -\log(p)$ where $p$ is computed from D;
$E$ is the LDDMM energy given by Equations~\ref{eq:LDDMM} or~\ref{eq:MCCLDDMM}; and $\lambda$ is the
weight for balancing the adversarial and the generative losses.
For each sample pair $(I_0^w,I_1)$, G is fed with the pair of images and updates the network parameters  
from the back-propagation of the information of the loss function values coming from the LDDMM energy 
and the discriminator probability of being a pair generated by G.

In the early stages of learning, it is expected that the generator network provides misaligned images
and the discriminator penalizes the system with high probabilities for the negative cases.
As the learning progresses, the generator is trained to fool the discriminator, so the generated warped 
sources will be diffeomorphically transformed to resemble $I_1$ as much as possible according to the 
convex linear model.
The discrimination will eventually hardly distinguish the generated warped sources from the true population,
yielding low probabilities for the negative cases, and learning will be considered to converge.

\subsection{Proposed GAN architecture}

\subsubsection{Generator network.}

In this work, the diffeomorphic registration network G is intended to learn LDDMM diffeomorphic 
registration parameterized on 
the space of steady velocity fields 
or 
the space of initial velocity fields subject to the EPDiff equation (Equation~\ref{eq:EPDiff}).
The diffeomorphic transformation $(\phi^{v}_{1})^{-1}$ is obtained from these velocity fields 
either from scaling and squaring~\cite{Vercauteren_09,Hernandez_14} 
or 
the solution of the deformation state equation~\cite{Beg_05}. 
Euler integration is used as PDE solver for all the involved differential equations.

A number of different generator network architectures have been proposed in the recent literature, 
with predominance of simple fully convolutional (FC)~\cite{Mahapatra_18} or U-Net like architectures~\cite{Duan_18,Fan_19b}. 
In this work, we propose to use the architecture by Duan et al.~\cite{Duan_18} adapted to fit our purposes. 
The network follows the general U-net design of utilizing a encoder- decoder structure with skip connections. 
However, during the encoding phase, the source and target images are fed to two encoding streams.
The first stream follows a fully convolutional design (FC) while the second stream follows the traditional U-net 
encoding pattern.
For each resolution level, the parameters from the two streams are combined and fed to the next level.
In contrast to simpler U-net architectures, the combination of the two encoding streams allows a larger 
receptive field suitable to learn large deformations.
During the decoding phase, the encoding output is passed through an upsampling decoder to obtain the velocity 
field and the diffeomorphic transformation $(\phi^{v}_{1})^{-1}$.
The upsampling is performed with a deconvolutional operation based on transposed convolutional layers~\cite{Zeiler_11}.
We have empirically noticed that the learnable parameters of these layers help reducing typical checkerboard GAN artifacts in 
the decoding~\cite{Odena_16}.

\subsubsection{Discriminator network.}

The discriminator network D follows a very traditional CNN architecture. 
The two input images are concatenated and passed through five convolutional blocks. 
Each block includes a convolutional layer, 
a RELU activation function, 
and a size-two max-pooling layer. 
After the convolutions, the 4D volume is flattened and passed through three fully connected layers.
The output of the last layer is the probability of the input images to come from a 
registered pair not generated by G.


\subsubsection{Generative-Discriminative integration layer.}

The generator and the discriminator networks G and D are connected trough an integration layer in the 
shape of a spatial transformation layer.
This integration layer allows calculating the diffeomorphism $(\phi^{v}_{1})^{-1}$ that warps the source image $I_0$.
The selected integration layer depends on the velocity parameterization: stationary or EPDiff-constrained time-dependent.
The computed diffeomorphisms are applied to the source image via a second 3D spatial transformation layer~\cite{Jaderberg_15}
with no learnable parameters.
The gradients of the integration layer are back-propagated from D to train G.
In the following, the method with the stationary parameterization will be recalled as SVF-GAN, and the method with the
EPDiff-constrained parameterization will be recalled as EPDiff-GAN.



\subsubsection{Parameter selection and implementation details}

We selected the parameters $\lambda = 1000$, $\sigma^2 = 1.0$, $\alpha = 0.0025$, and $s = 4$ and a unit-domain 
discretization of the image domain $\Omega$~\cite{Beg_05}.
Scaling and squaring and Euler integration were performed in 10 time samples.
The parameter $\beta$ for the convex linear modeling of warped images was selected equal to $0.2$.
This means that the discriminator is trained to learn deformable image registration results with a $20\%$ of
$MSE_{rel}$ level of accuracy.

Both the generator network and the discriminator network were trained with Adam's optimizer with default 
parameters and learning rates of $5e^{-5}$ for G and $1e^{-6}$ for D, respectively. 

The experiments were run on a machine equipped with one NVidia Titan RTX with 24 GBS of video memory and an Intel 
Core i7 with 64 GBS of DDR3 RAM. 
The codes were developed in the GPU with Keras and a TensorFlow backend.

\section{Results}
\label{sec:Results}

In this section we demonstrate the effectiveness of our non-supervised proposed methods by training and testing
on a 2D simulated dataset and 3D brain MRI datasets.

\subsection{Datasets}

\subsubsection{2D simulated dataset.} 

We simulated a total of 2560 torus images by varying the parameters of two ellipse equations, similarly to~\cite{Wang_20}.
The parameters were drawn from two Gaussian distributions: $\mathcal{N}(4,2)$ for the inner ellipse and $\mathcal{N}(12,4)$
for the outer ellipse. The simulated images were of size 64 $\times$ 64.
Our GANs were trained during 1000 epochs with a batch size of 64 samples.

\subsubsection{3D brain MRI datasets.} 

For adversarial training, we used a total of 2113 T1-weighted brain MRI images from the Alzheimer's Disease Neuroimaging Initiative 
(ADNI, adni.loni.usc.edu).
The ADNI was launched in 2003 as a public-private
partnership, led by Principal Investigator Michael W. Weiner, MD. The primary goal of ADNI has been to
test whether serial magnetic resonance imaging (MRI), positron emission tomography (PET), other
biological markers, and clinical and neuropsychological assessment can be combined to measure the
progression of mild cognitive impairment (MCI) and early Alzheimer’s disease (AD).
The images were acquired at the baseline visit and belong to all the available ADNI projects (1, 2, Go, and 3).
The images were preprocessed with N3 bias field correction, 
affinely registered to the MNI152 atlas, 
skull-stripped,
and affinely registered to the skull-stripped MNI152 atlas.

The evaluation of our generated GAN models in the task of diffeomorphic registration was performed in NIREP 
dataset~\cite{Christensen_06}.
This dataset was released for the evaluation of non-rigid registration. 
The geometry of the segmentations in NIREP provides a specially challenging framework for deformable 
registration evaluation.
The images were acquired from 8 males and 8 females with a mean age of 32.5 $\pm$ 8.4 and 29.8 $\pm$ 5.8 years, 
respectively.
The substantial age differences between train and test subjects are intended to demonstrate the generalization
capability of our non-supervised models.

Both the encoder and decoder networks in G were implemented to work with images of size 
$176 \times 224 \times 176$. 
Our GANs were trained during 50 epochs with a batch size of 1 sample.
This selection of image size and batch sampling was performed due to VRAM memory issues.

\subsection{Results in the 2D simulated dataset}

\begin{figure}[!t]
\centering

\scriptsize sources \\
\begin{tabular}{ccccc}
\includegraphics[width=0.065\textwidth]{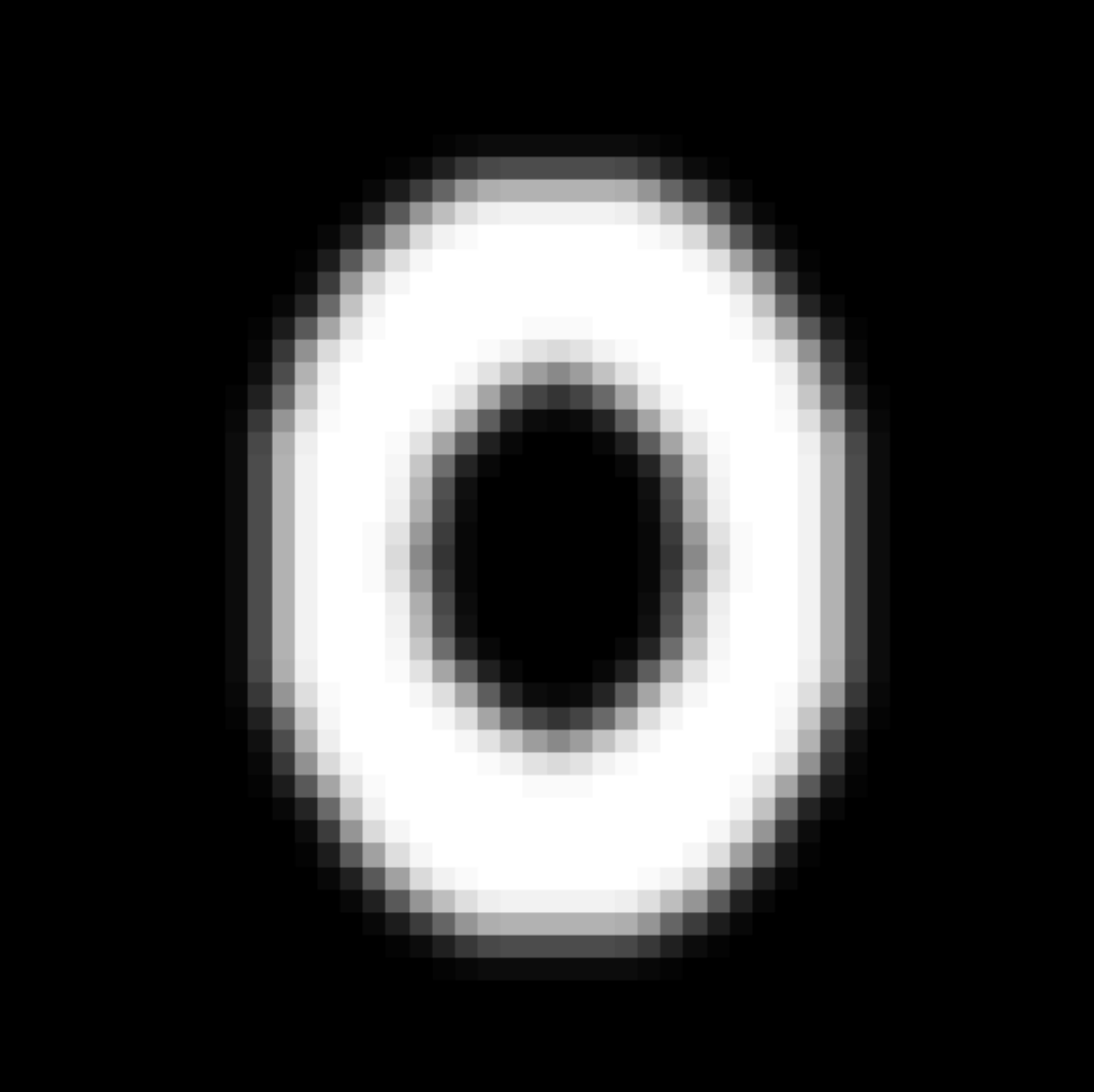} & 
\includegraphics[width=0.065\textwidth]{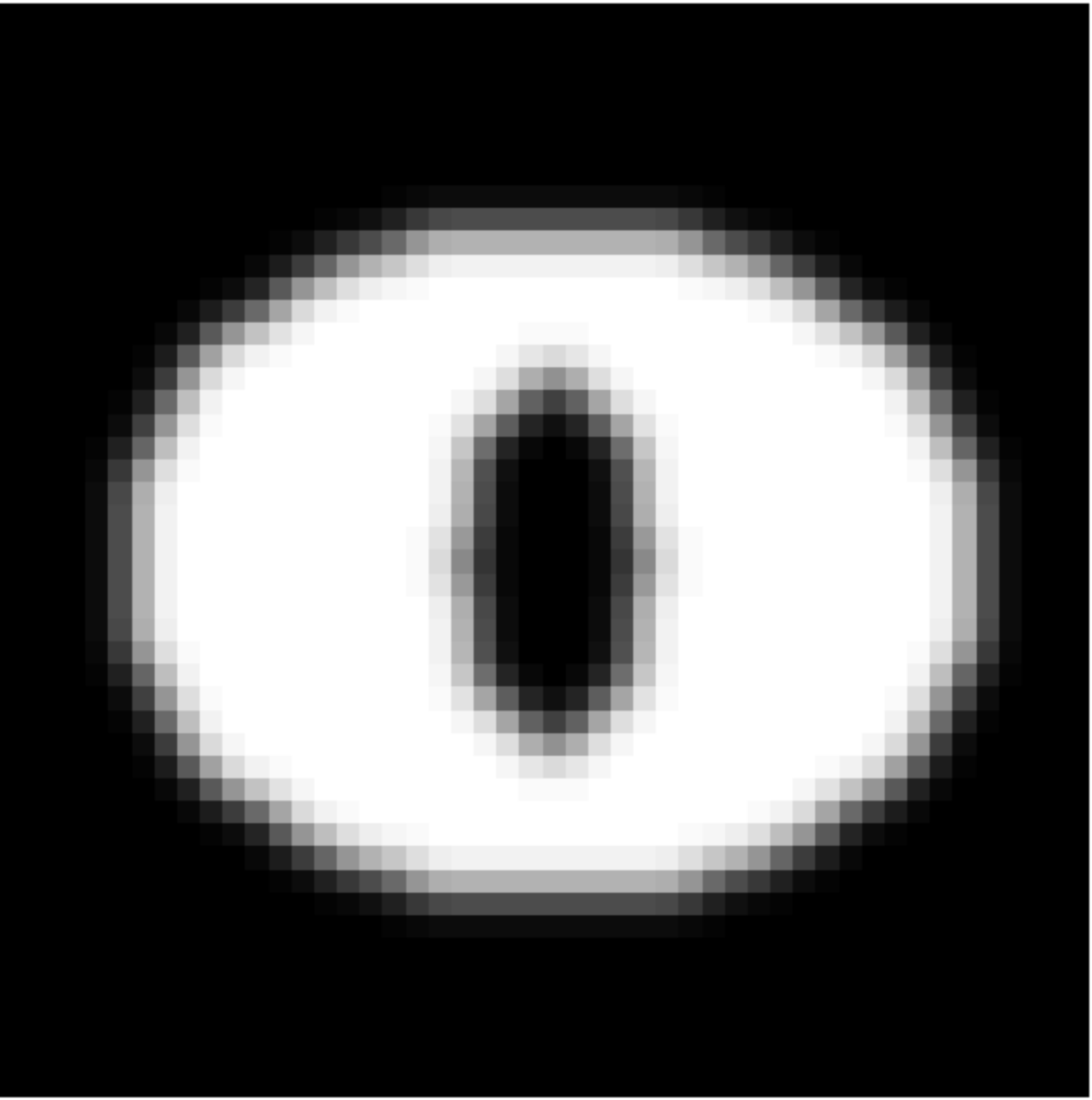} & 
\includegraphics[width=0.065\textwidth]{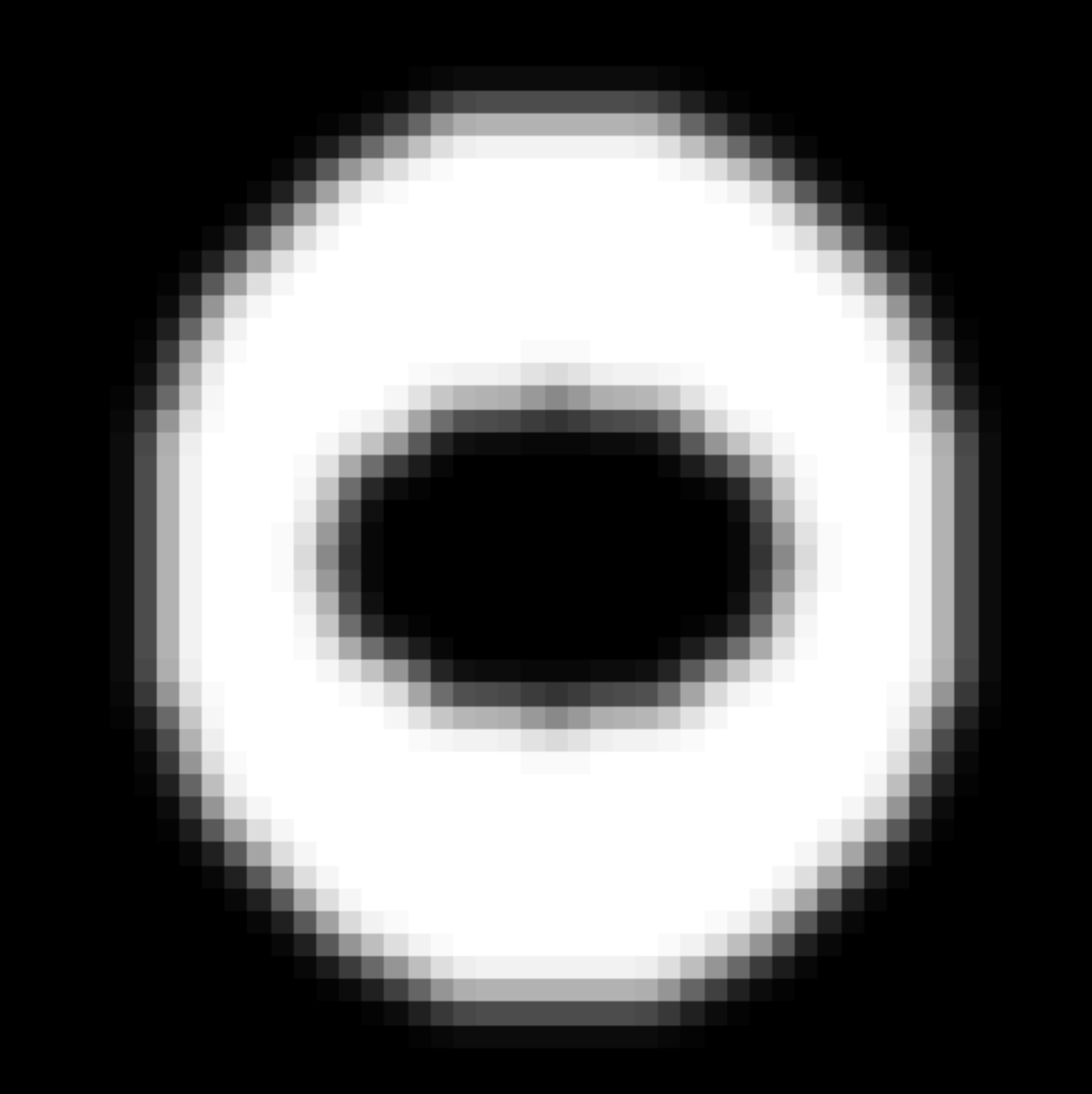} & 
\includegraphics[width=0.065\textwidth]{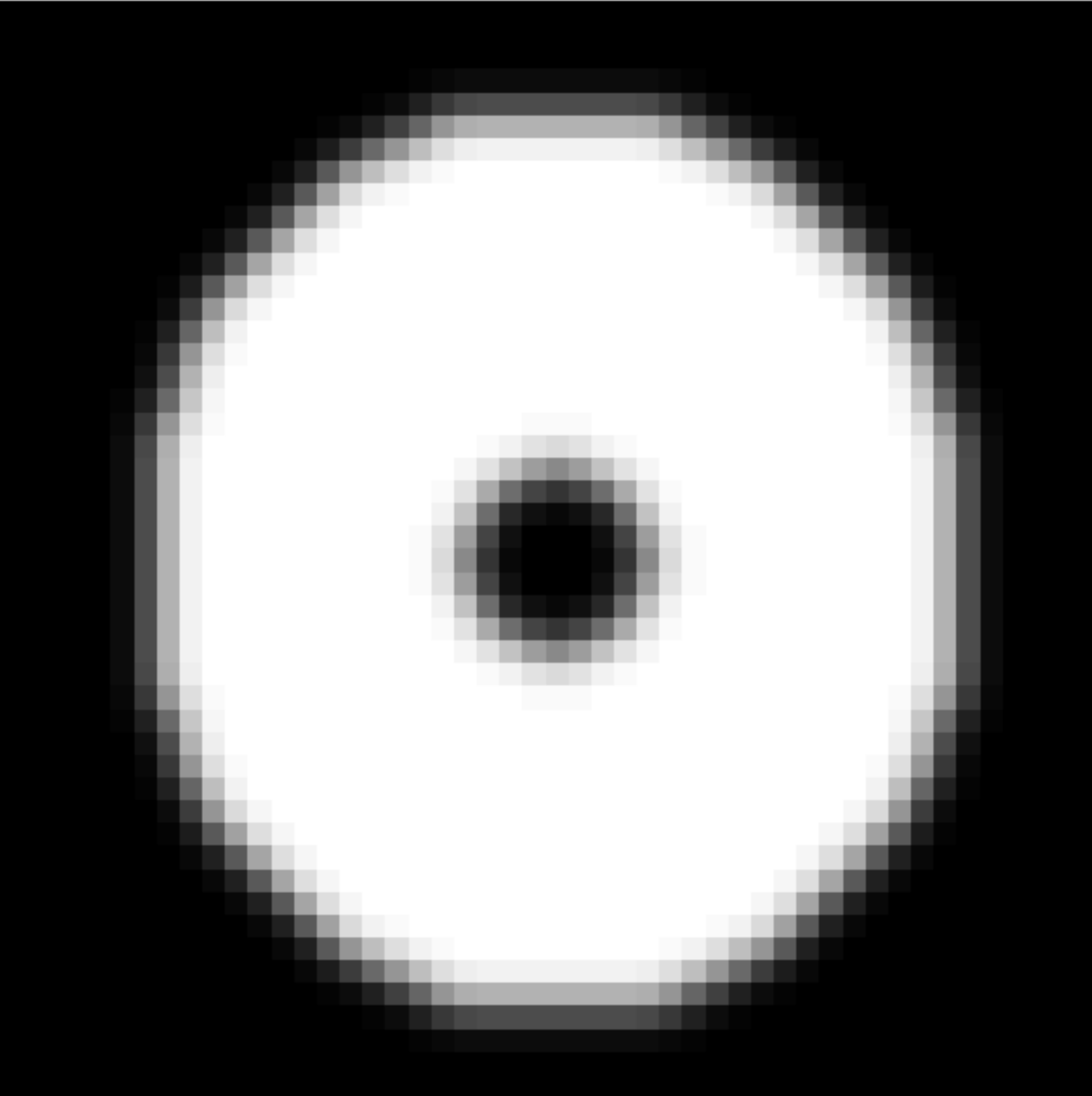} & 
\includegraphics[width=0.065\textwidth]{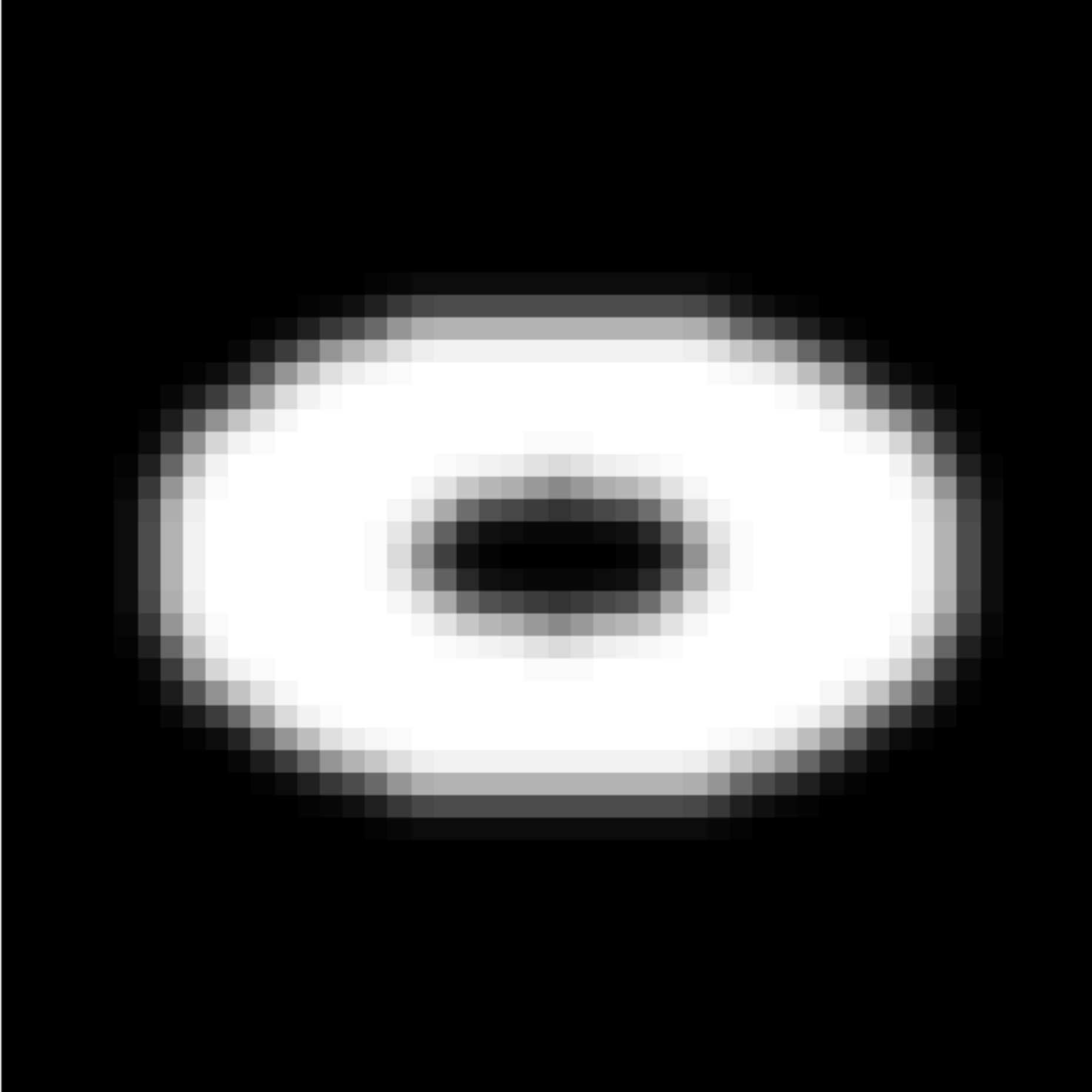} \\
\end{tabular}
\\
\scriptsize targets \\
\begin{tabular}{ccccc}
\includegraphics[width=0.065\textwidth]{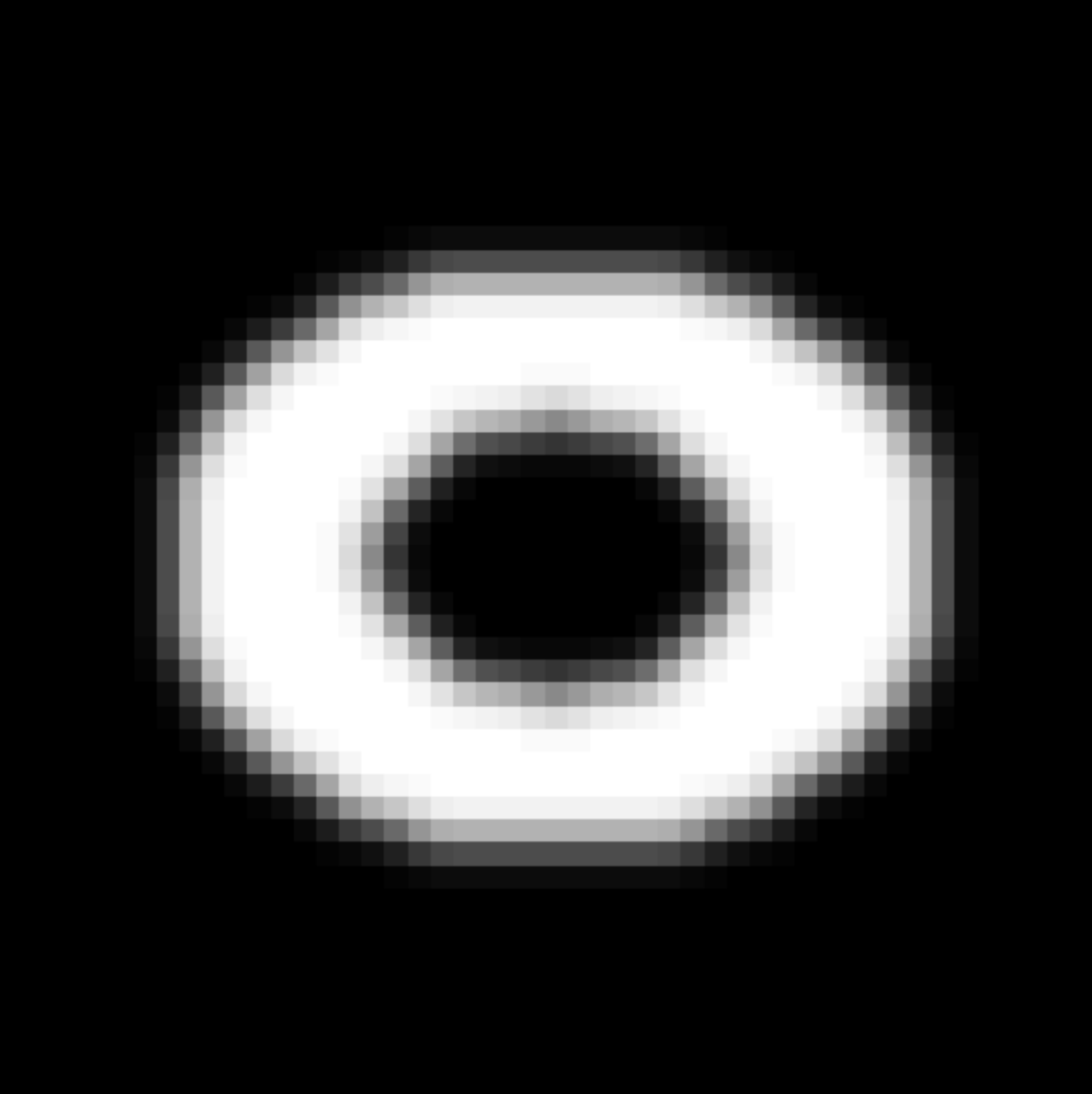} & 
\includegraphics[width=0.065\textwidth]{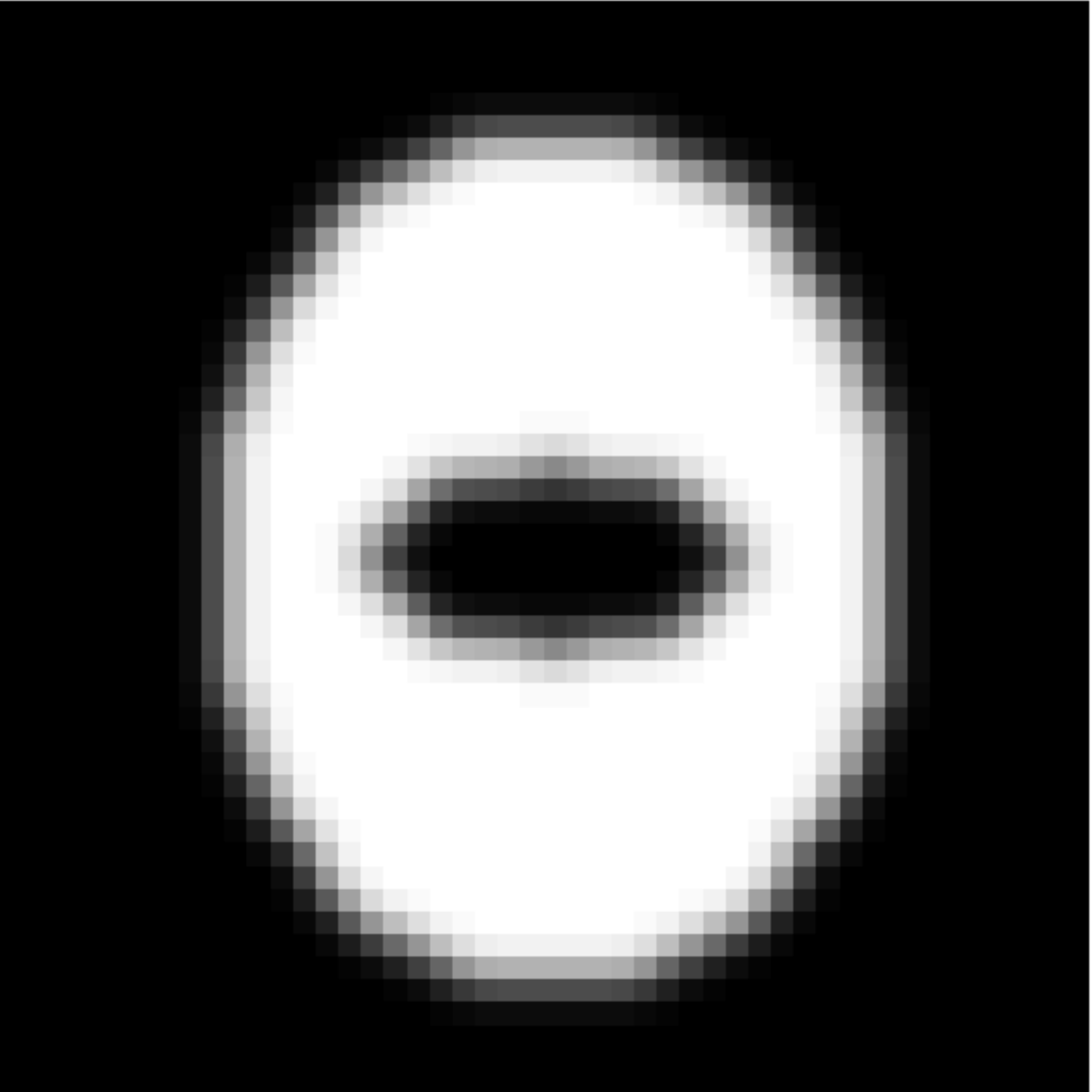} & 
\includegraphics[width=0.065\textwidth]{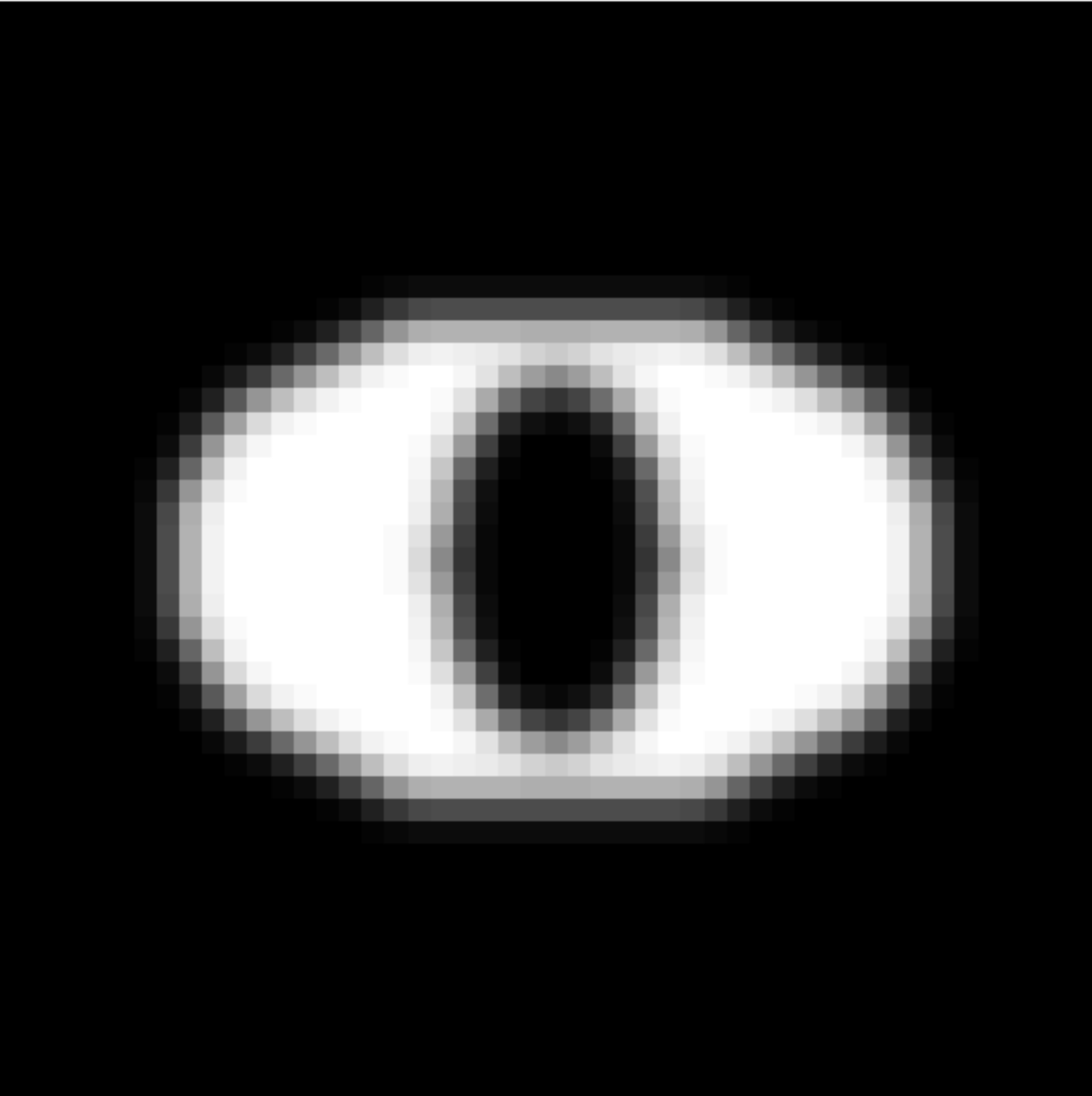} & 
\includegraphics[width=0.065\textwidth]{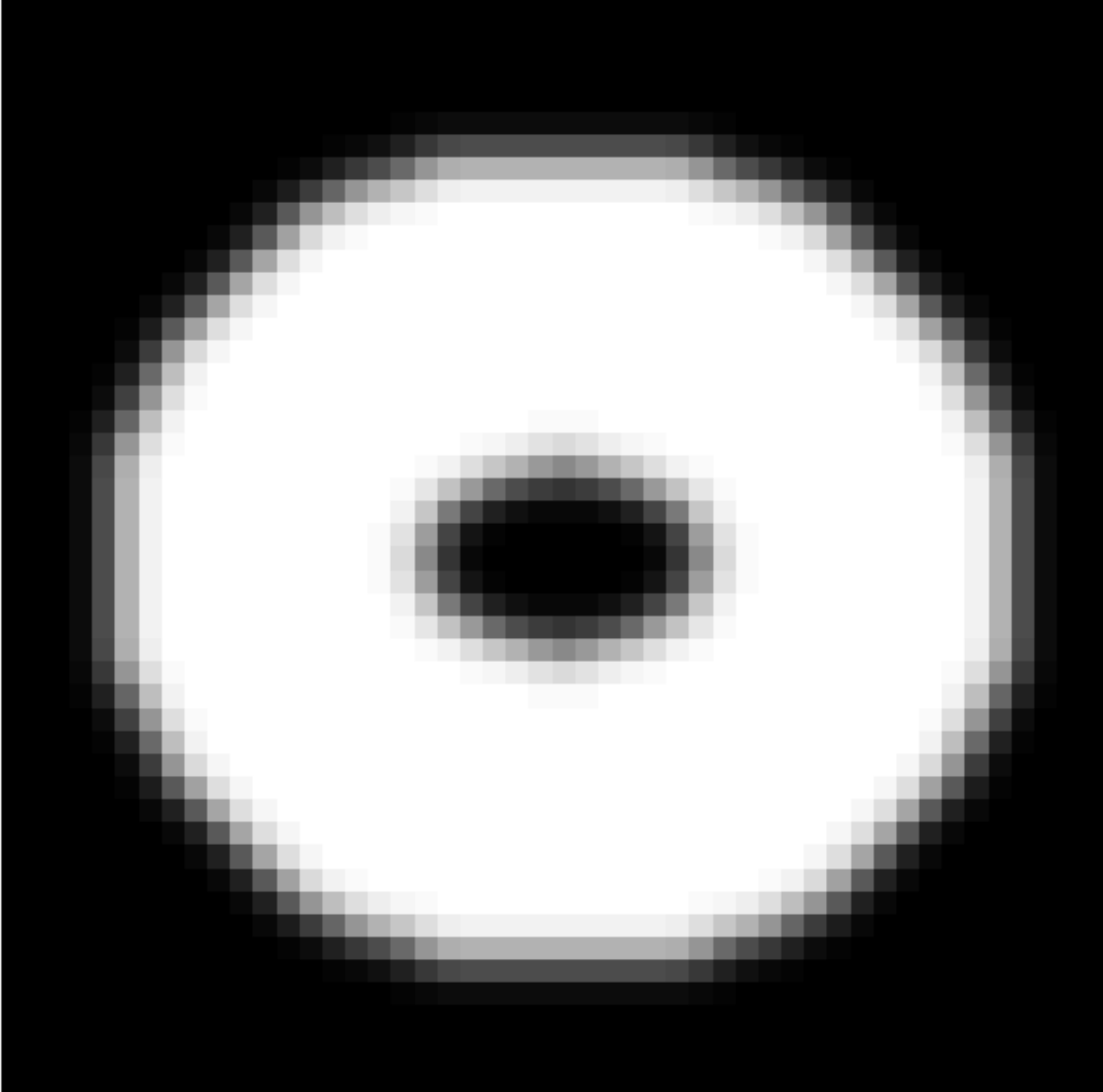} & 
\includegraphics[width=0.065\textwidth]{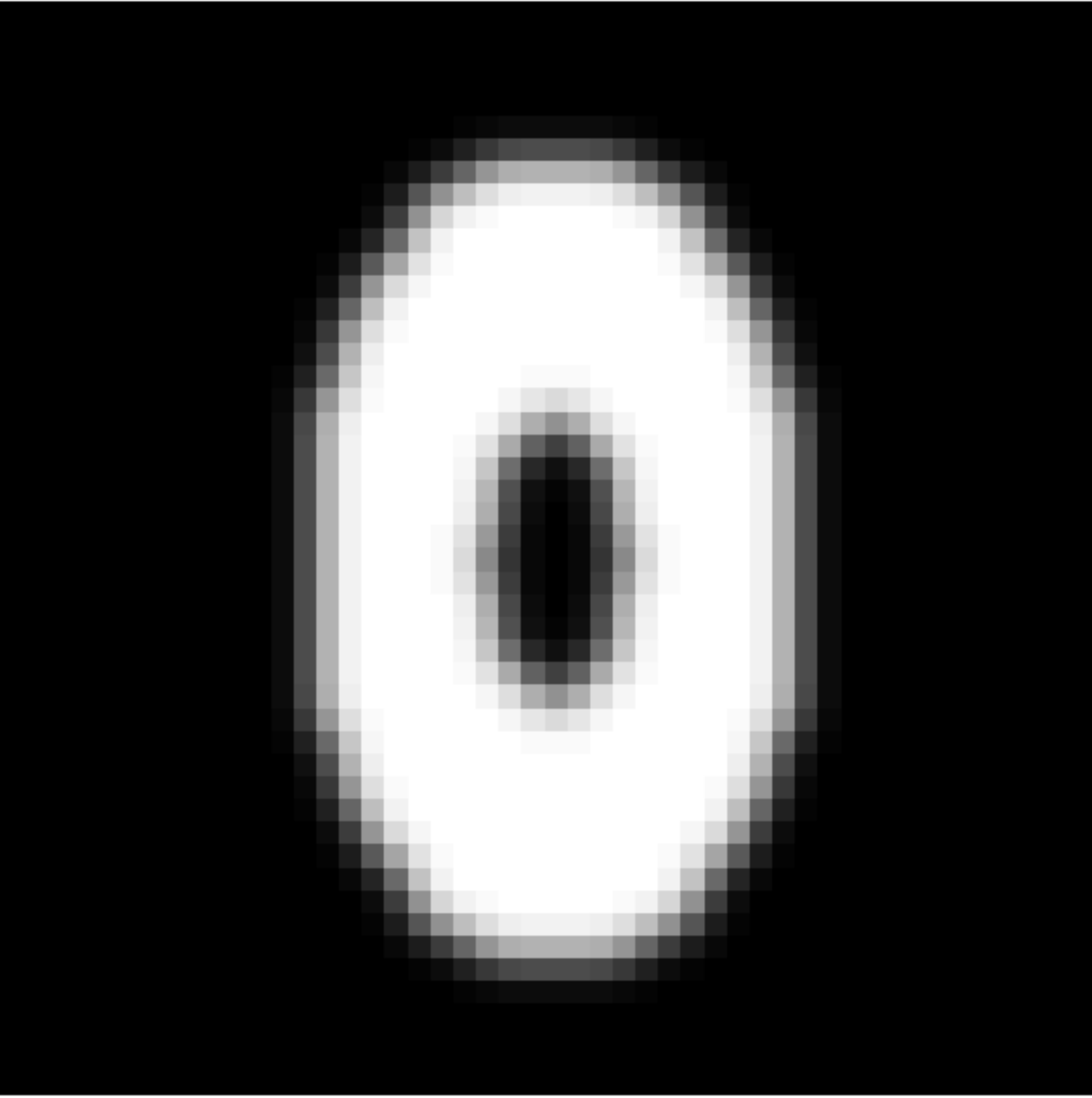} \\ 
\end{tabular}

\begin{tabular}{cc|cc|cc|cc|cc}
\scriptsize DD & \scriptsize SVF & \scriptsize St. LDDMM & \scriptsize SFV & \scriptsize Flash & \scriptsize $V_0$ & \scriptsize SVF-GAN & \scriptsize SVF & \scriptsize EPDiff-GAN & $V_0$ \\
\includegraphics[width=0.065\textwidth]{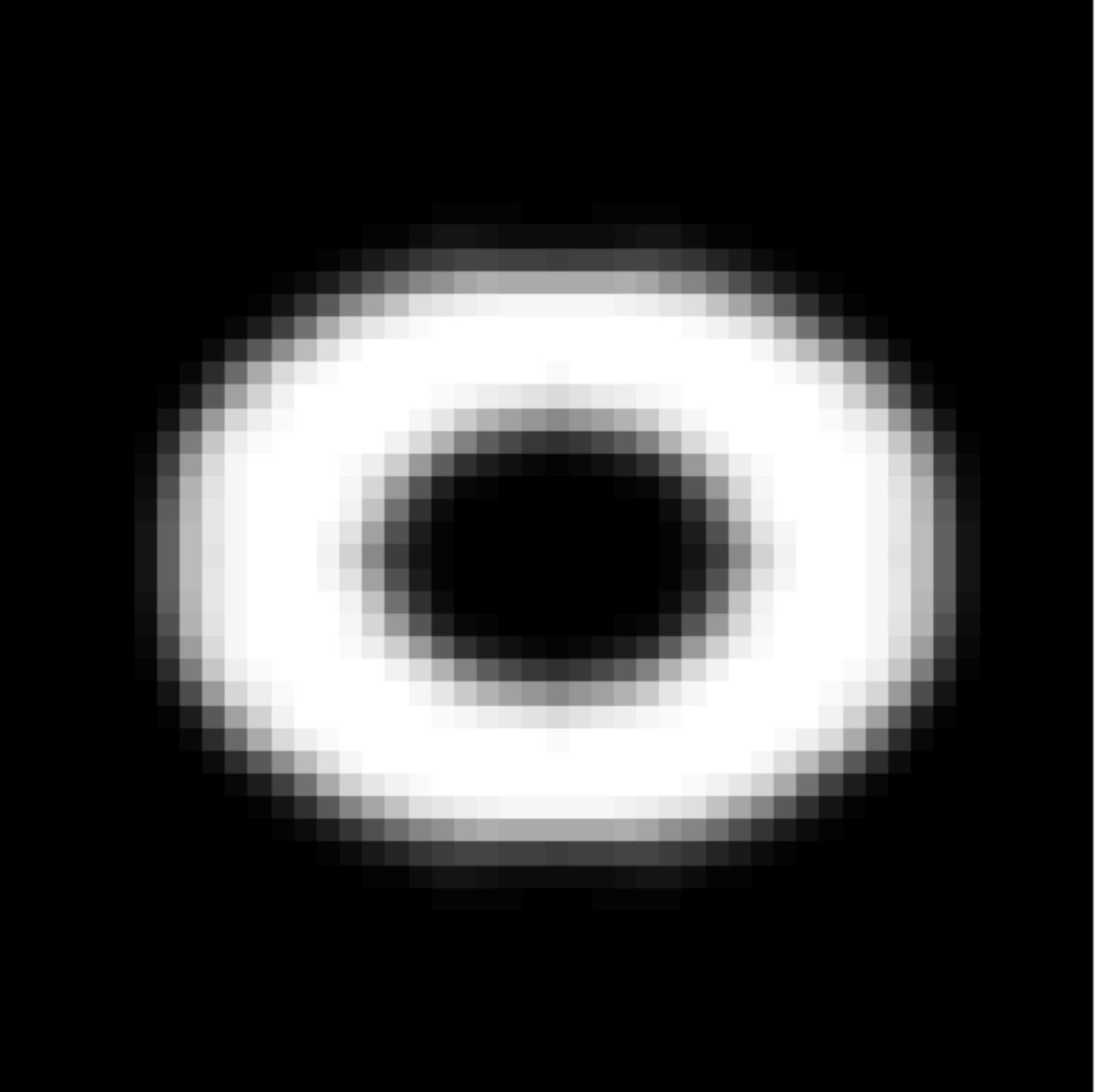} & 
\includegraphics[width=0.75cm, height = 0.75cm]{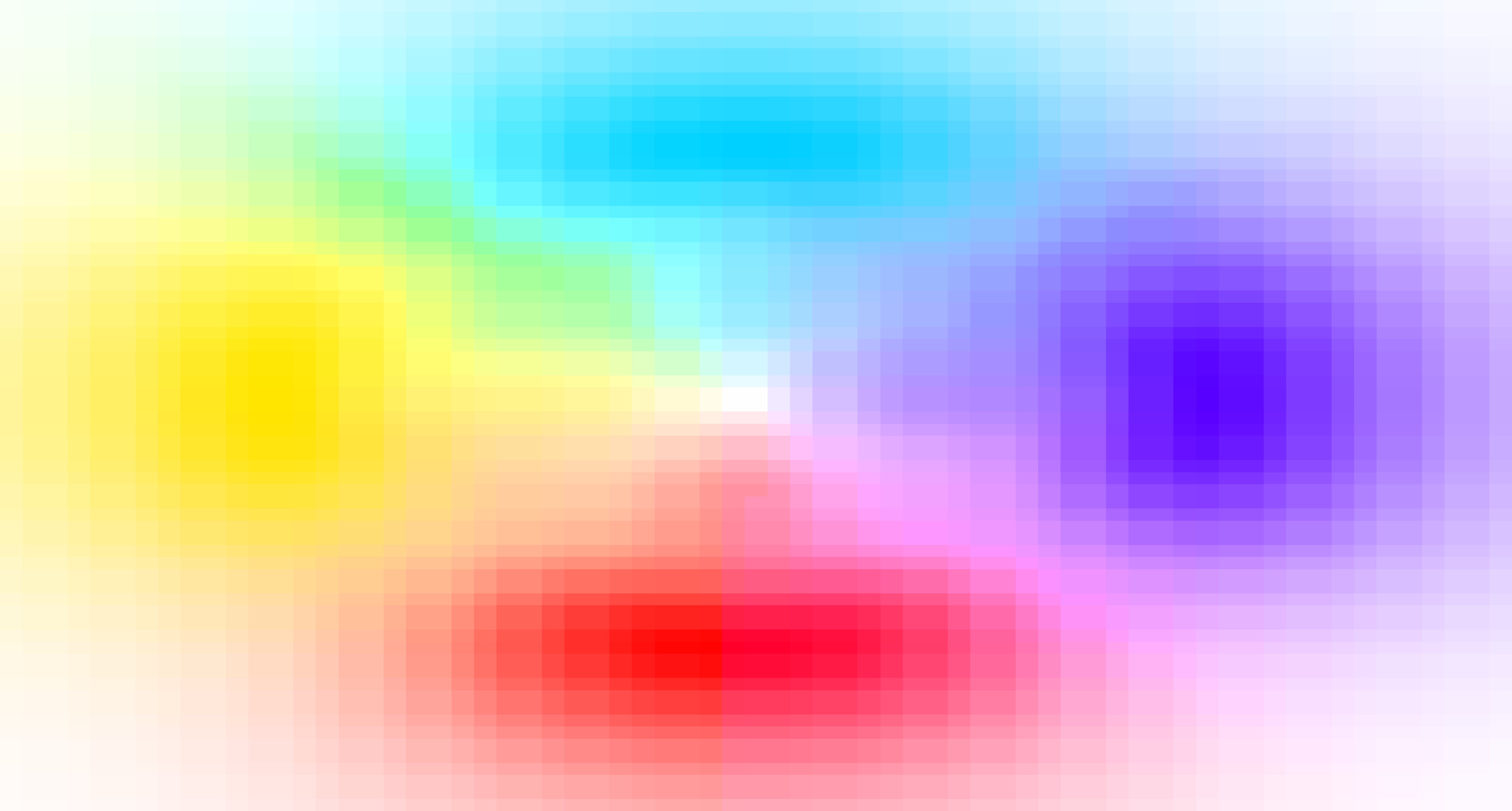} & 
\includegraphics[width=0.065\textwidth]{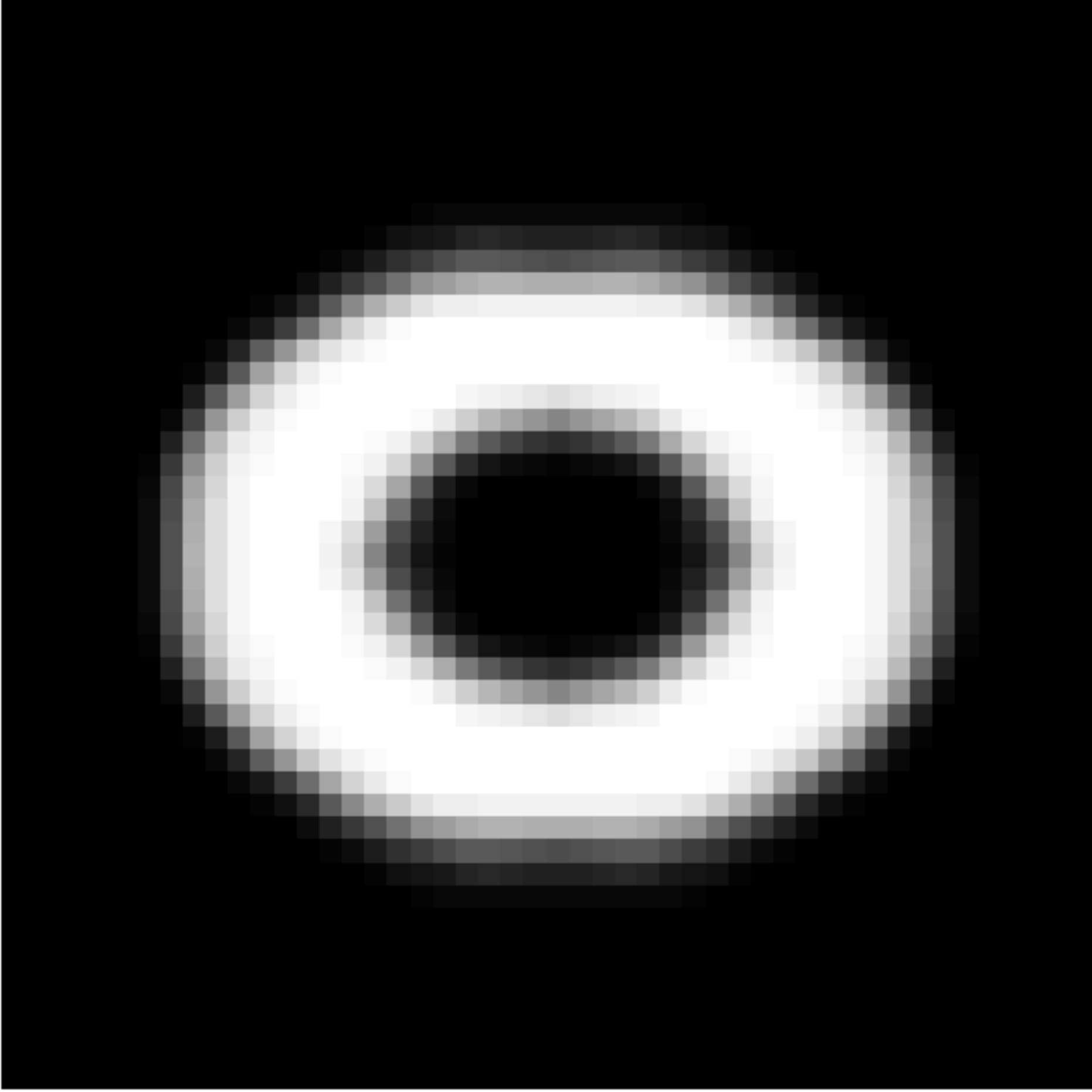} & 
\includegraphics[width=0.75cm, height = 0.75cm]{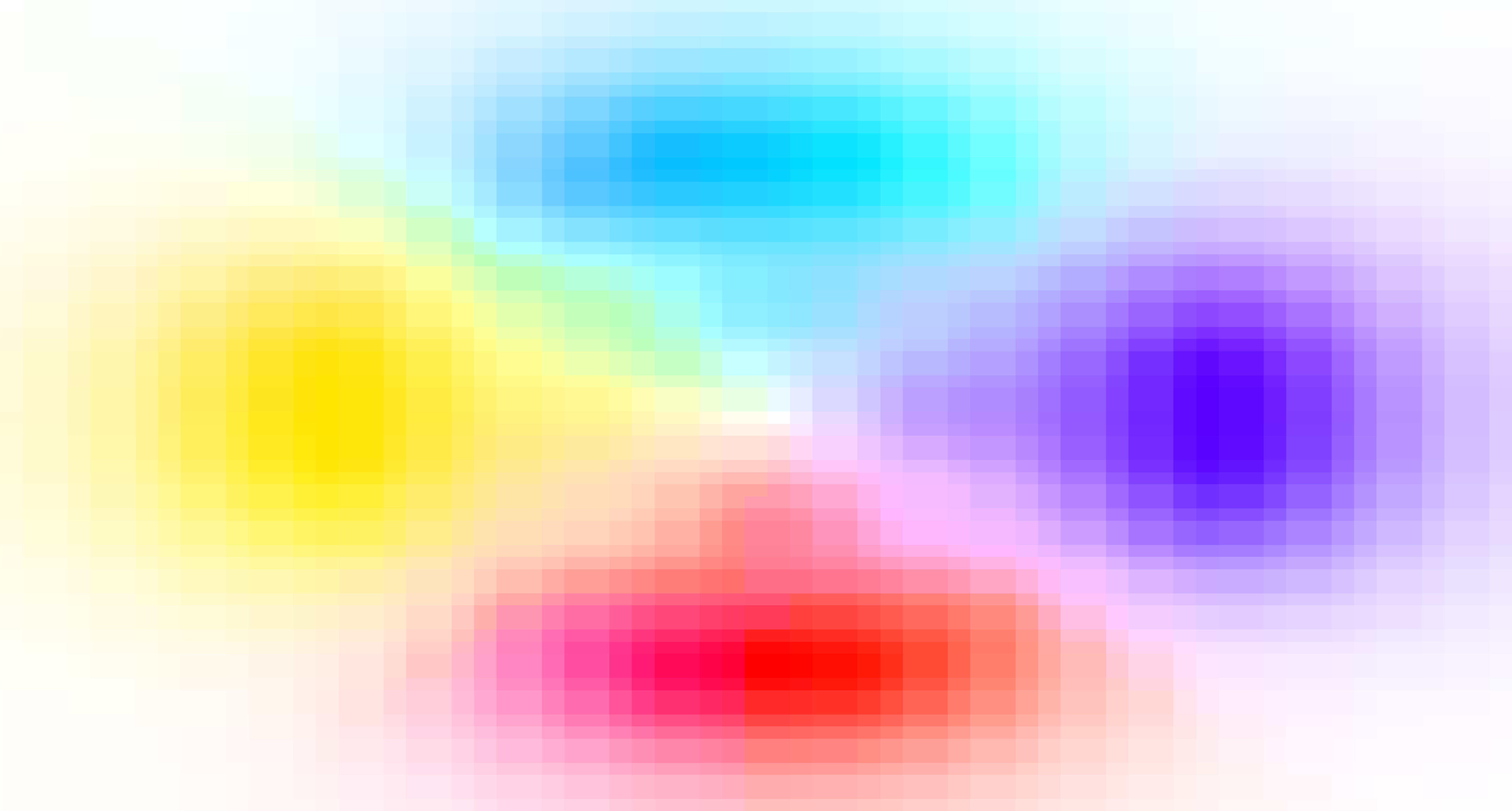} & 
\includegraphics[width=0.065\textwidth]{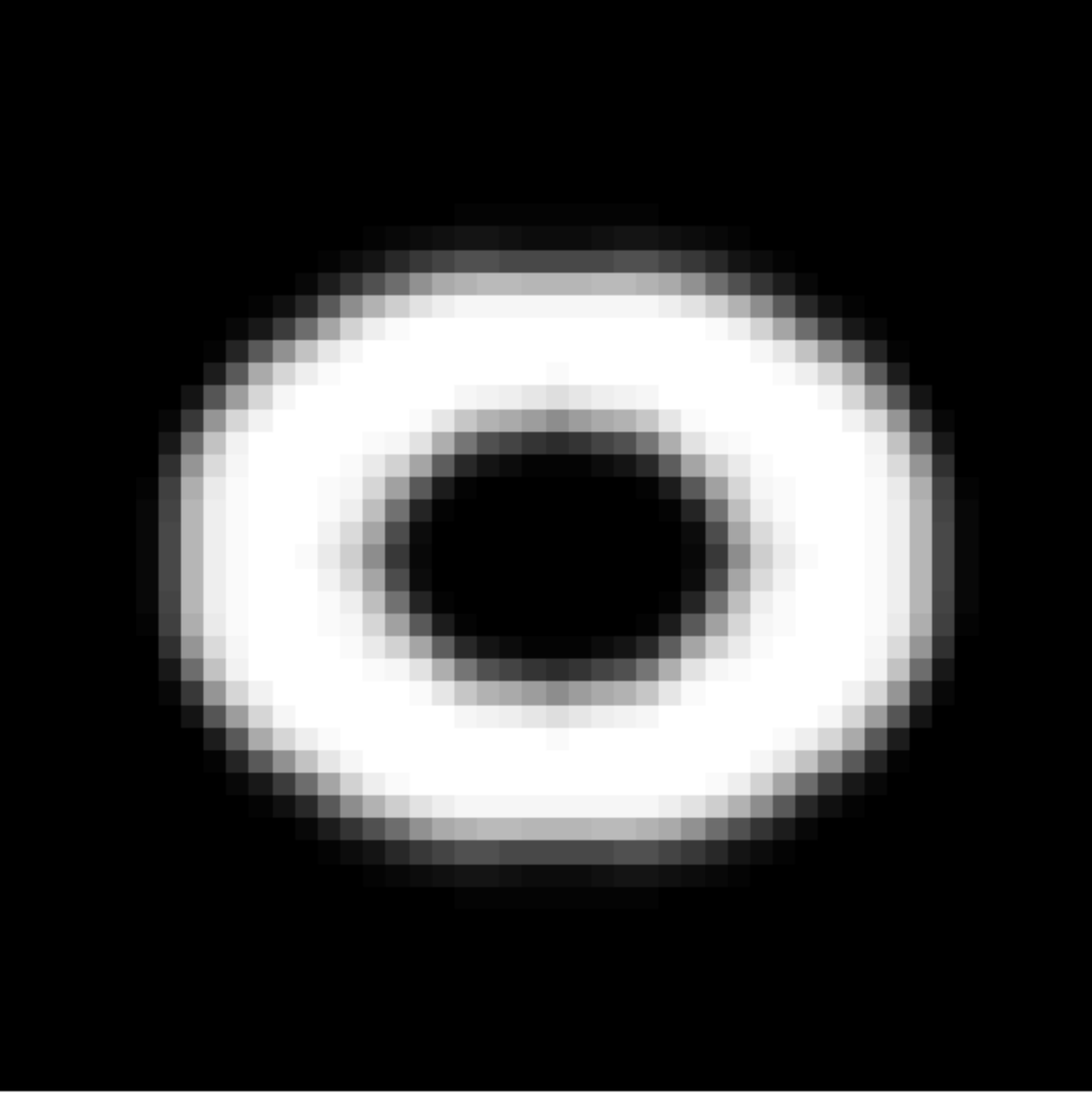} & 
\includegraphics[width=0.75cm, height = 0.75cm]{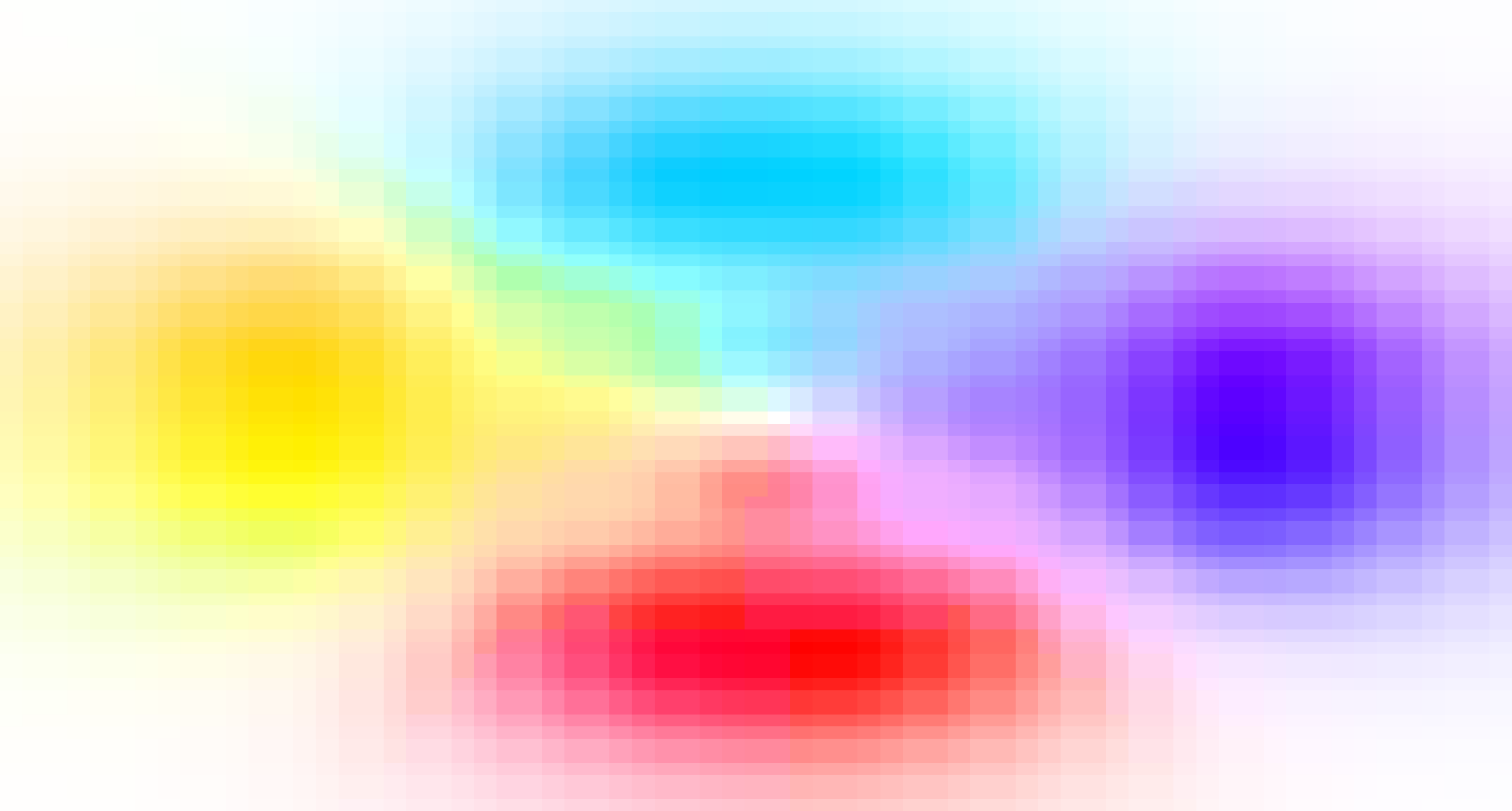} & 
\includegraphics[width=0.065\textwidth]{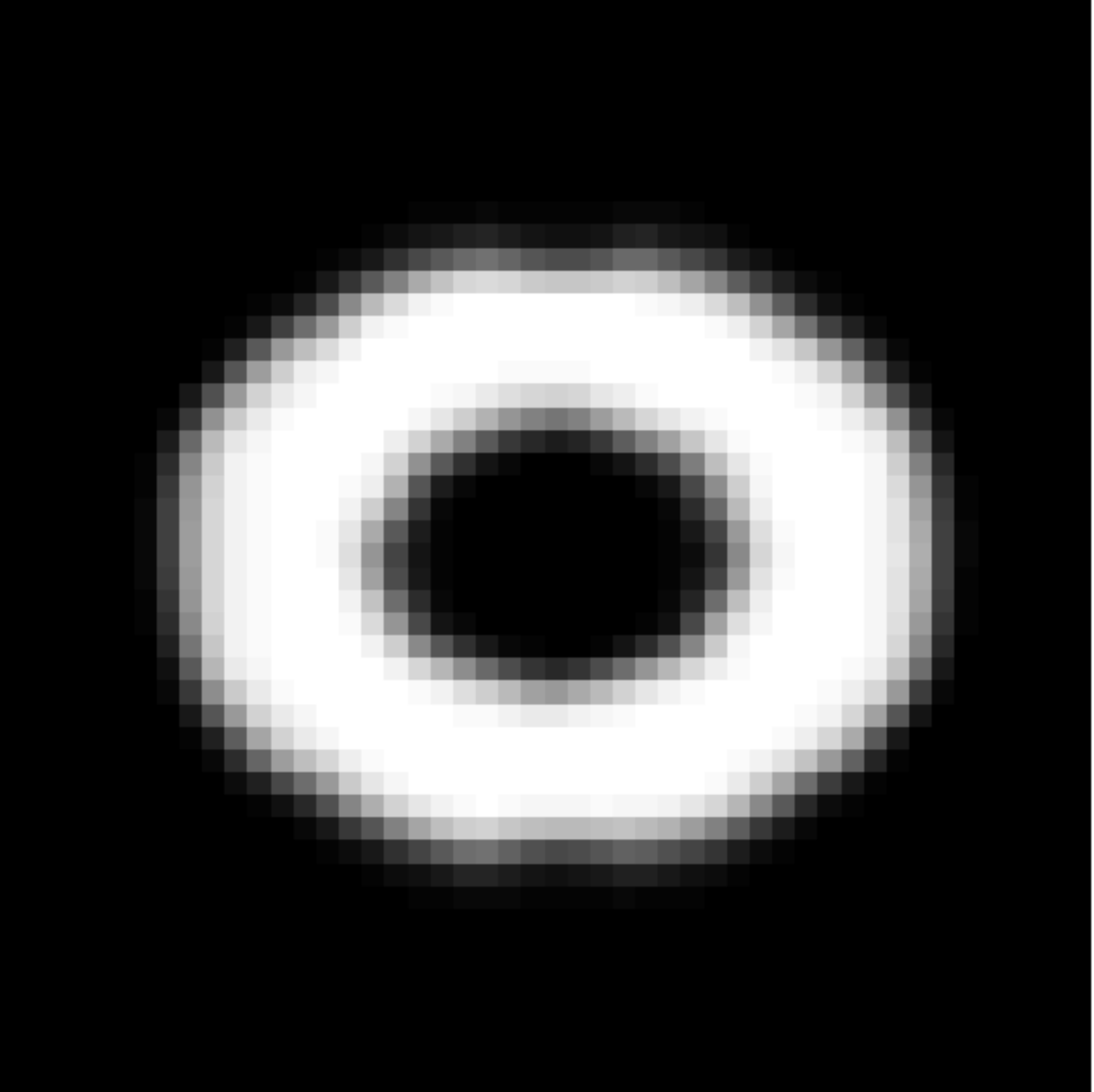} &
\includegraphics[width=0.75cm, height = 0.75cm]{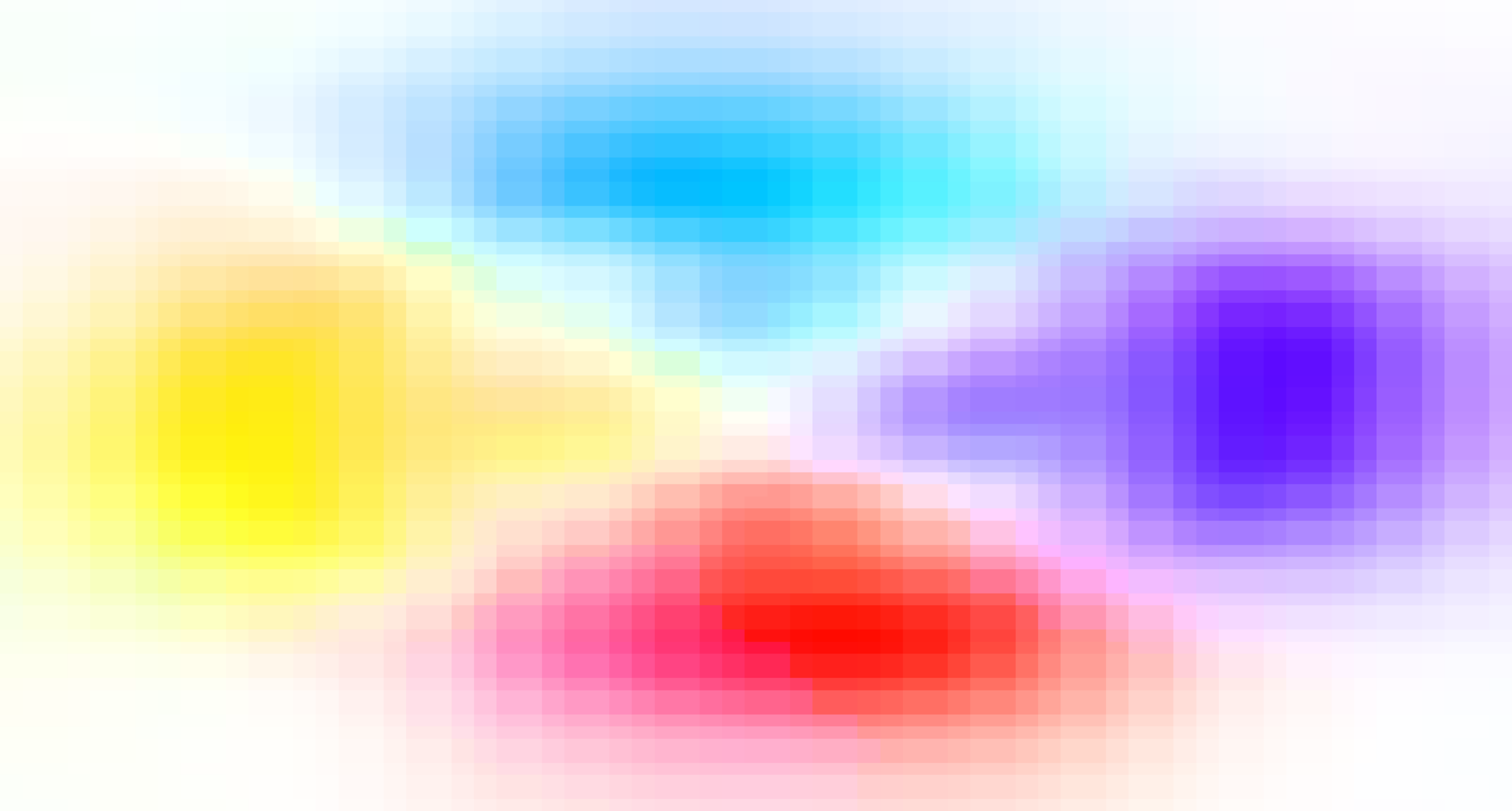} &
\includegraphics[width=0.065\textwidth]{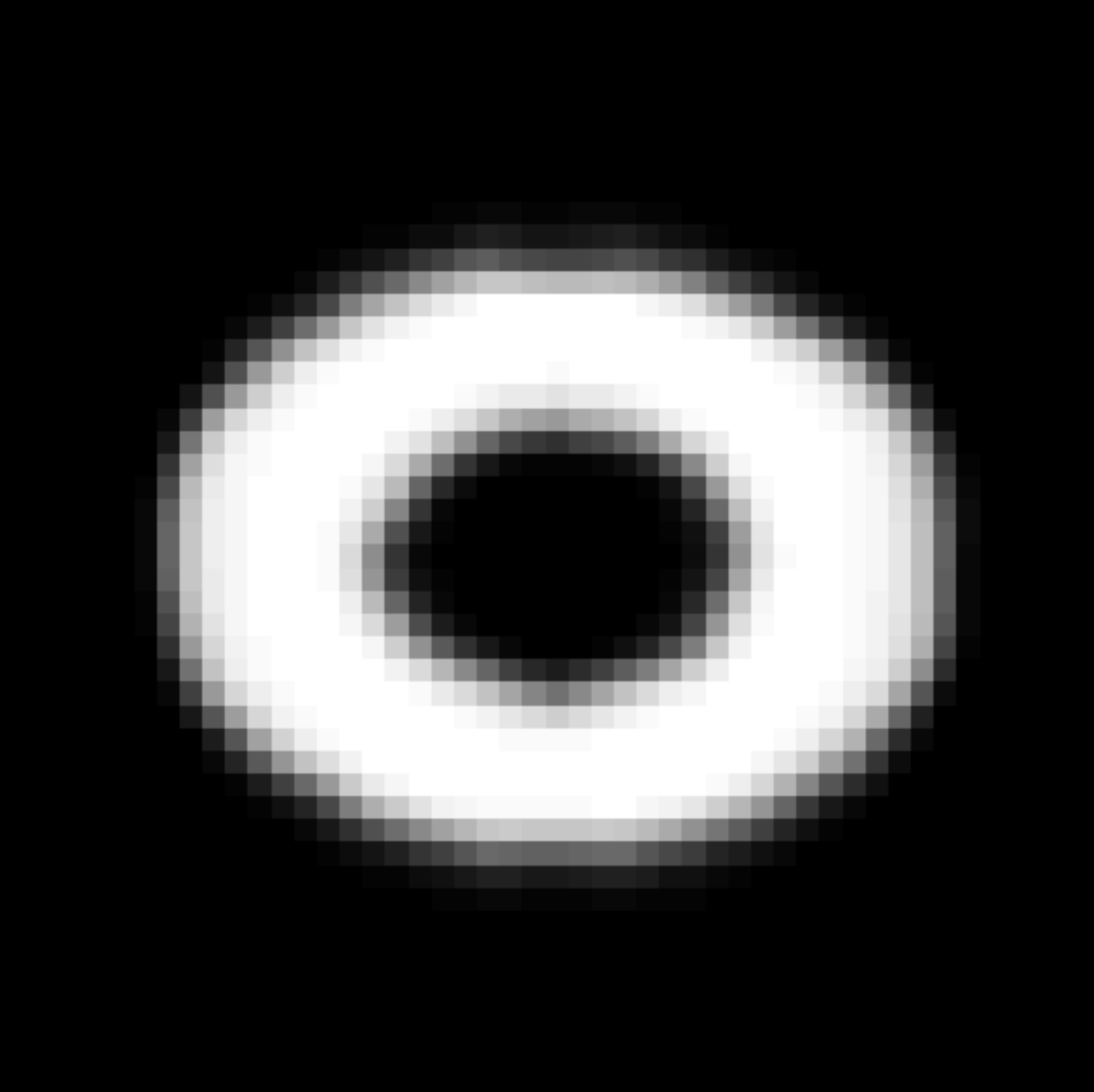} &
\includegraphics[width=0.75cm, height = 0.75cm]{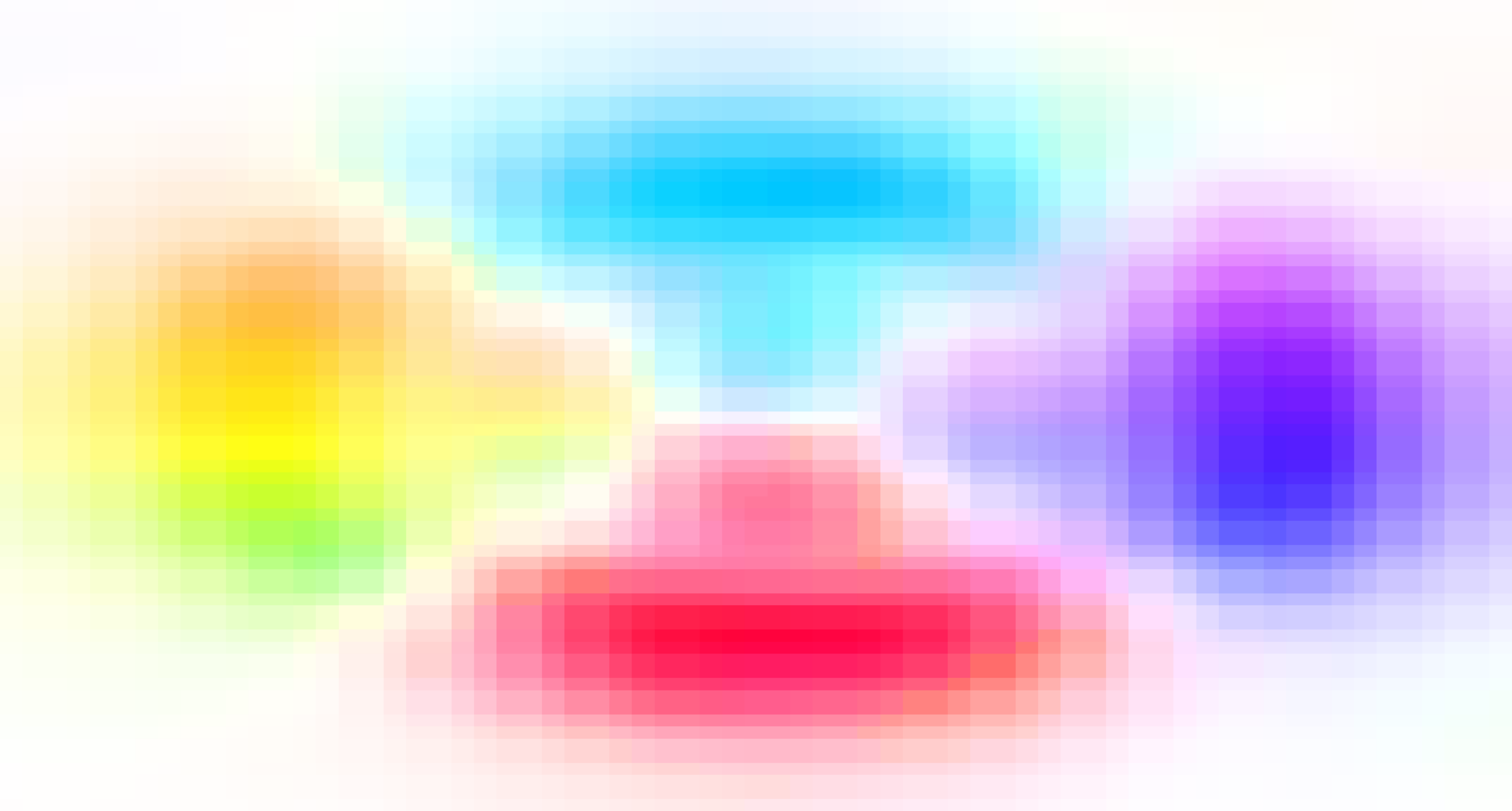} 
\\ 
\includegraphics[width=0.065\textwidth]{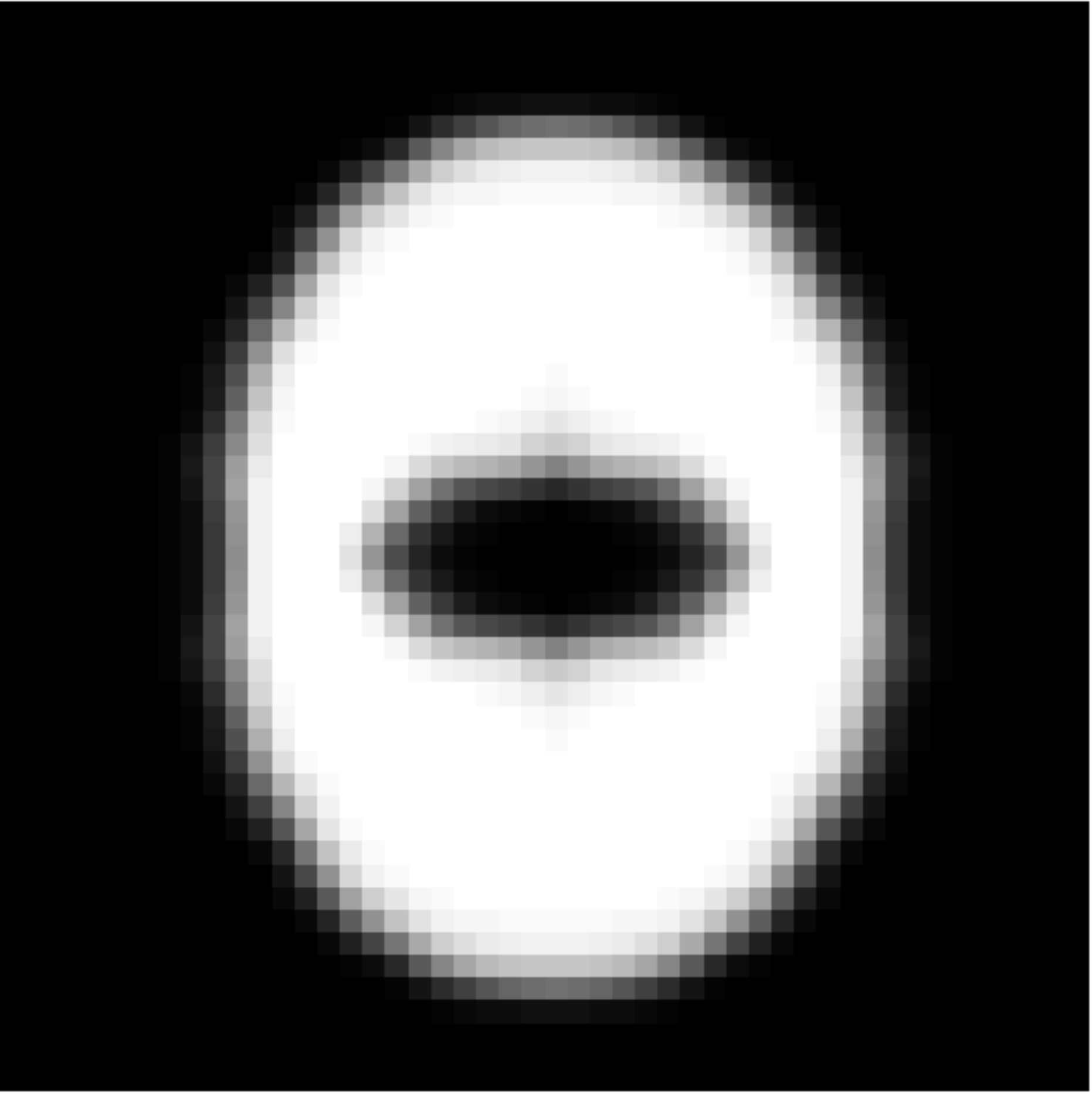} & 
\includegraphics[width=0.75cm, height = 0.75cm]{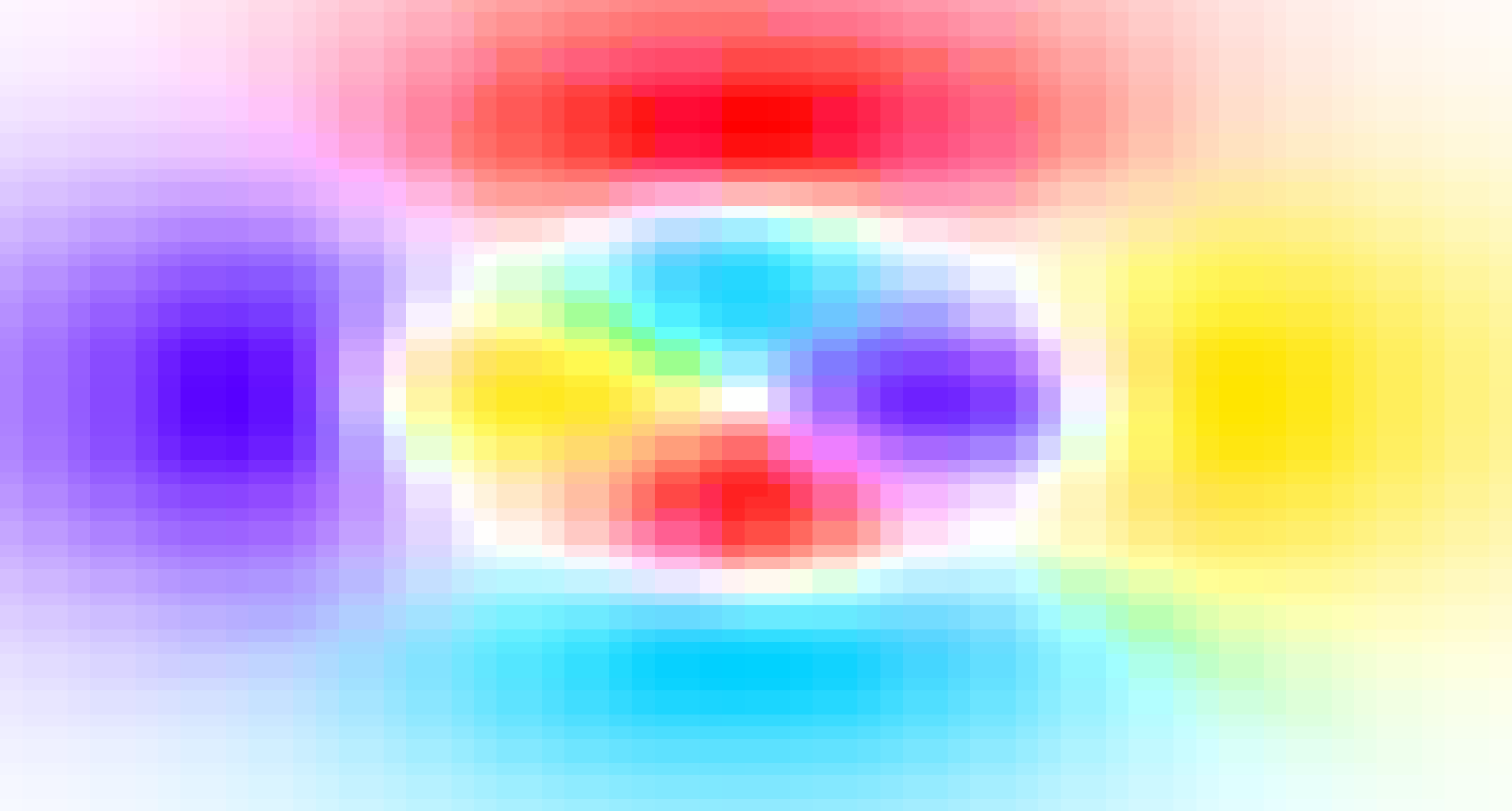} & 
\includegraphics[width=0.065\textwidth]{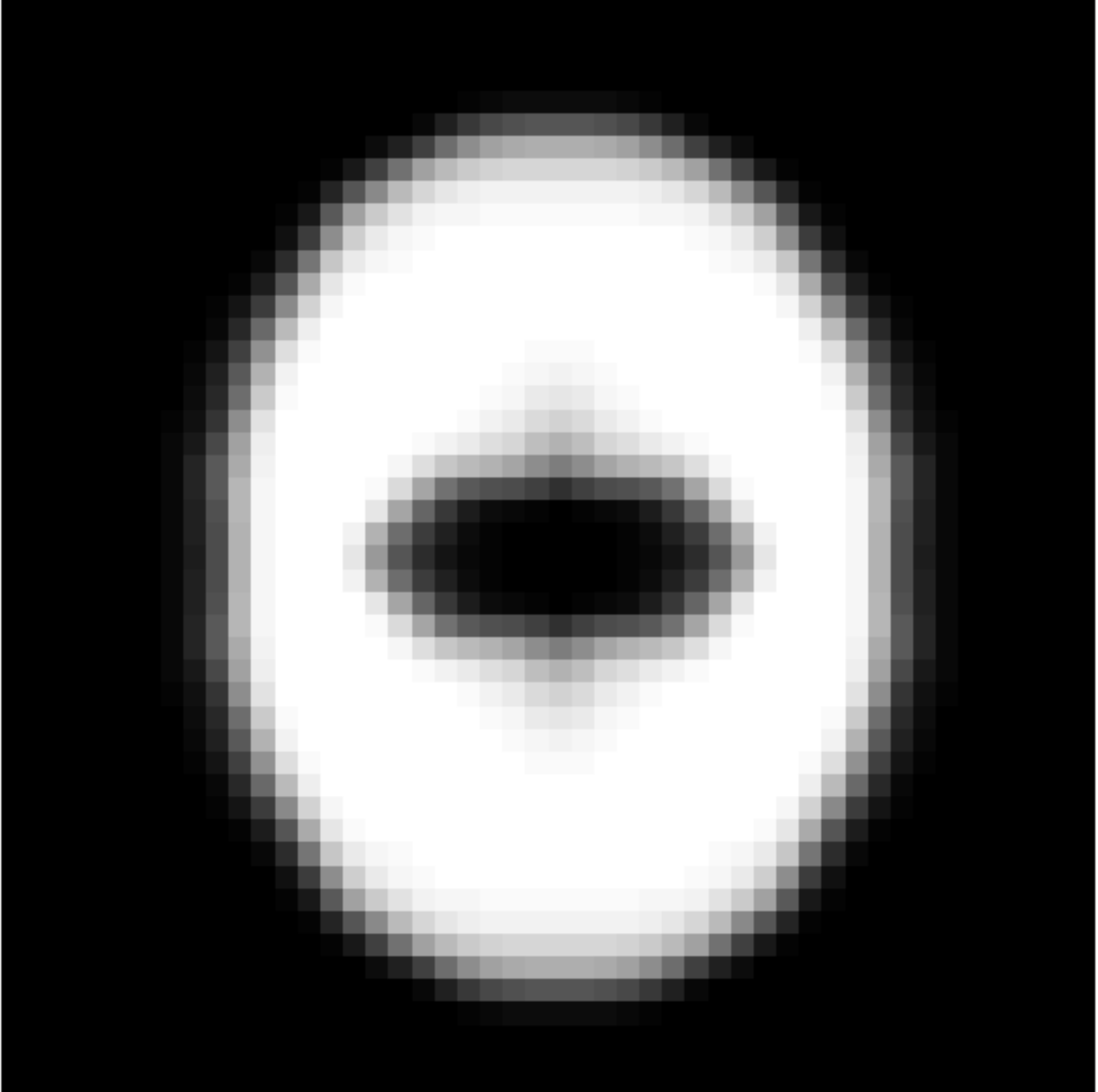} & 
\includegraphics[width=0.75cm, height = 0.75cm]{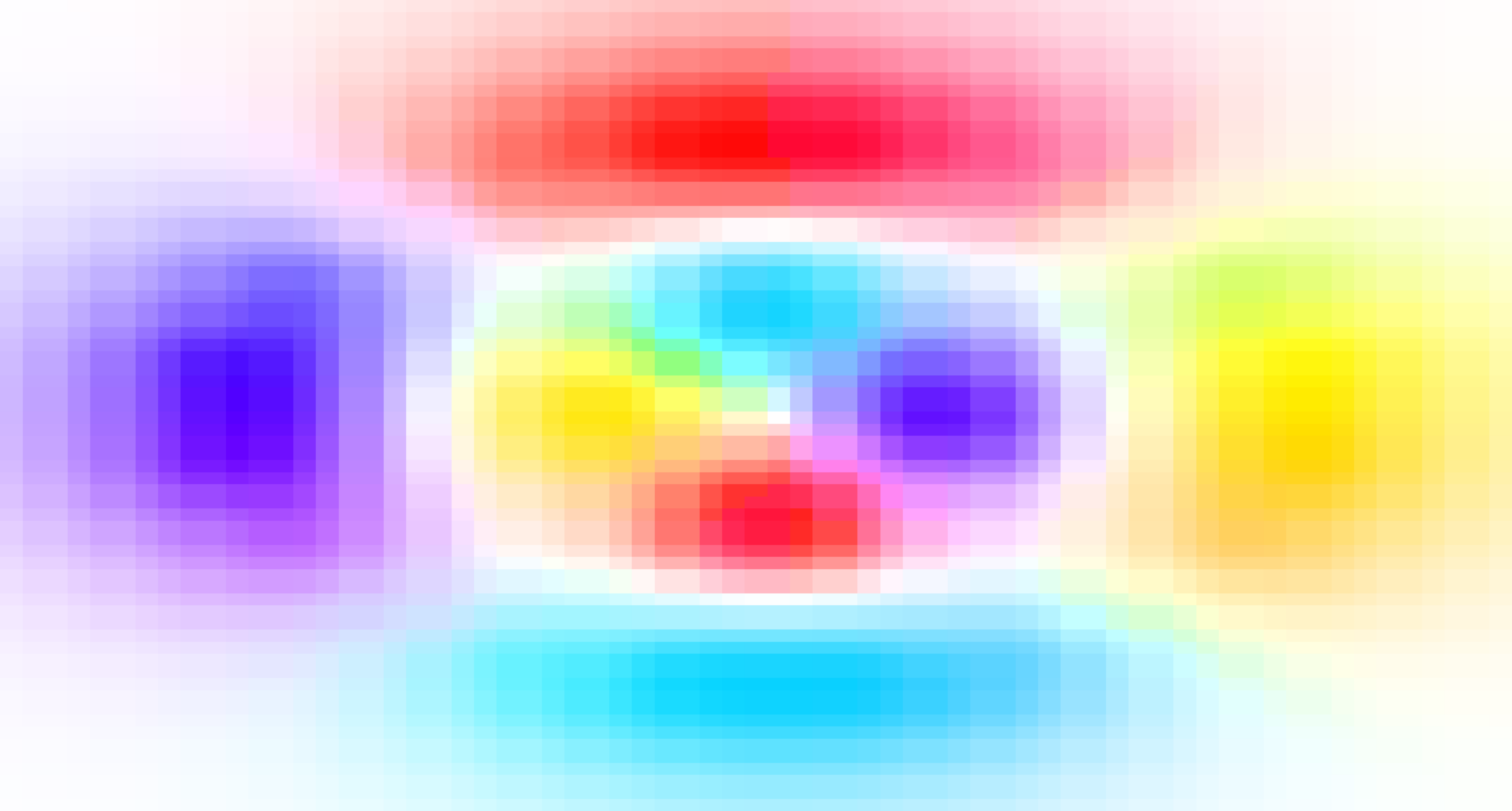} & 
\includegraphics[width=0.065\textwidth]{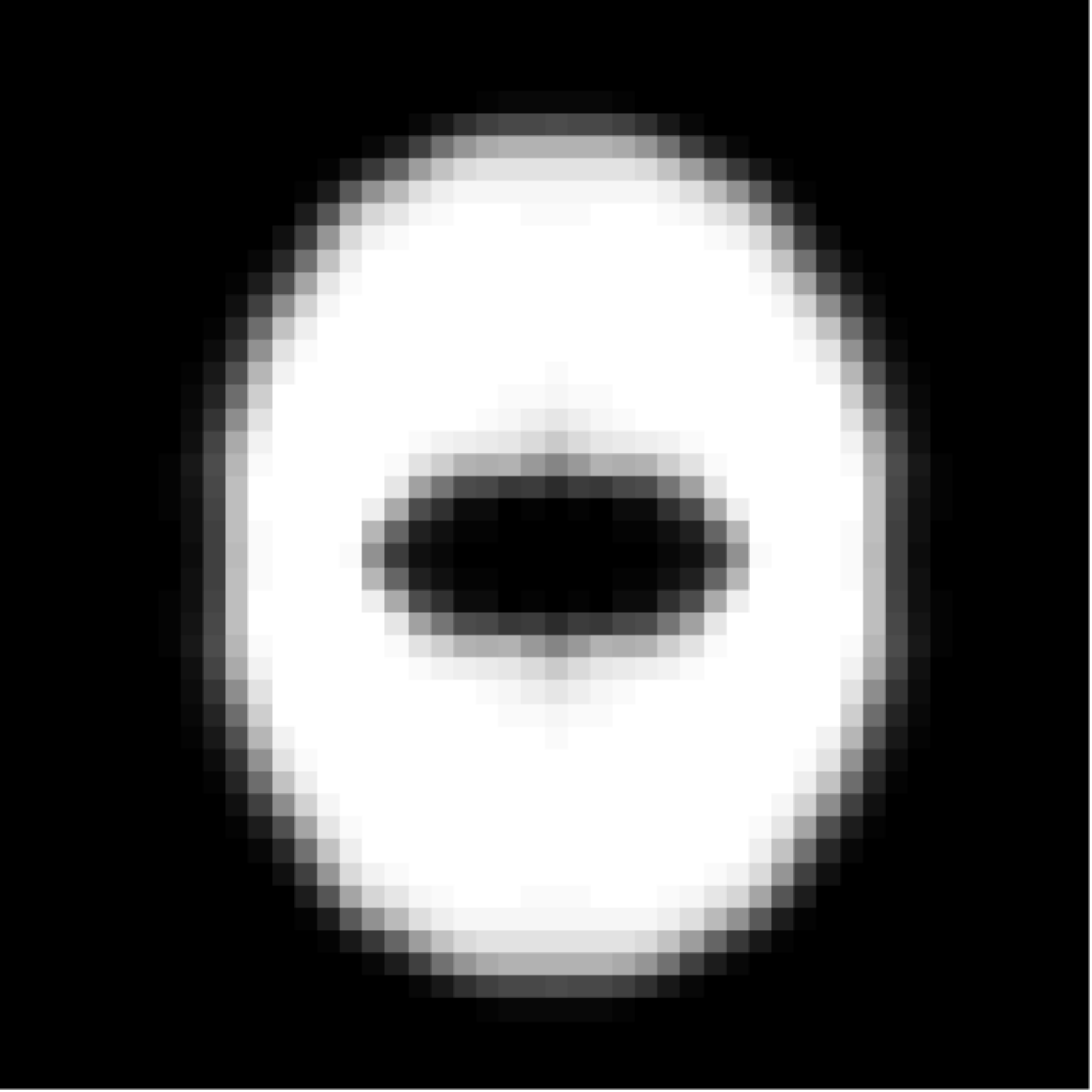} & 
\includegraphics[width=0.75cm, height = 0.75cm]{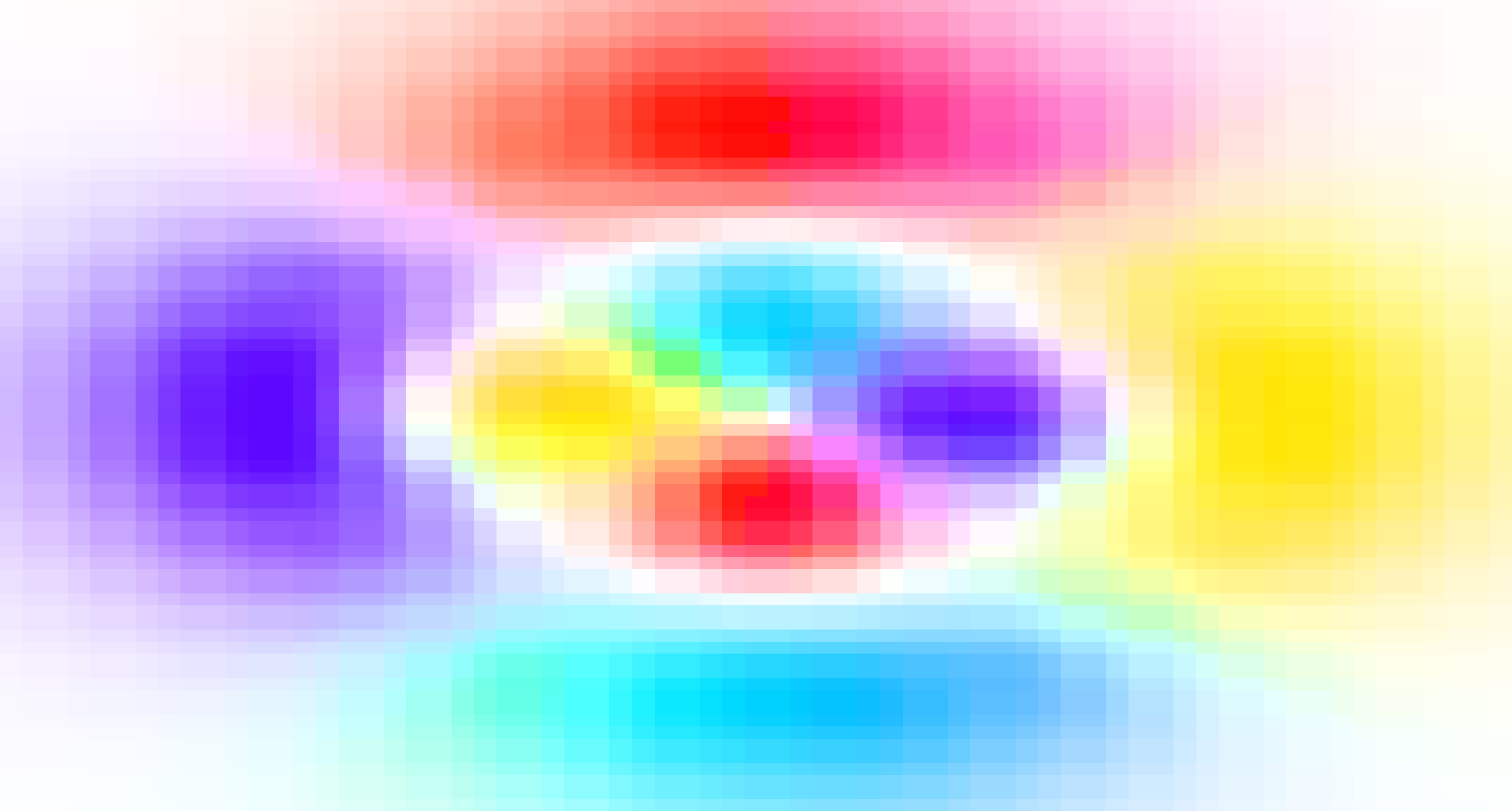} & 
\includegraphics[width=0.065\textwidth]{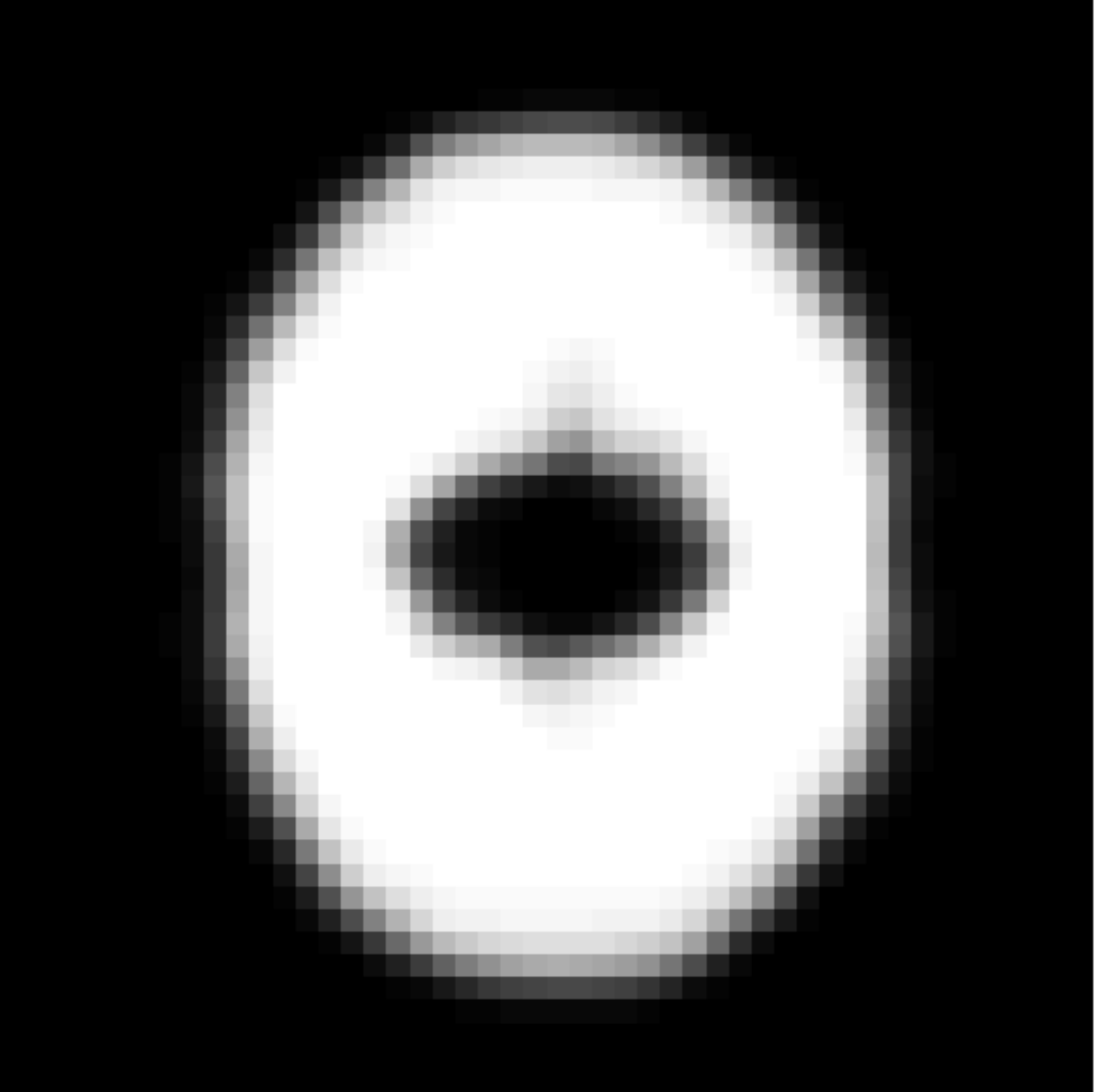} &
\includegraphics[width=0.75cm, height = 0.75cm]{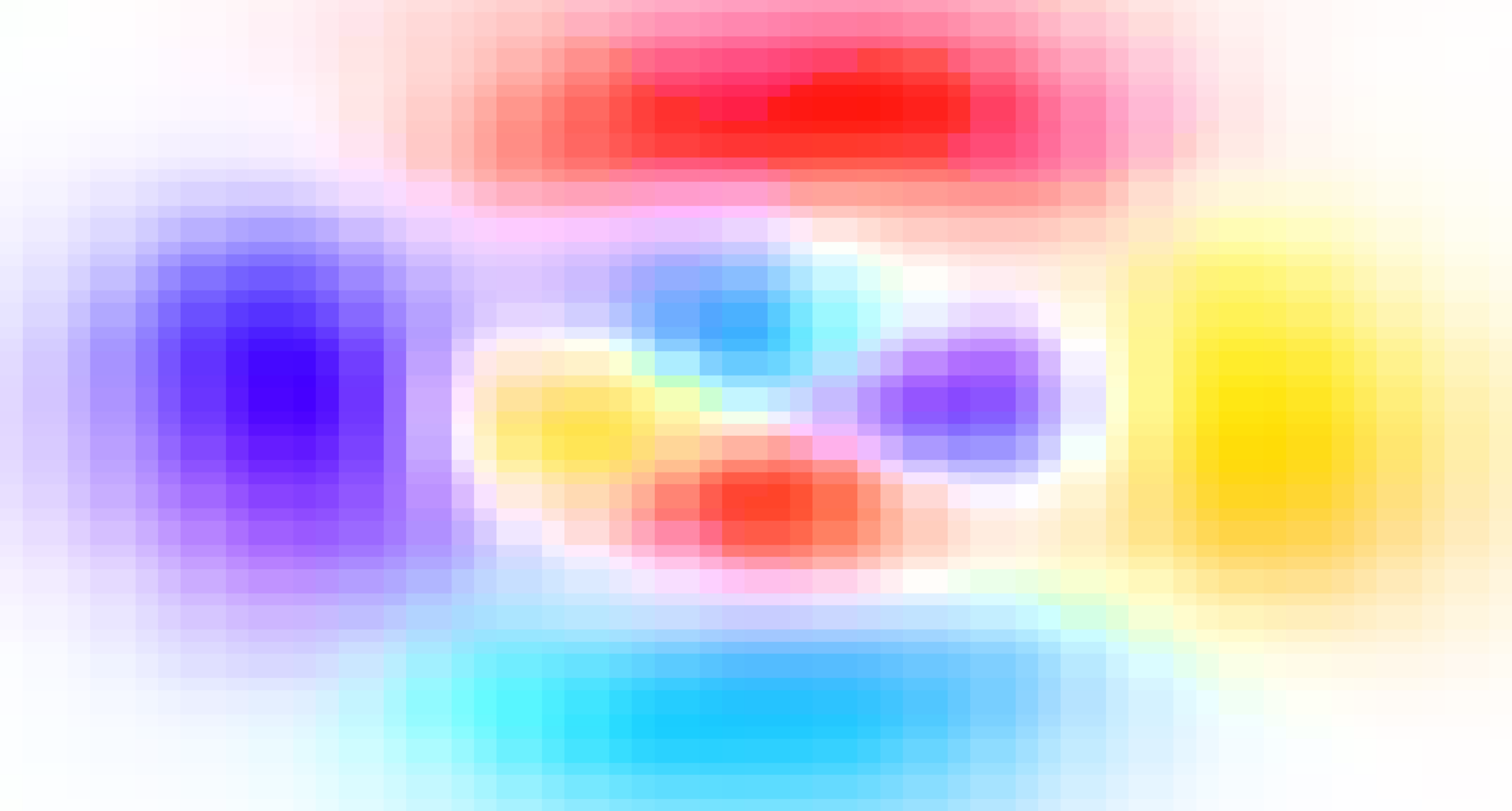} &
\includegraphics[width=0.065\textwidth]{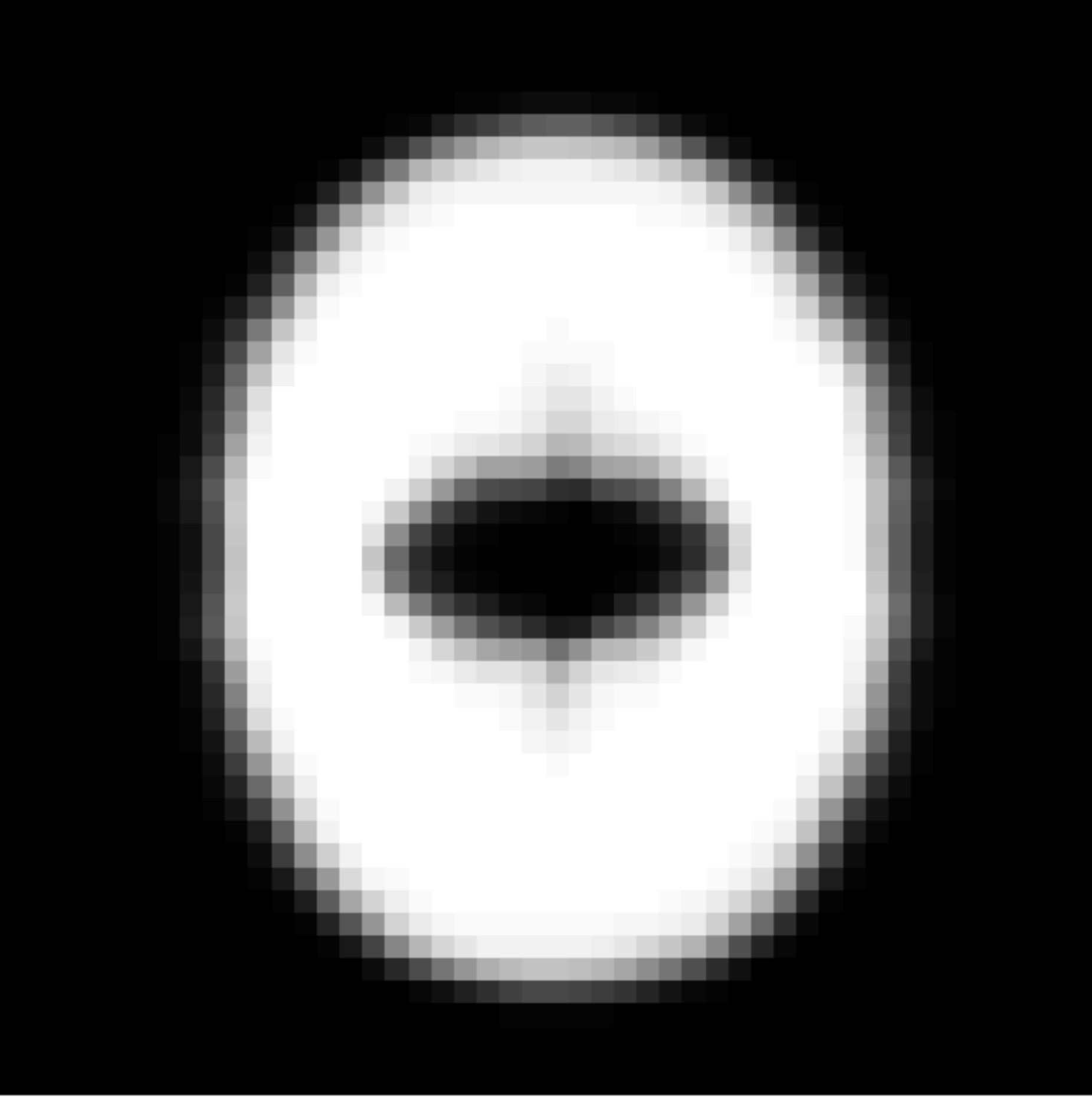} &
\includegraphics[width=0.75cm, height = 0.75cm]{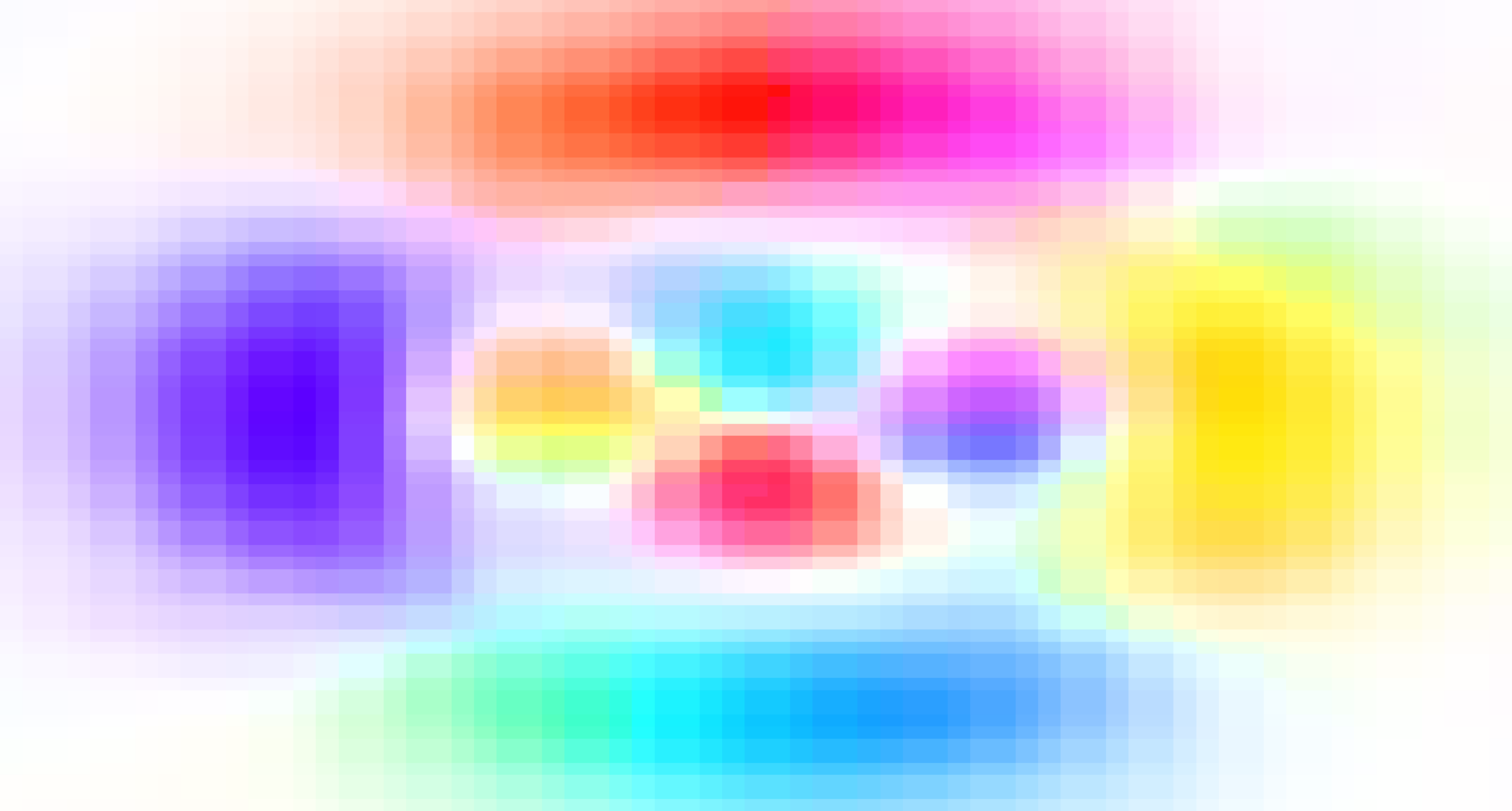} 
\\ 
\includegraphics[width=0.065\textwidth]{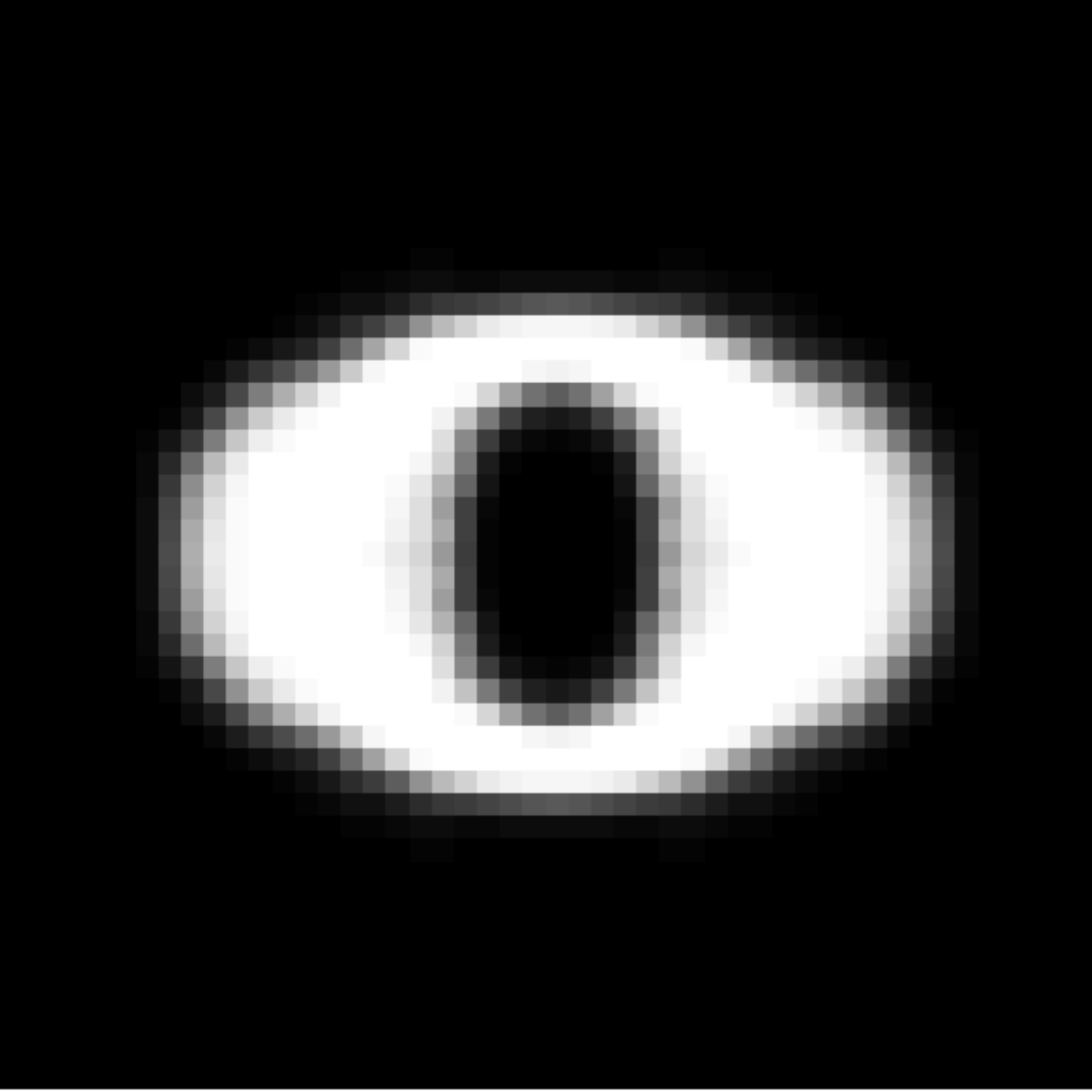} & 
\includegraphics[width=0.75cm, height = 0.75cm]{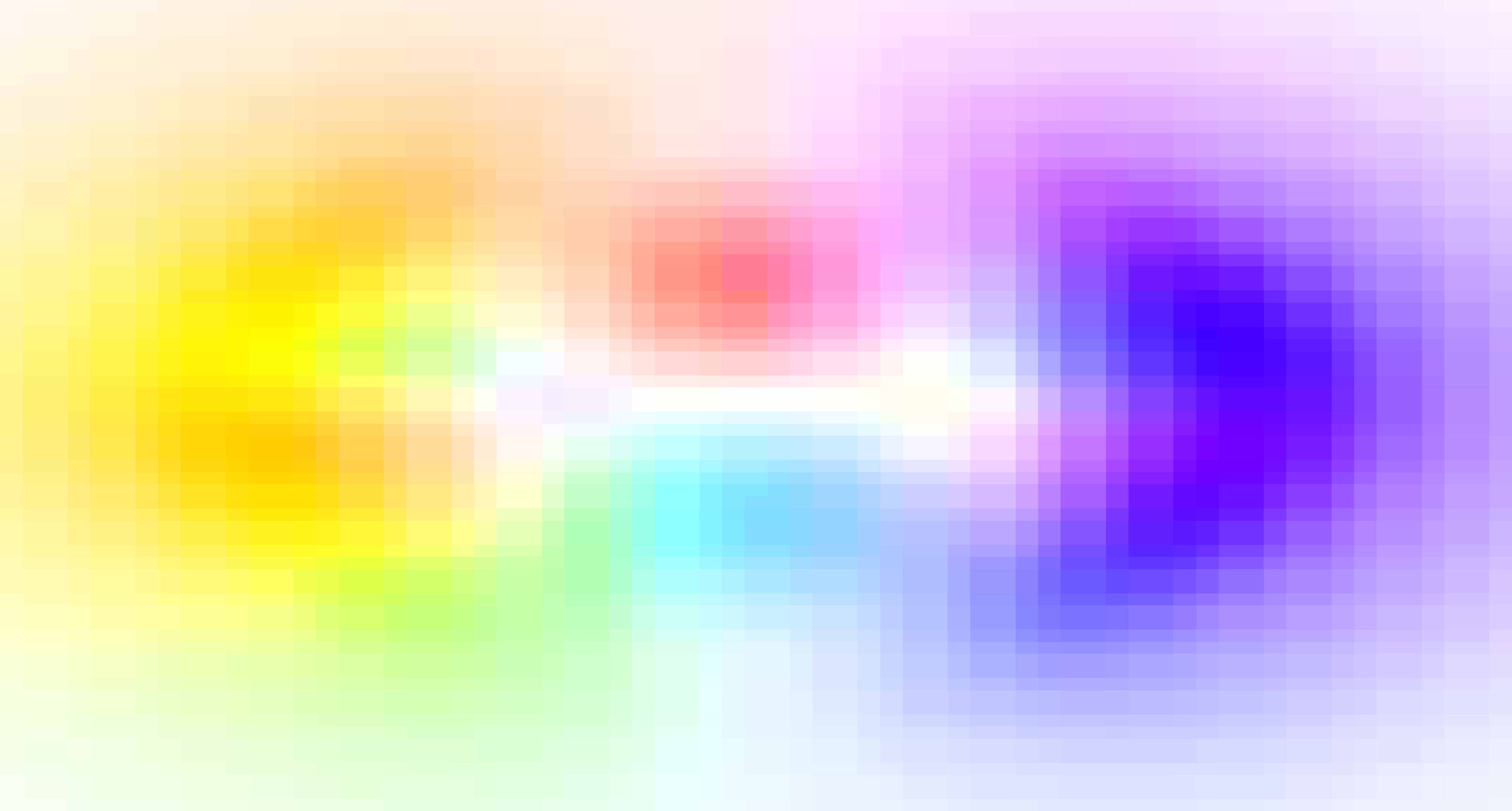} & 
\includegraphics[width=0.065\textwidth]{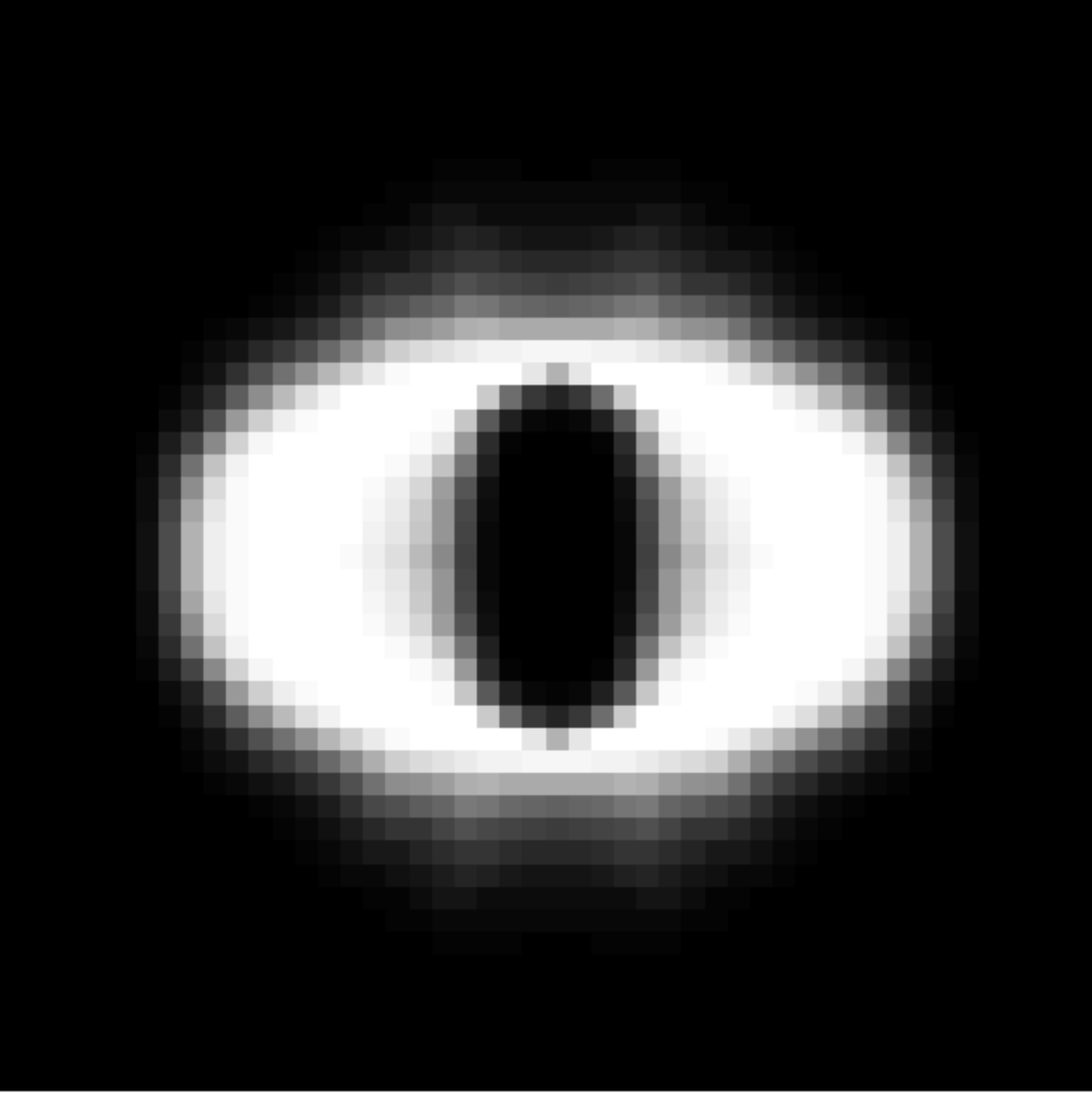} & 
\includegraphics[width=0.75cm, height = 0.75cm]{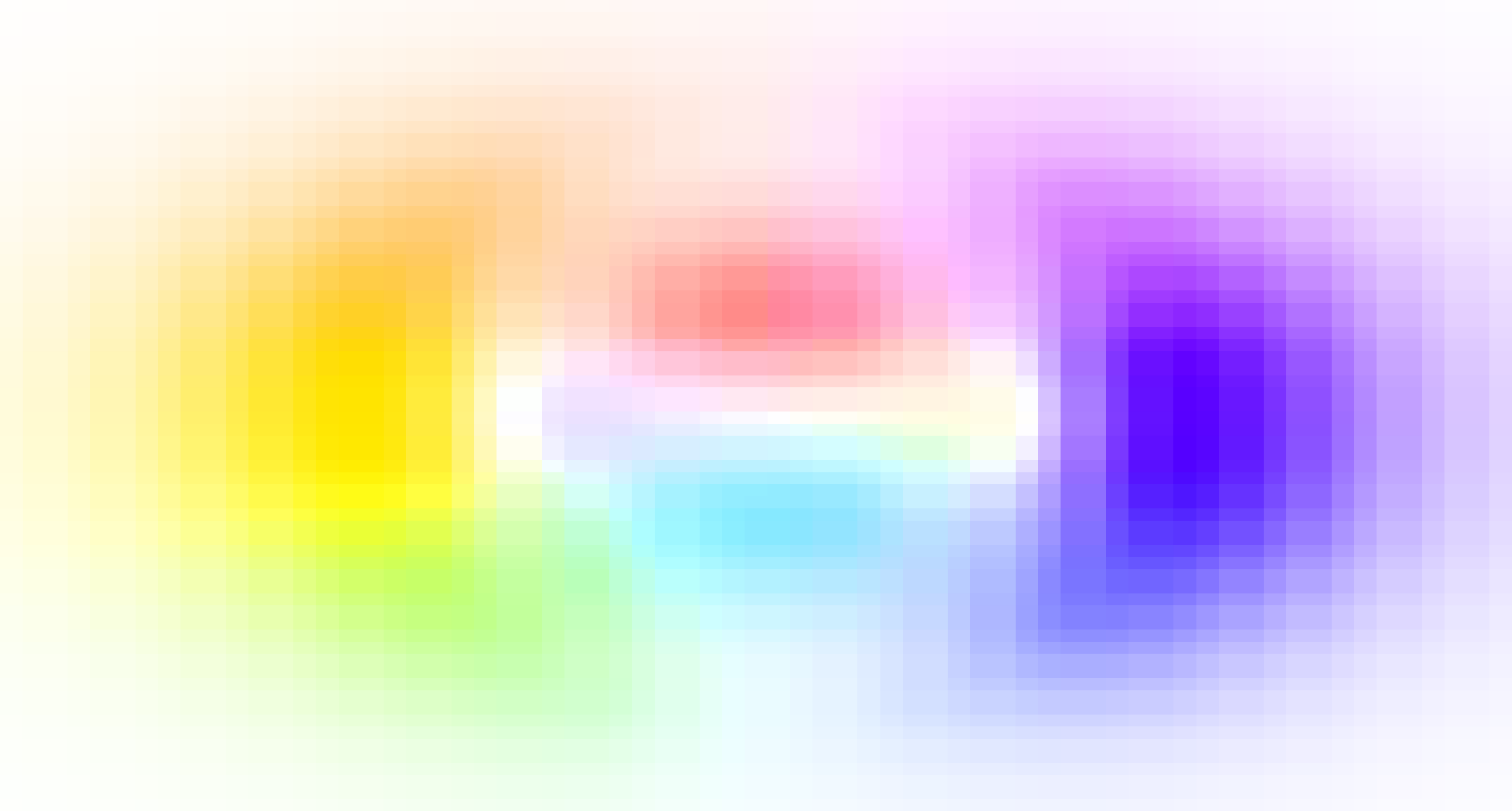} & 
\includegraphics[width=0.065\textwidth]{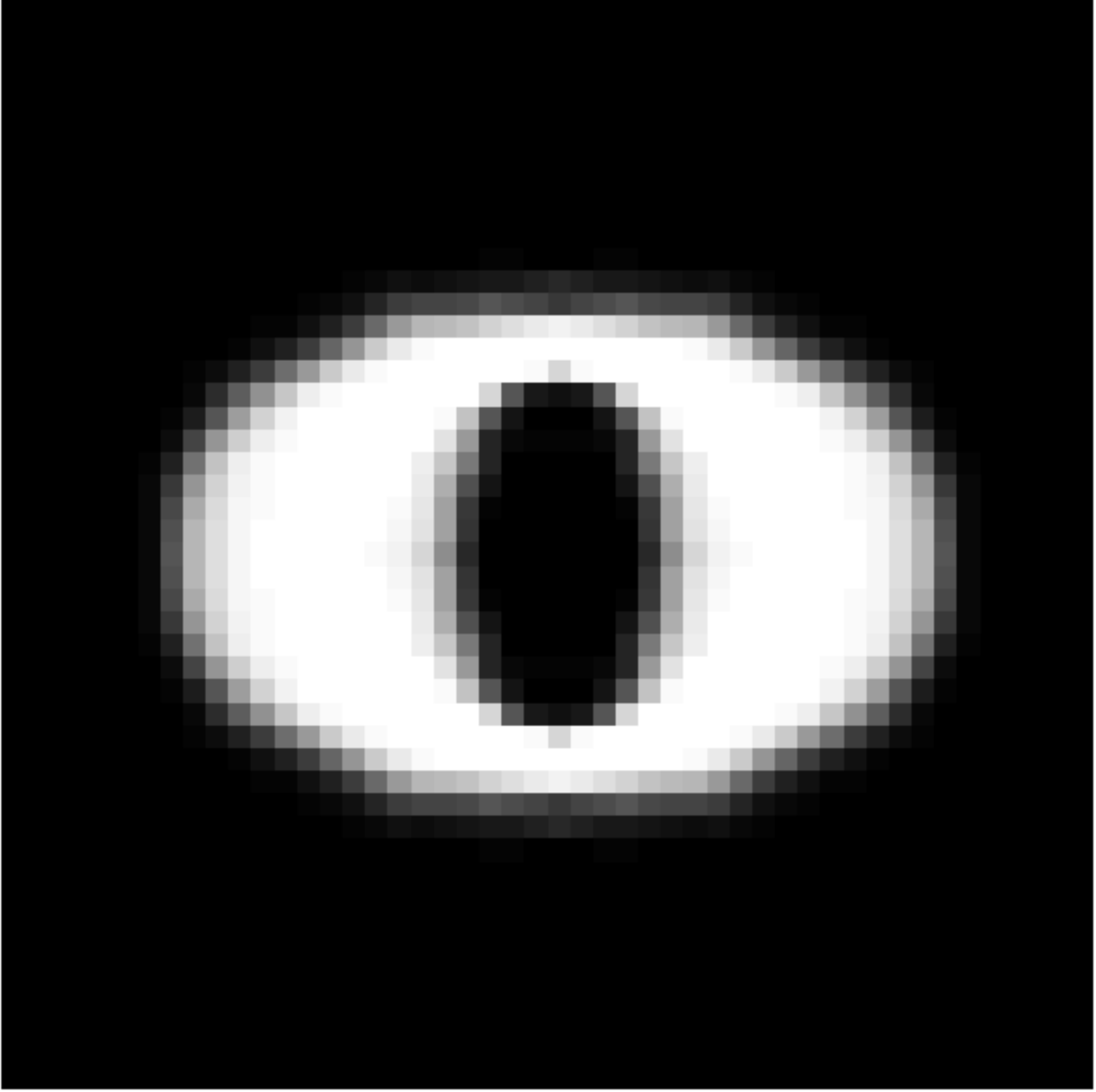} & 
\includegraphics[width=0.75cm, height = 0.75cm]{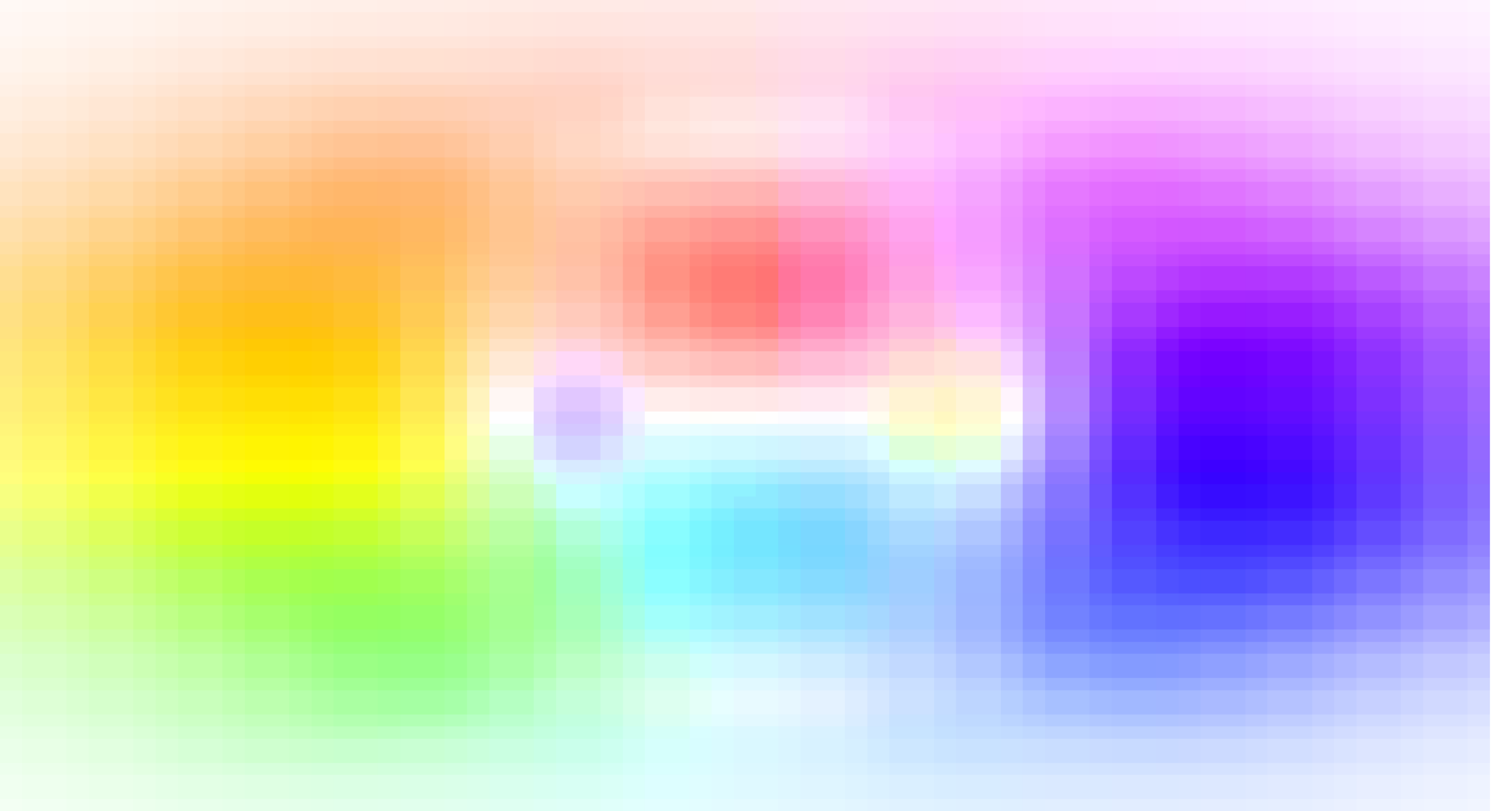} & 
\includegraphics[width=0.065\textwidth]{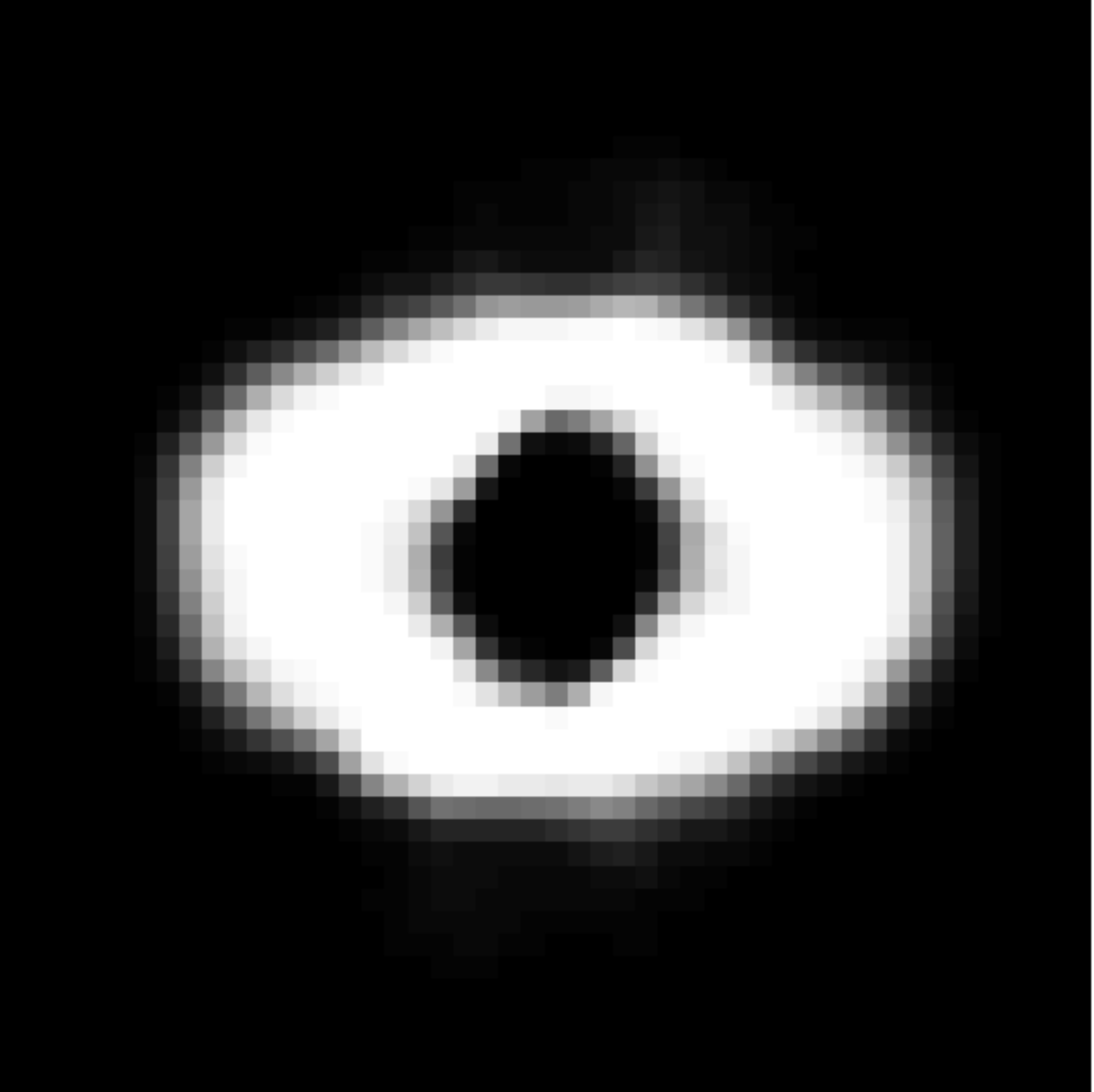} &
\includegraphics[width=0.75cm, height = 0.75cm]{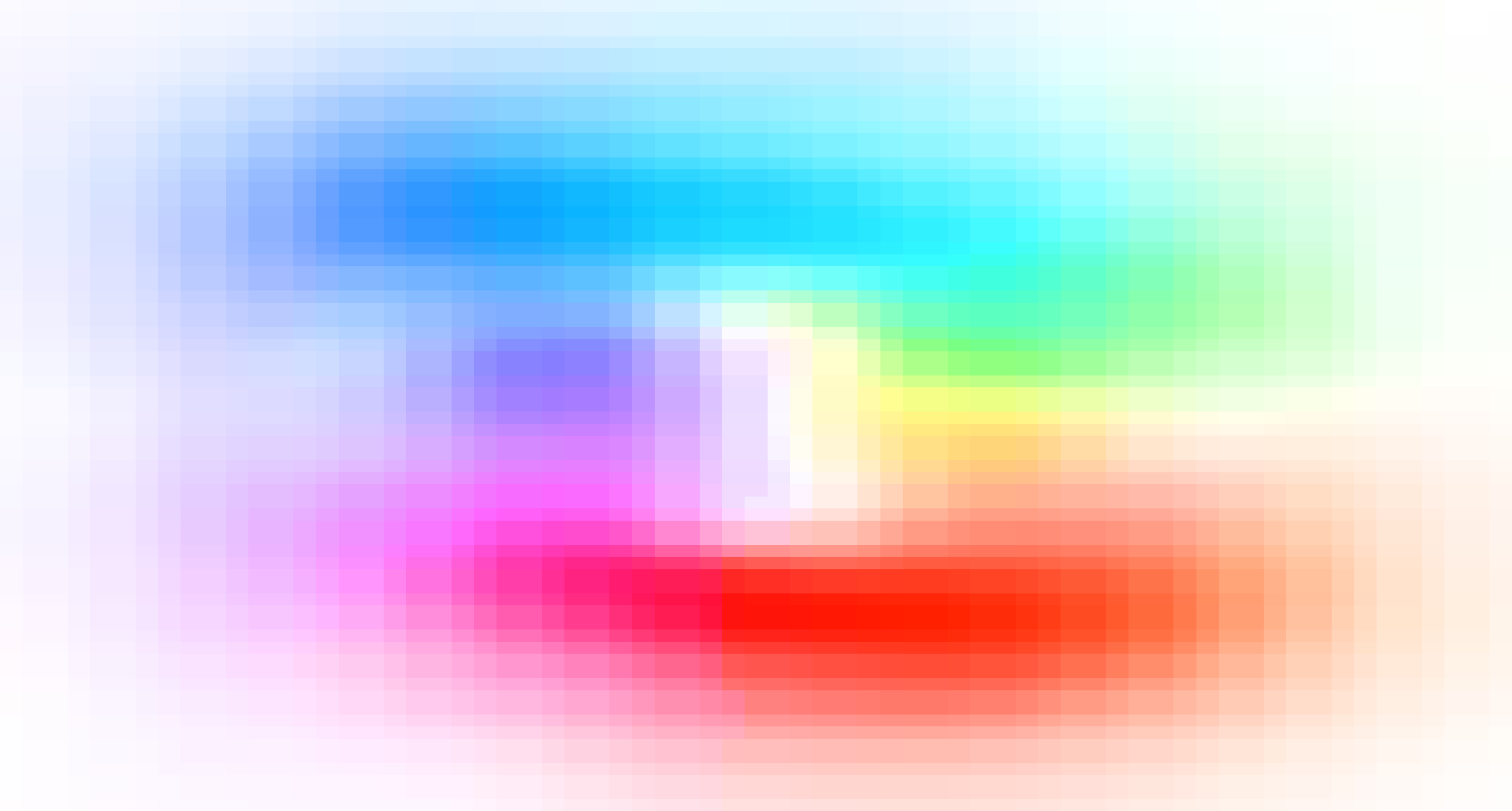} &
\includegraphics[width=0.065\textwidth]{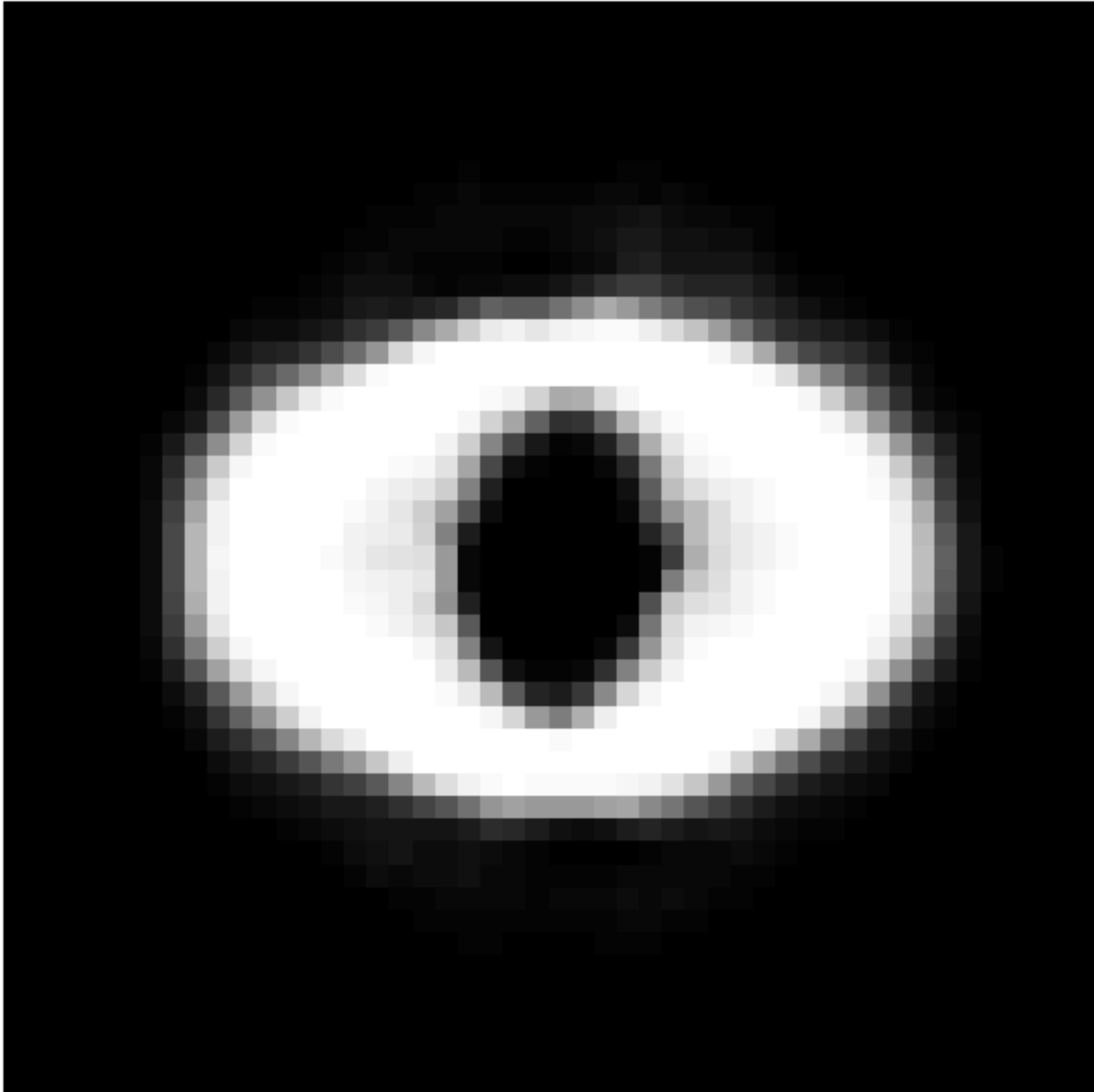} &
\includegraphics[width=0.75cm, height = 0.75cm]{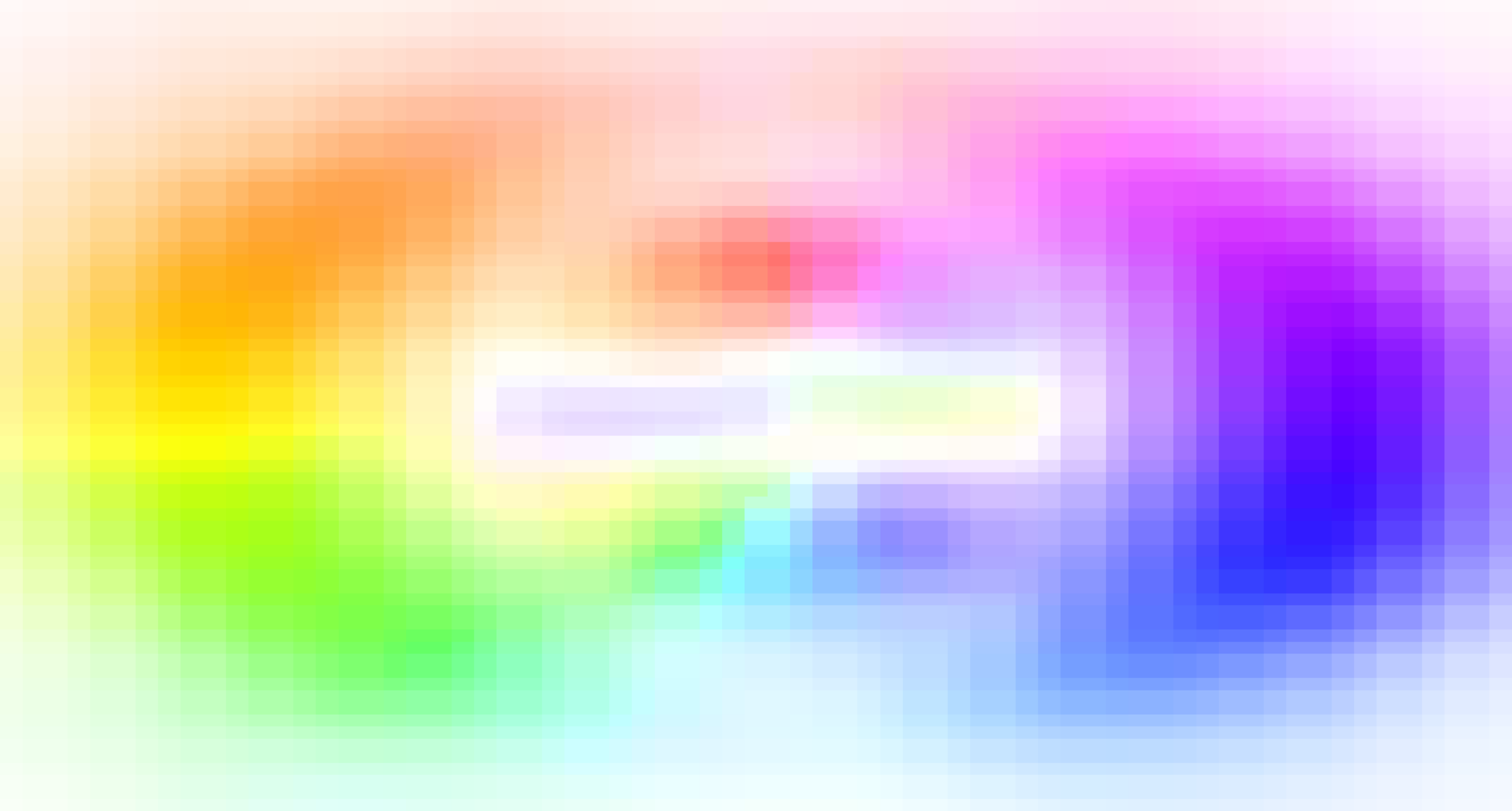} 
\\ 
\includegraphics[width=0.065\textwidth]{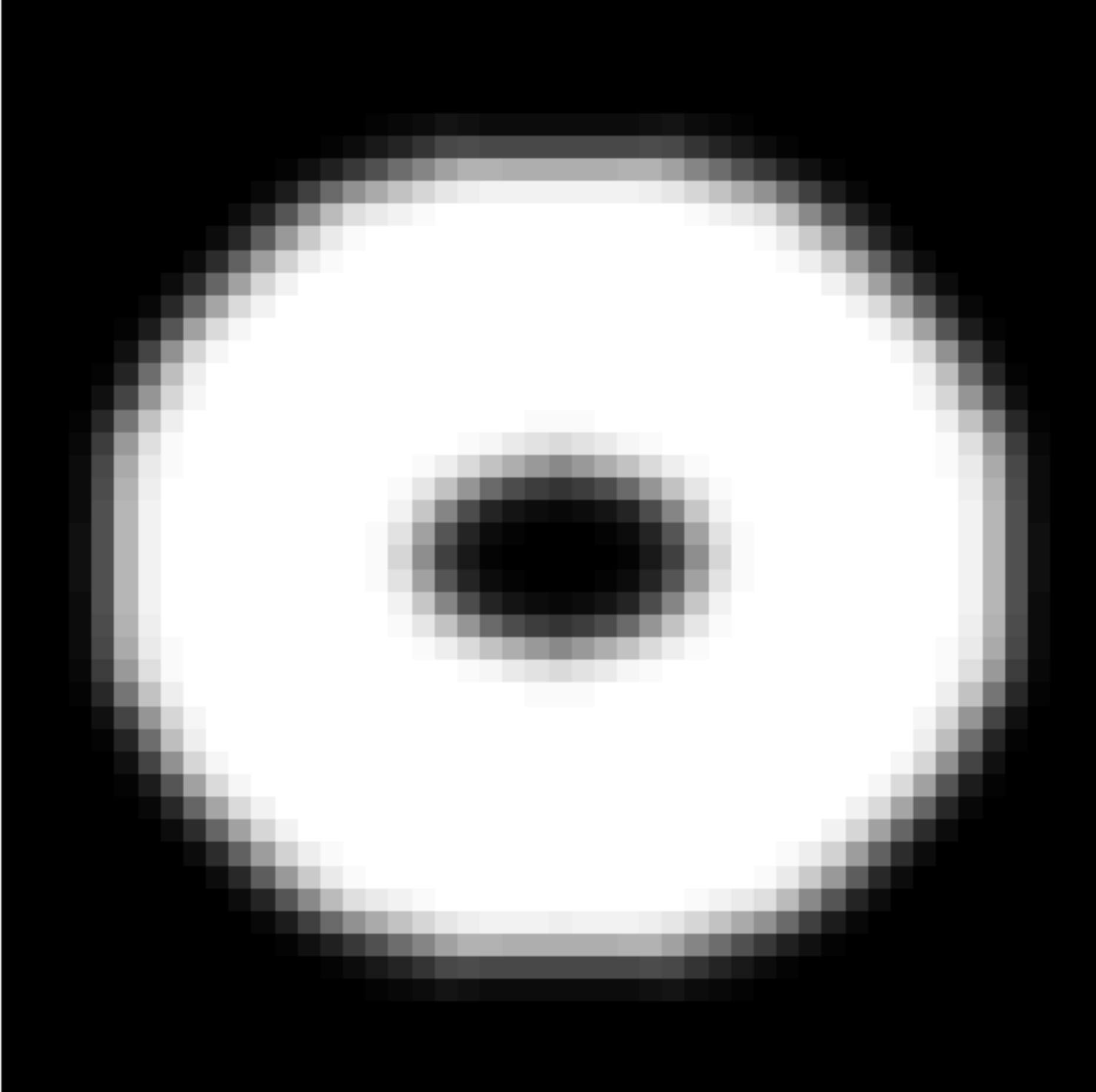} & 
\includegraphics[width=0.75cm, height = 0.75cm]{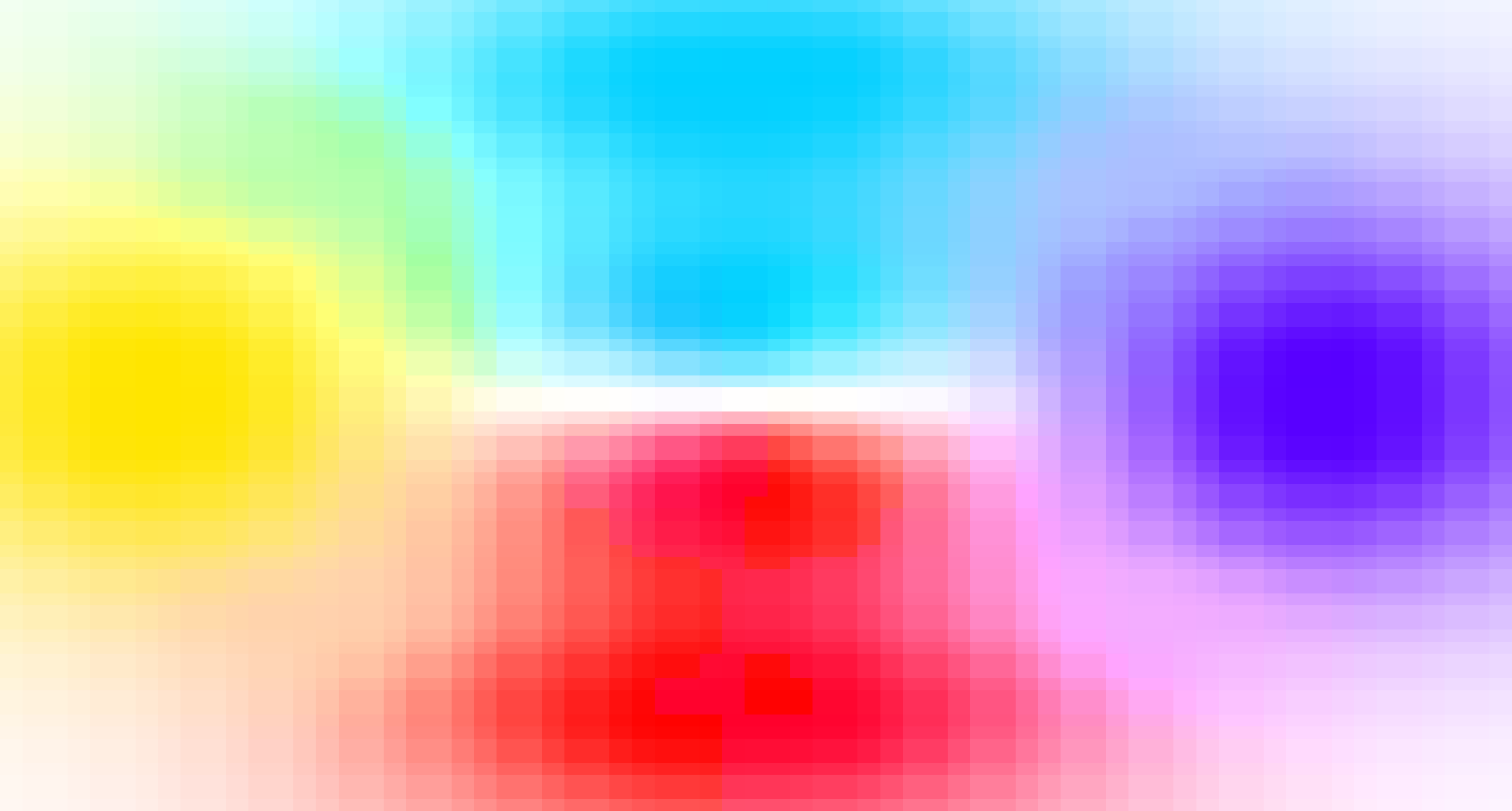} & 
\includegraphics[width=0.065\textwidth]{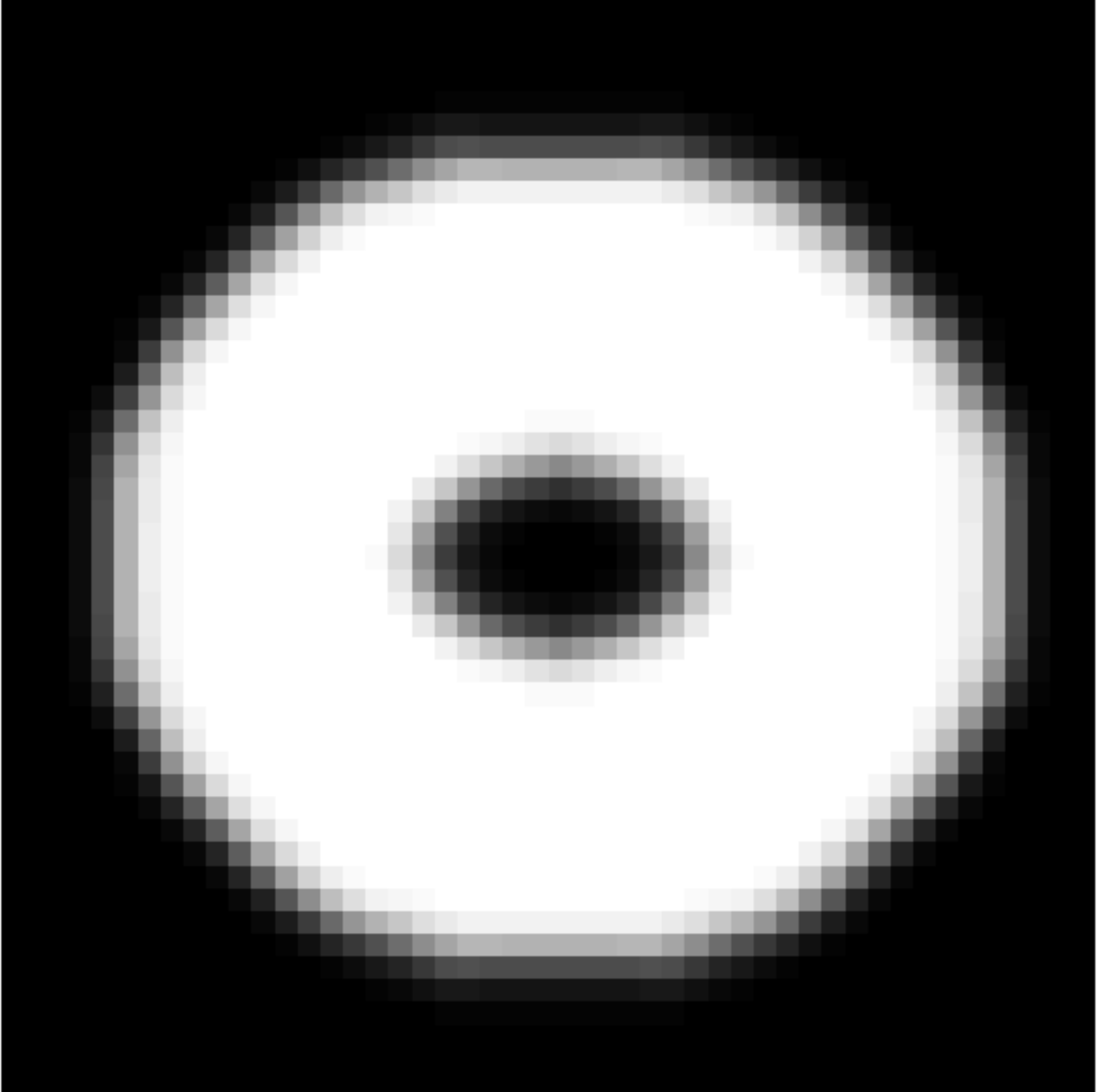} & 
\includegraphics[width=0.75cm, height = 0.75cm]{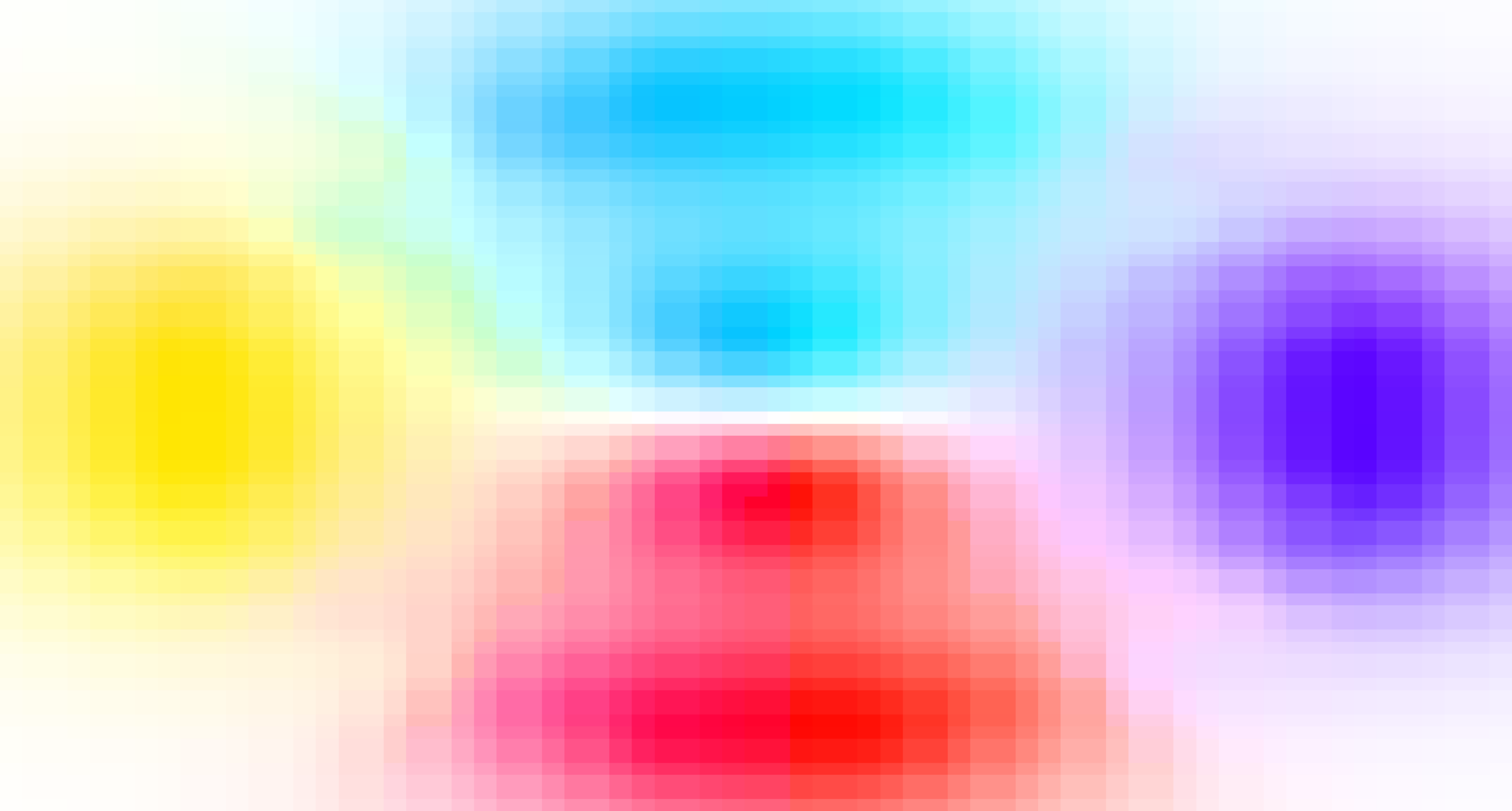} & 
\includegraphics[width=0.065\textwidth]{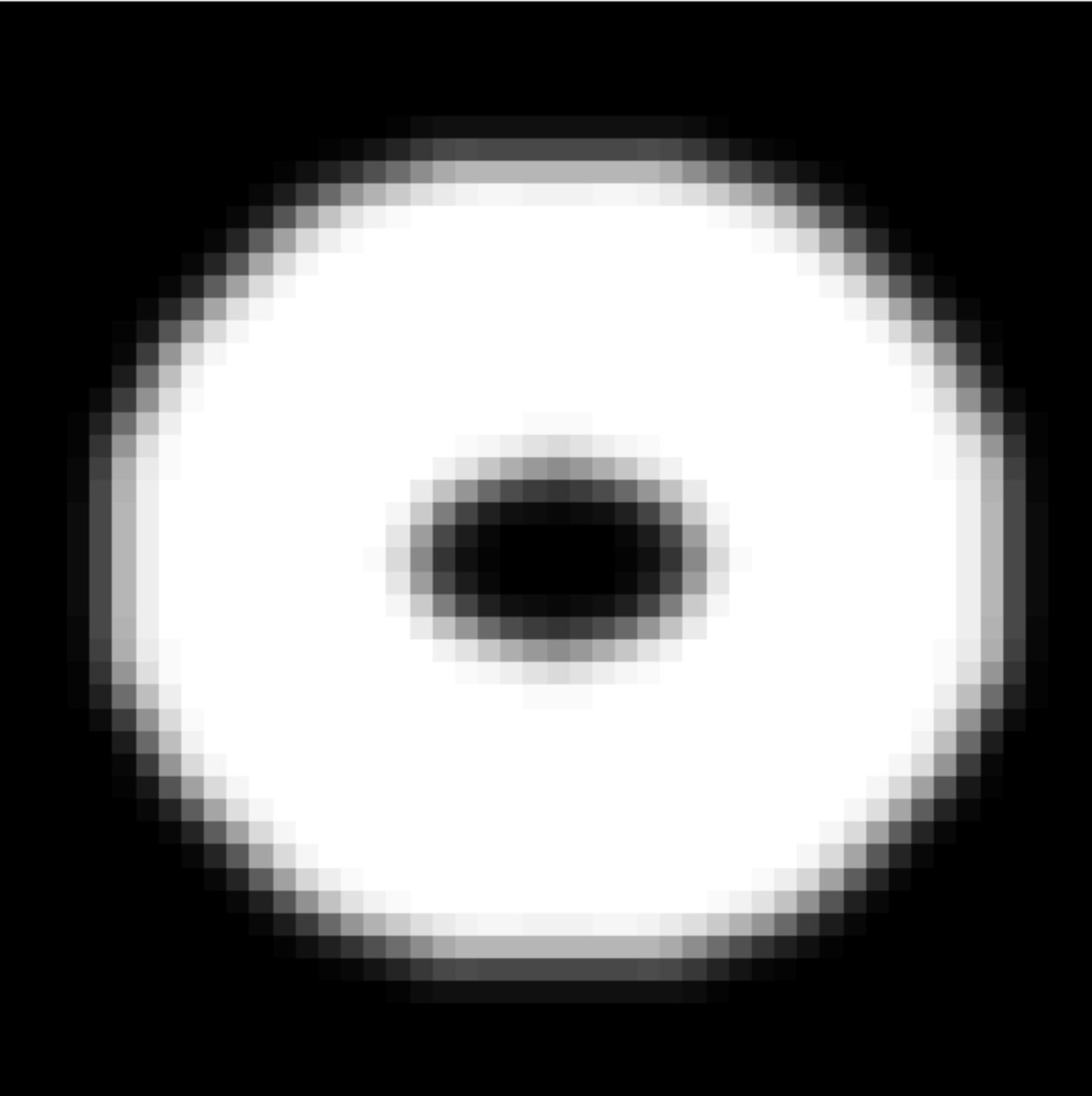} & 
\includegraphics[width=0.75cm, height = 0.75cm]{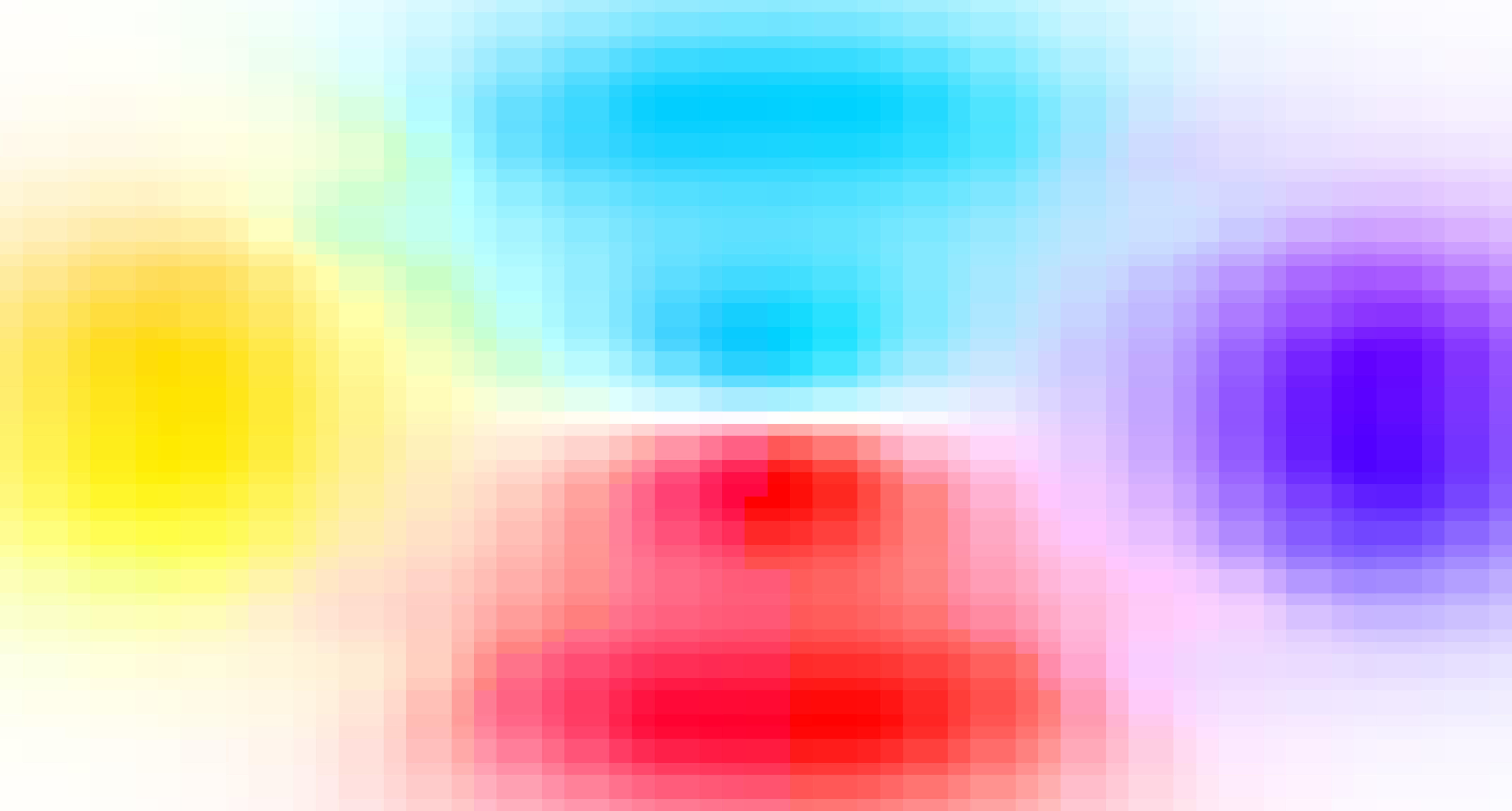} & 
\includegraphics[width=0.065\textwidth]{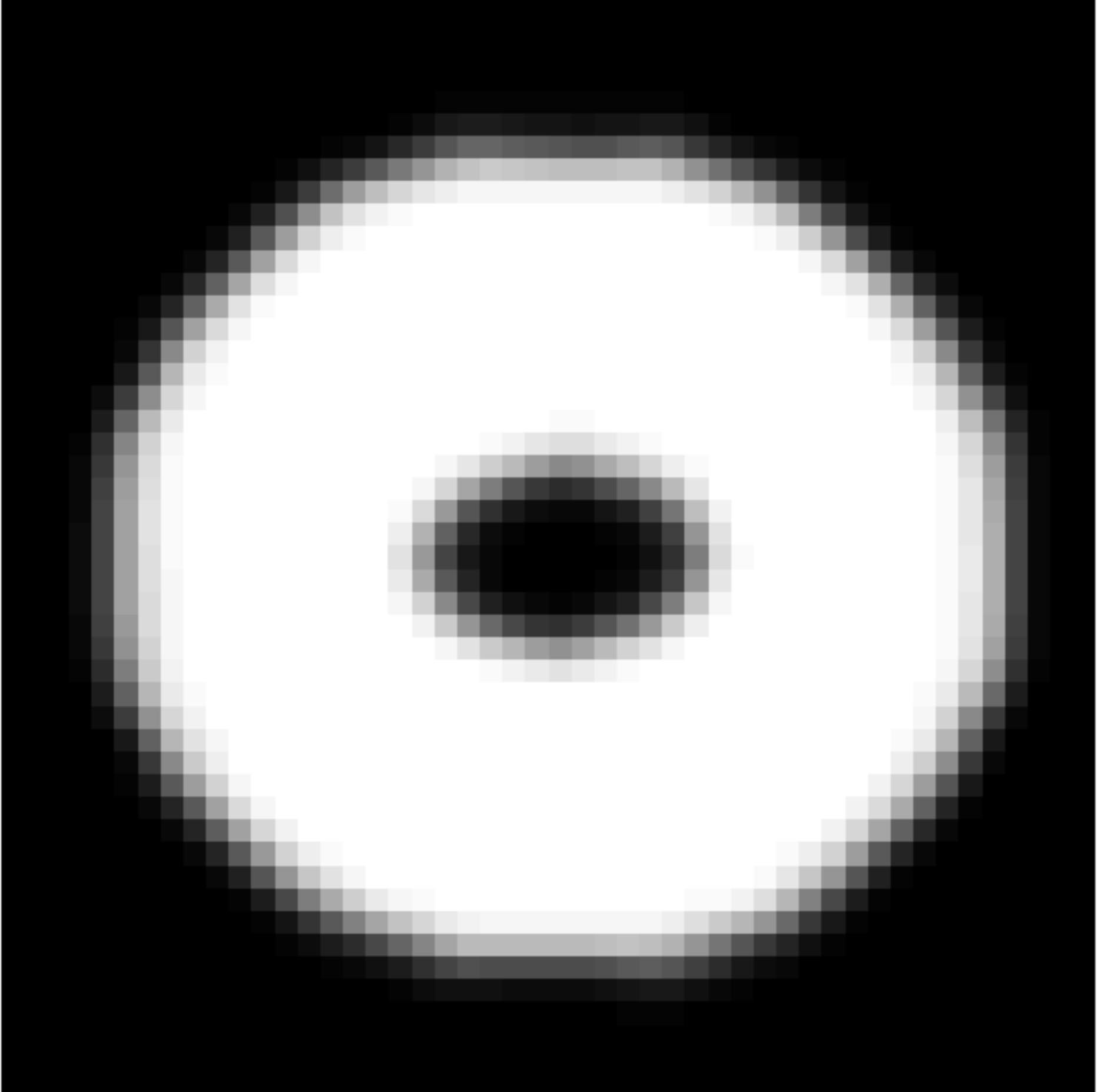} &
\includegraphics[width=0.75cm, height = 0.75cm]{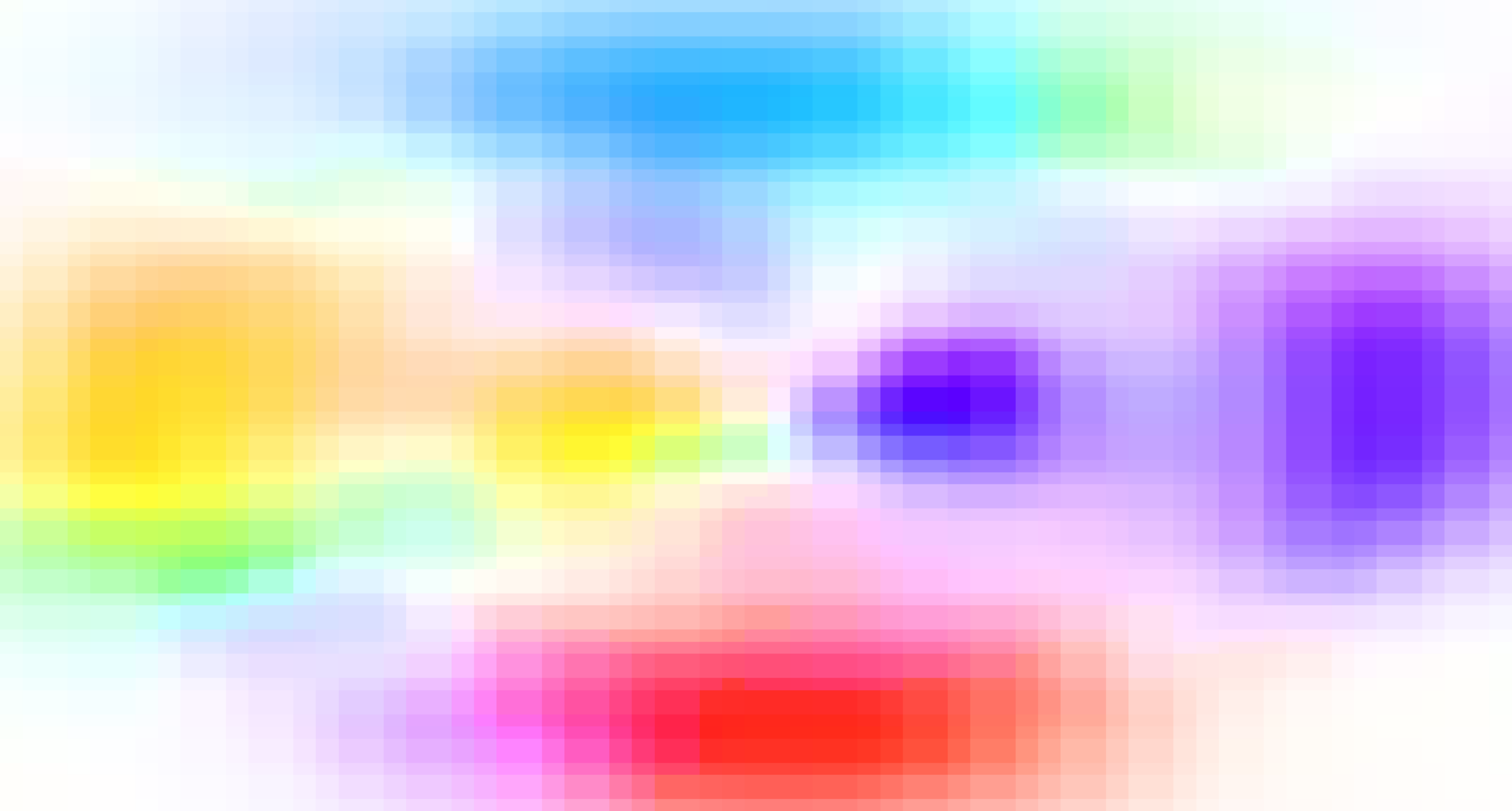} &
\includegraphics[width=0.065\textwidth]{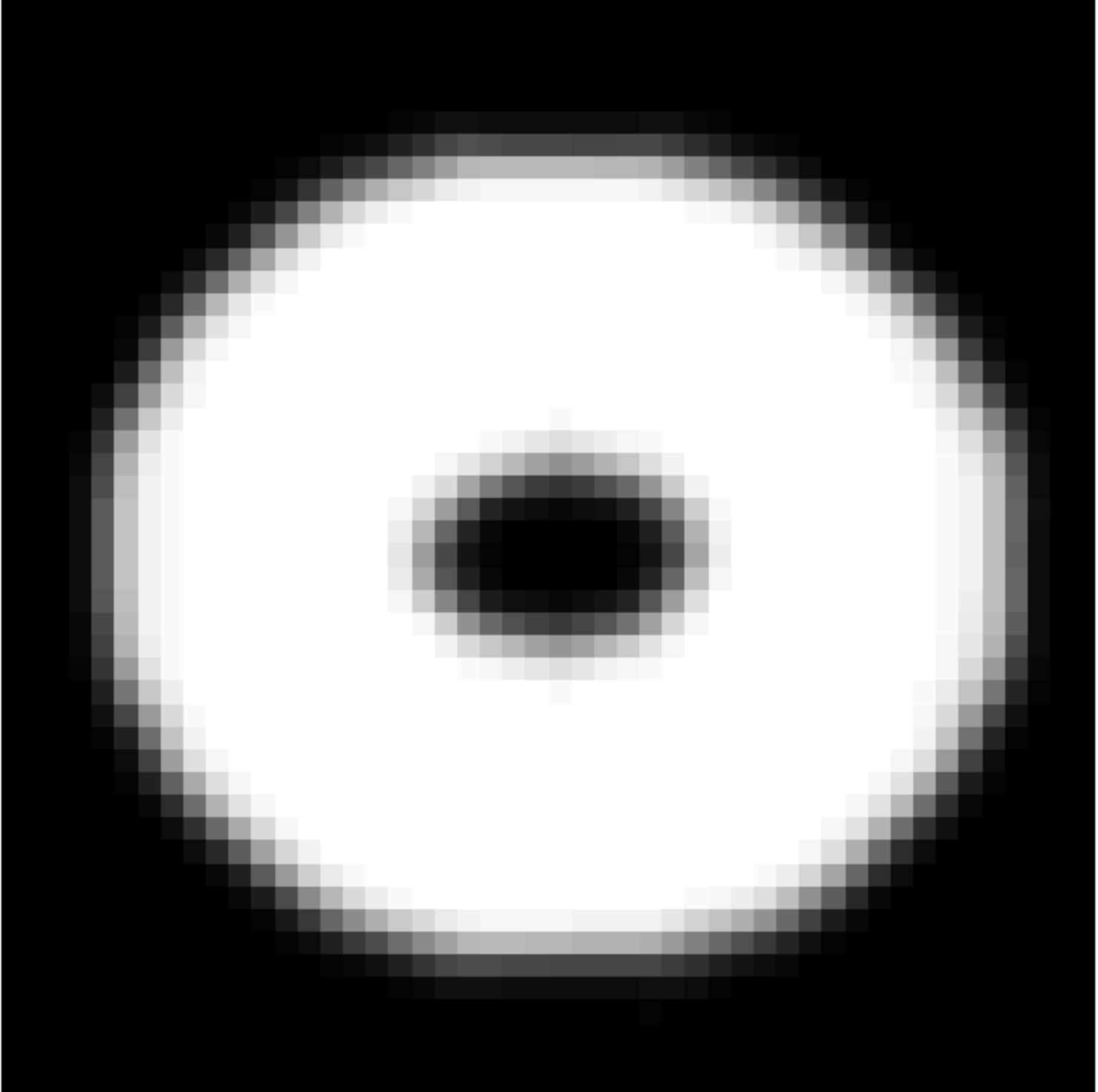} &
\includegraphics[width=0.75cm, height = 0.75cm]{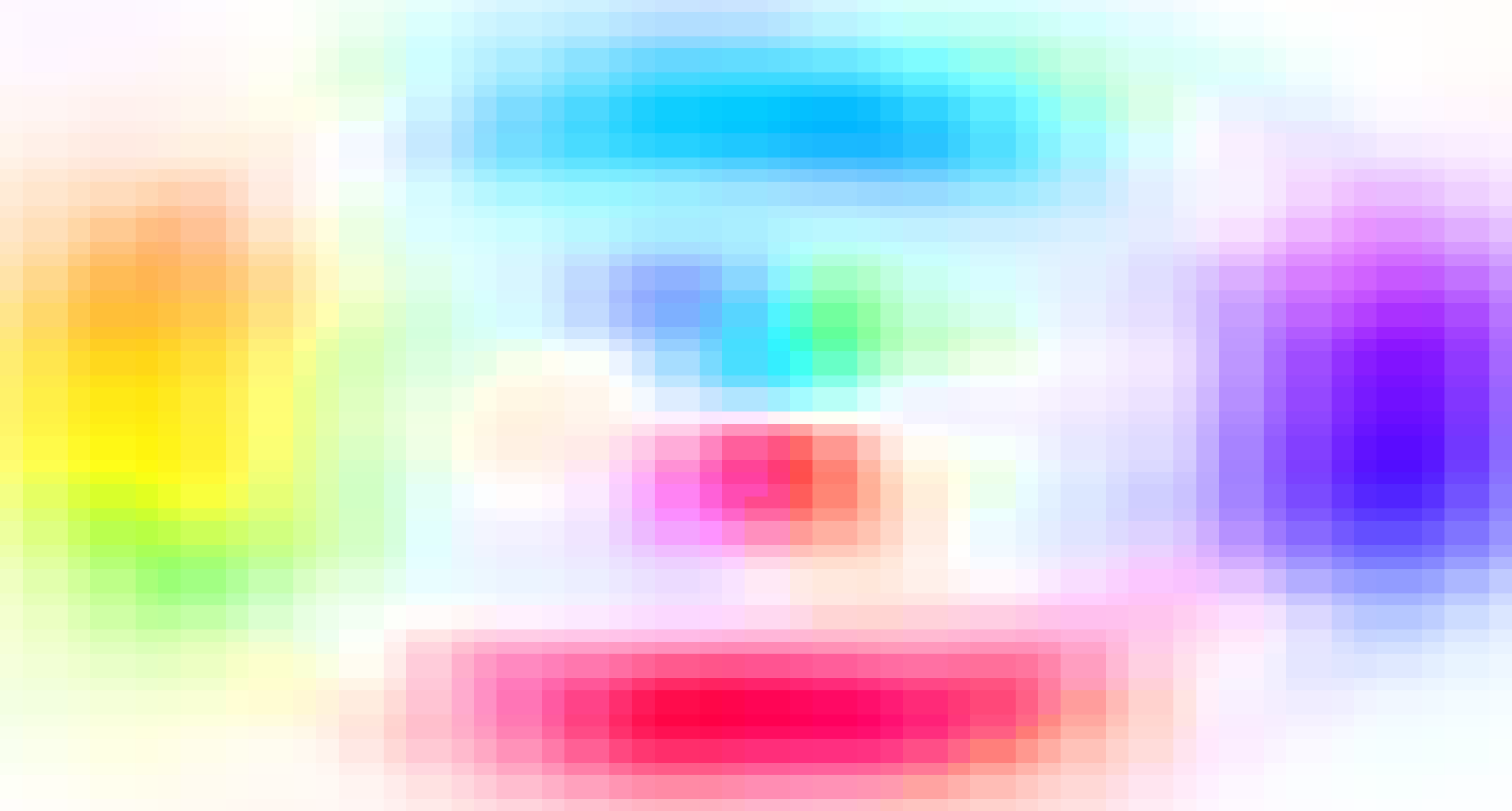} 
\\ 
\includegraphics[width=0.065\textwidth]{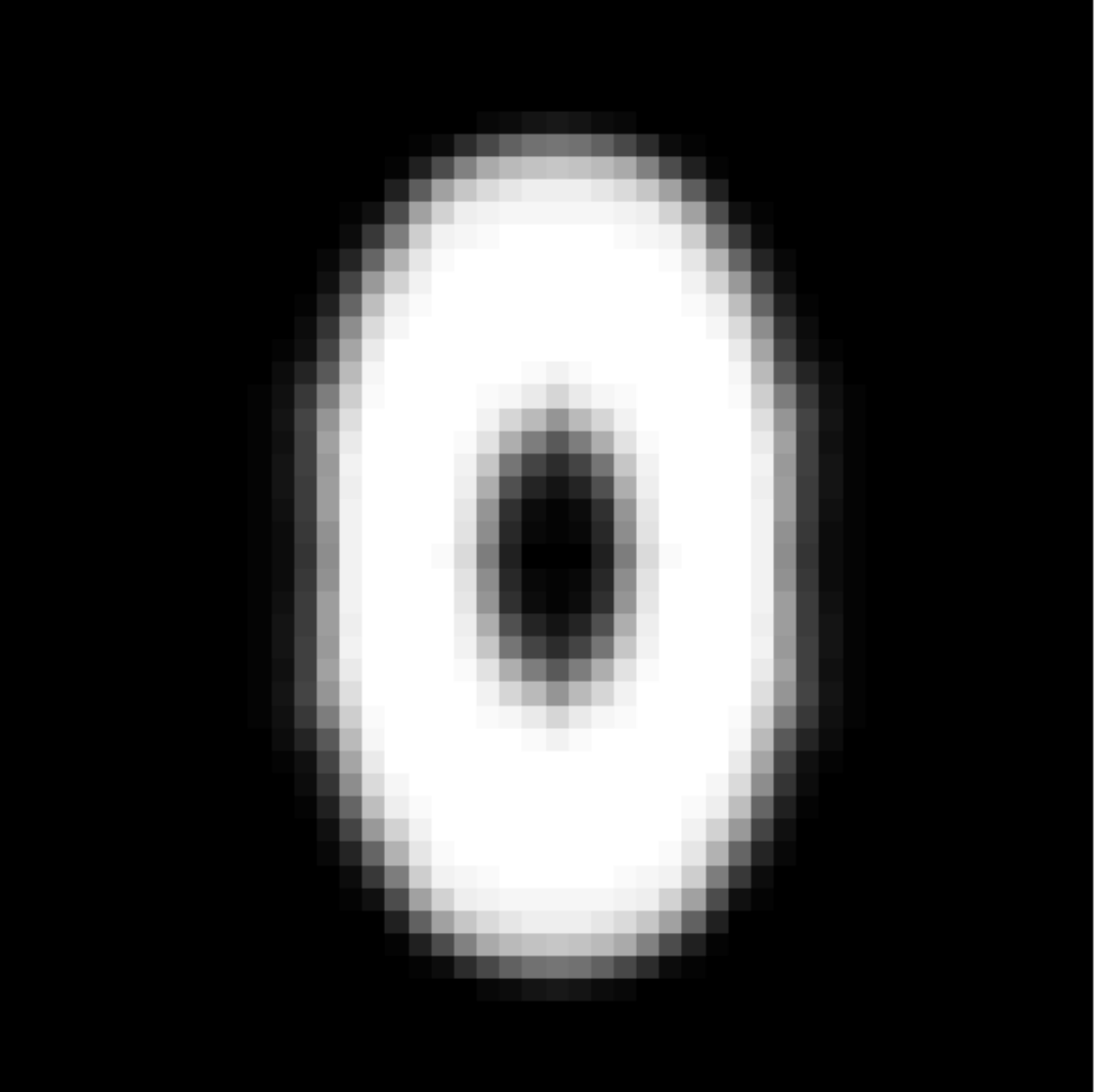} & 
\includegraphics[width=0.75cm, height = 0.75cm]{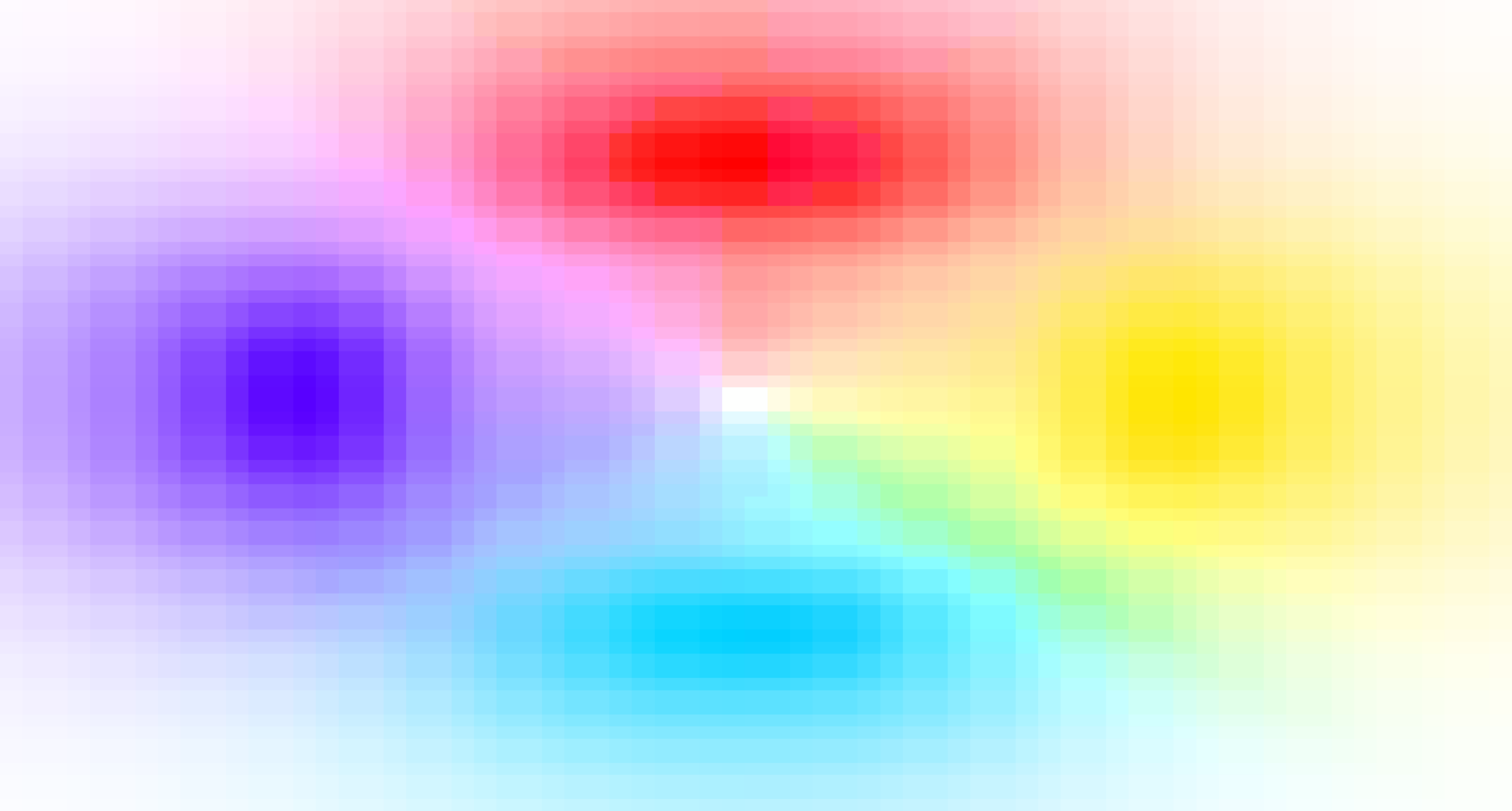} & 
\includegraphics[width=0.065\textwidth]{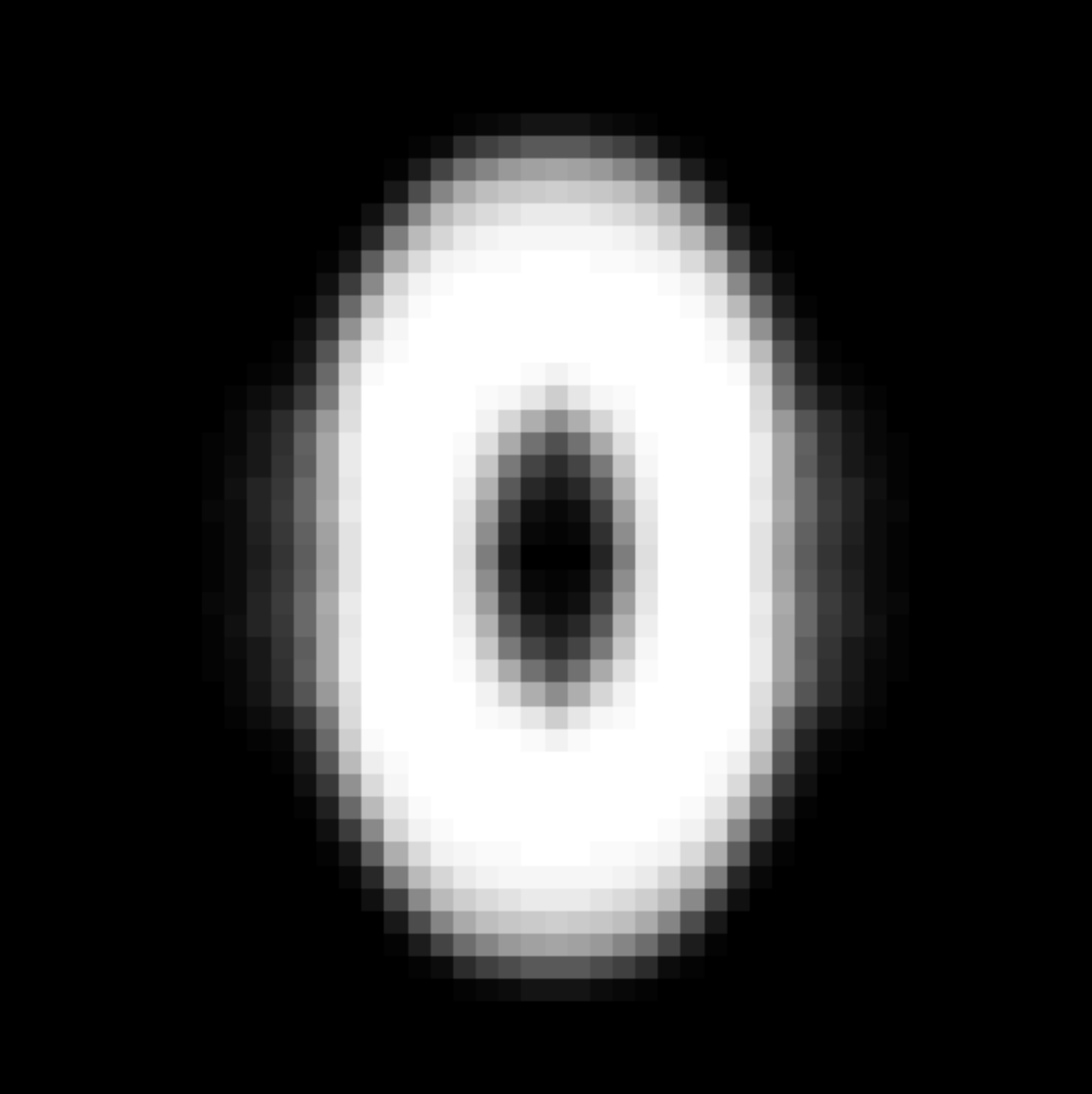} & 
\includegraphics[width=0.75cm, height = 0.75cm]{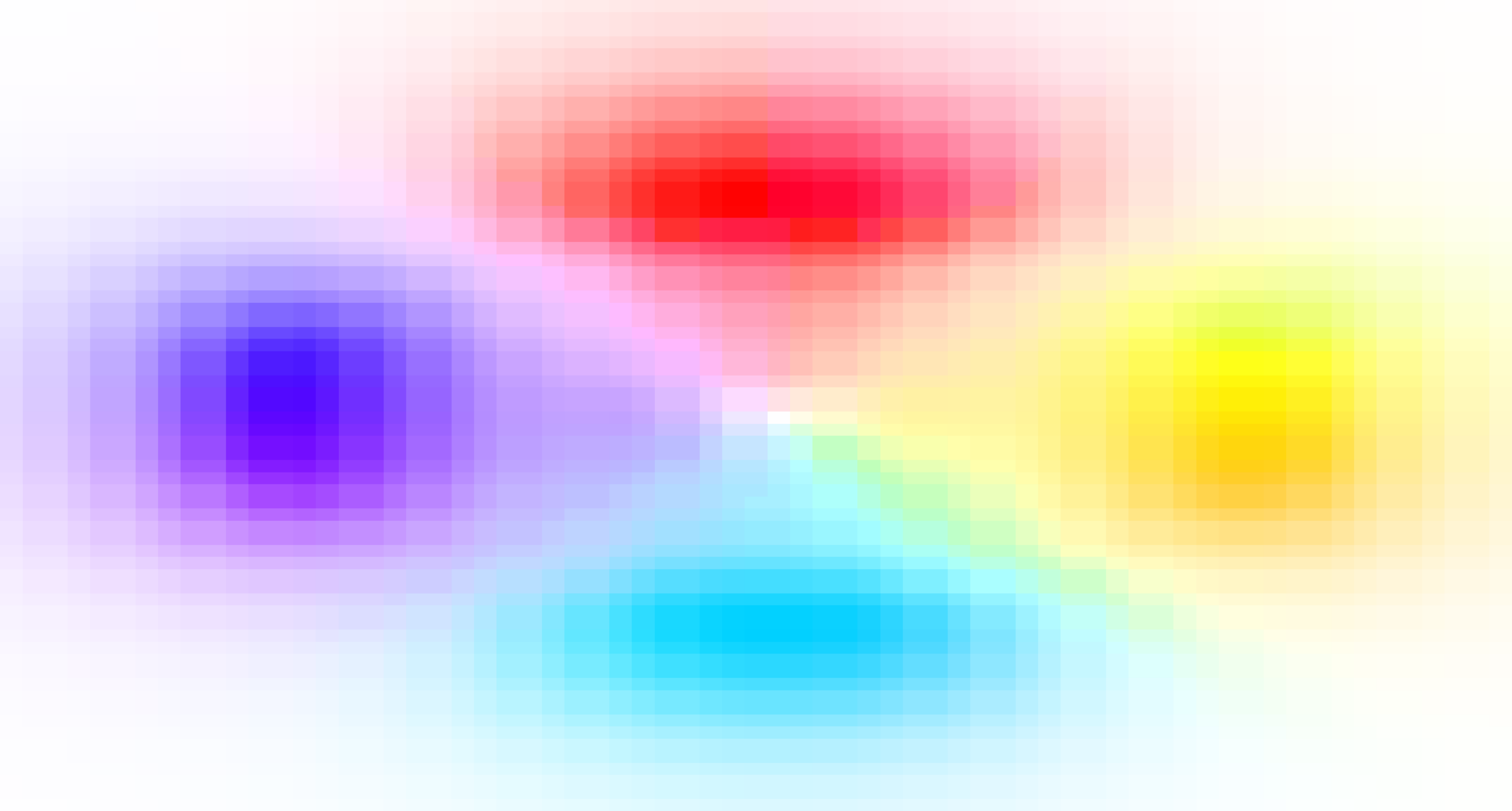} & 
\includegraphics[width=0.065\textwidth]{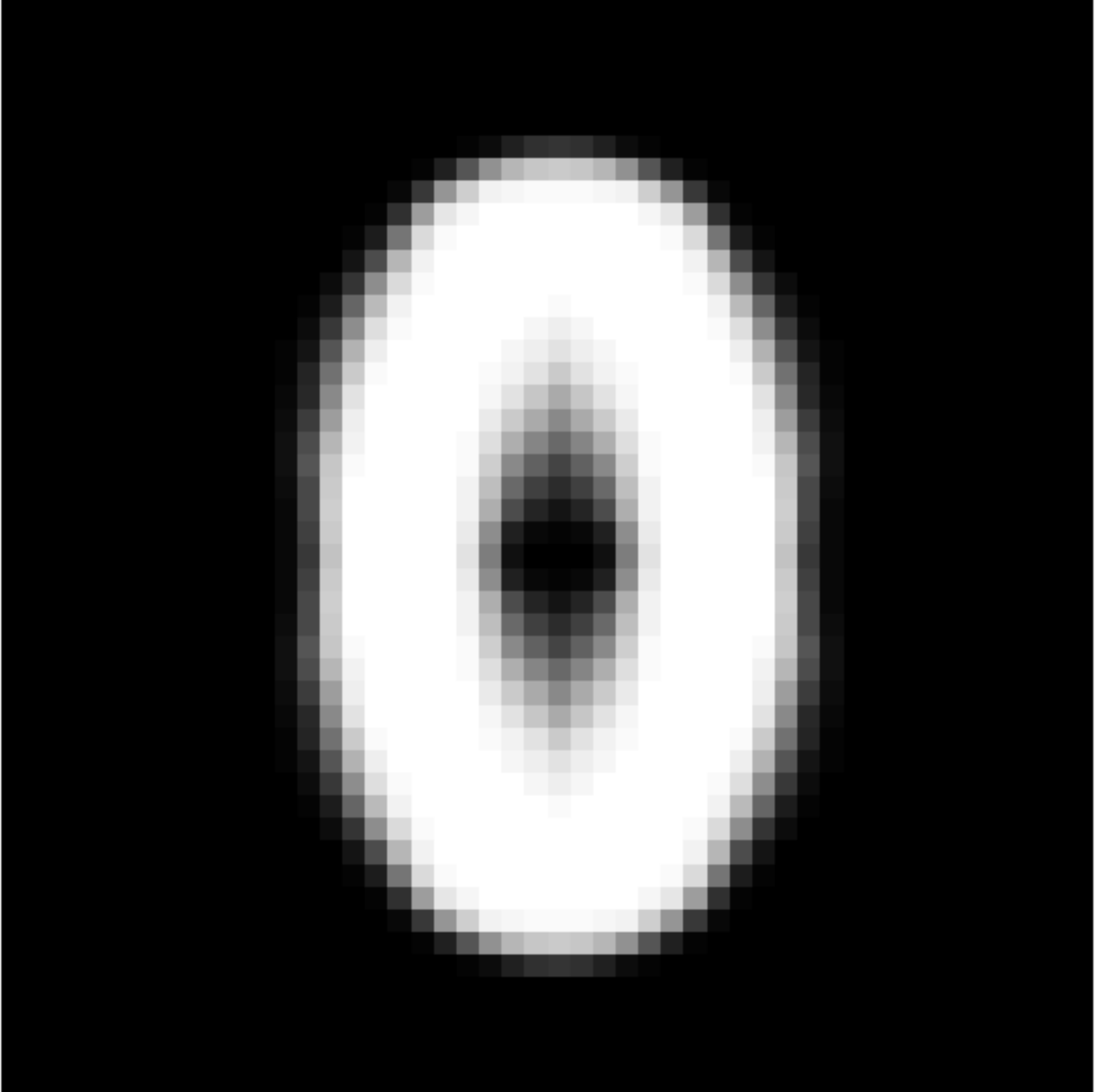} & 
\includegraphics[width=0.75cm, height = 0.75cm]{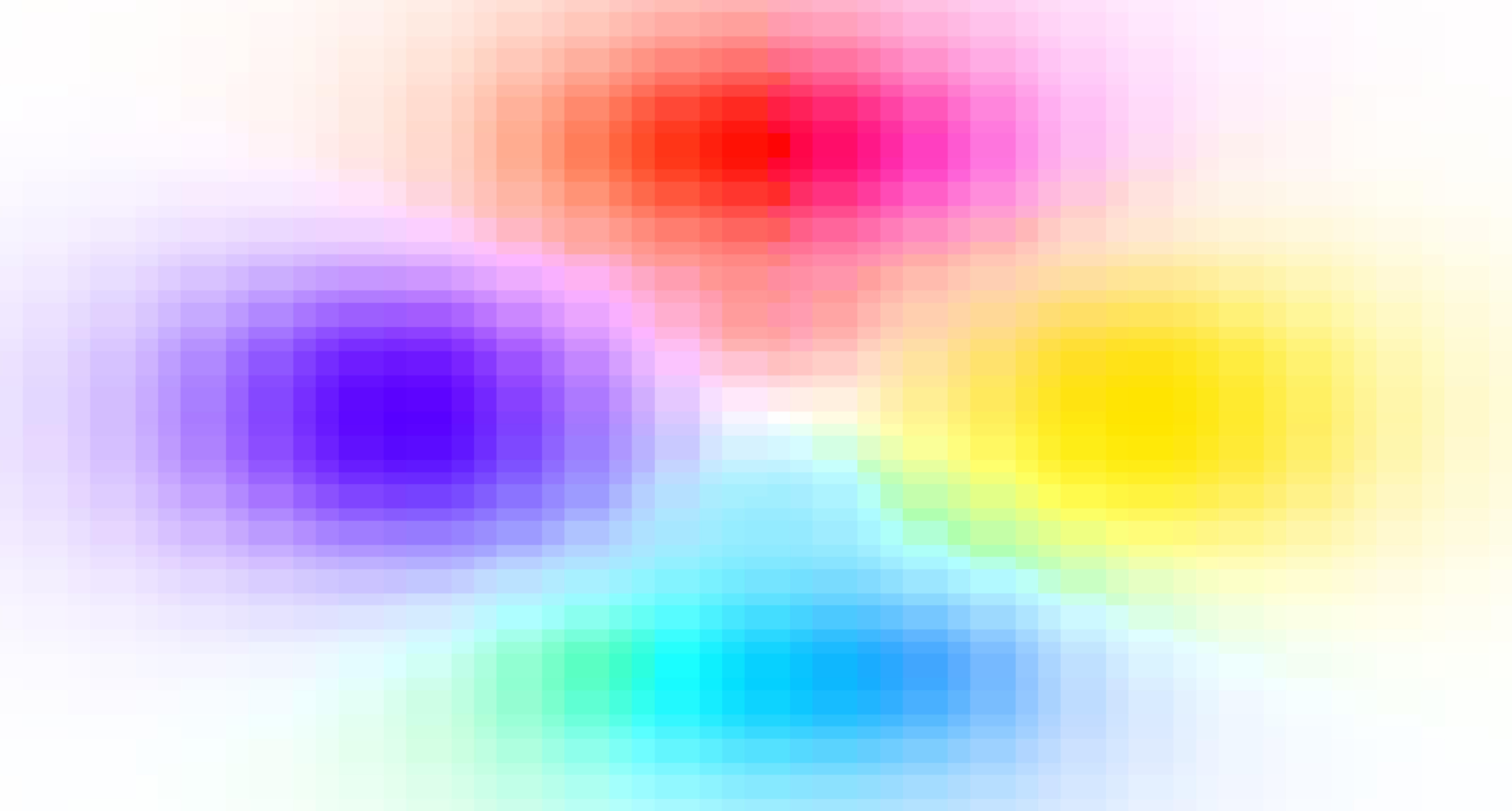} & 
\includegraphics[width=0.065\textwidth]{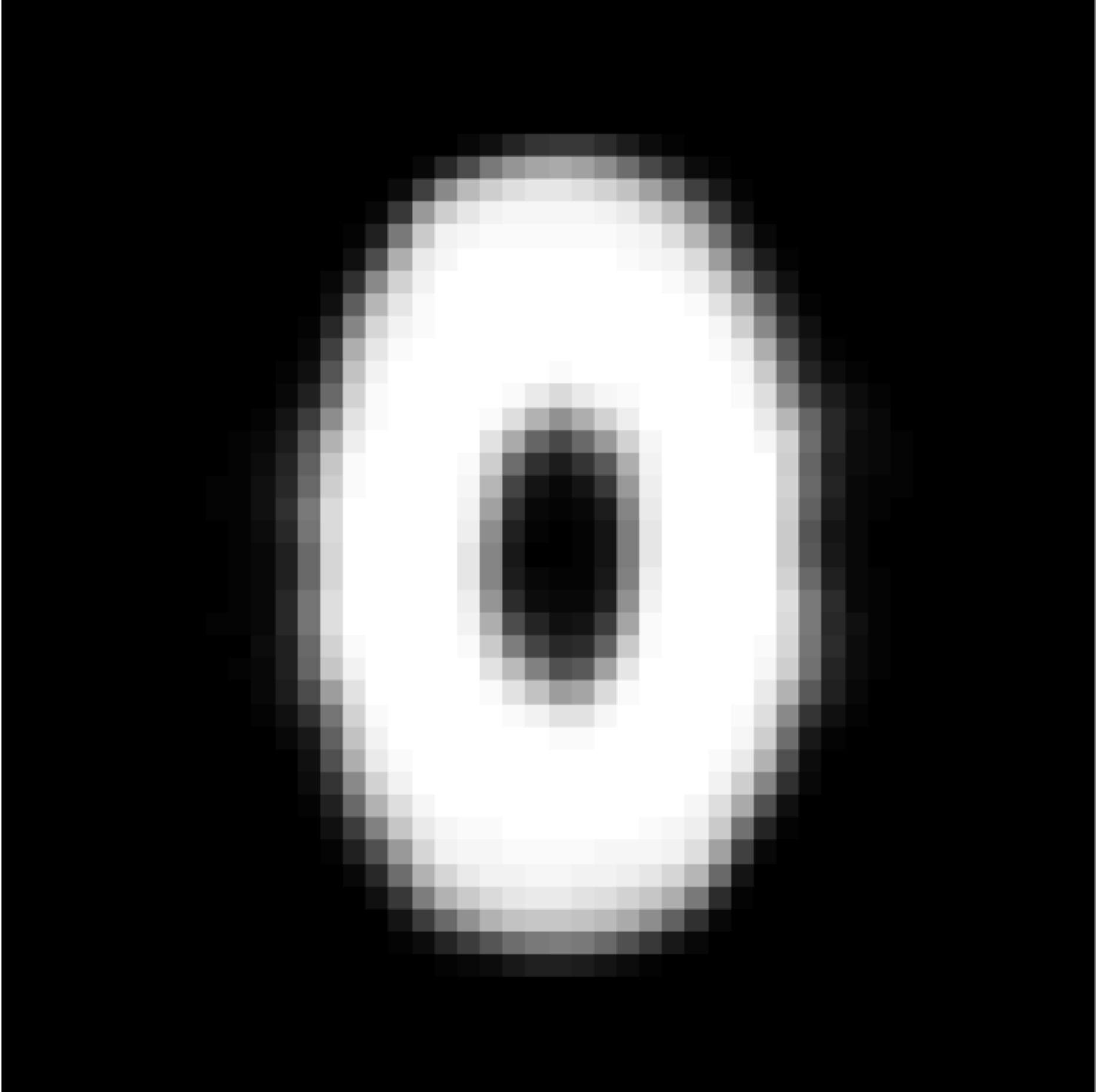} &
\includegraphics[width=0.75cm, height = 0.75cm]{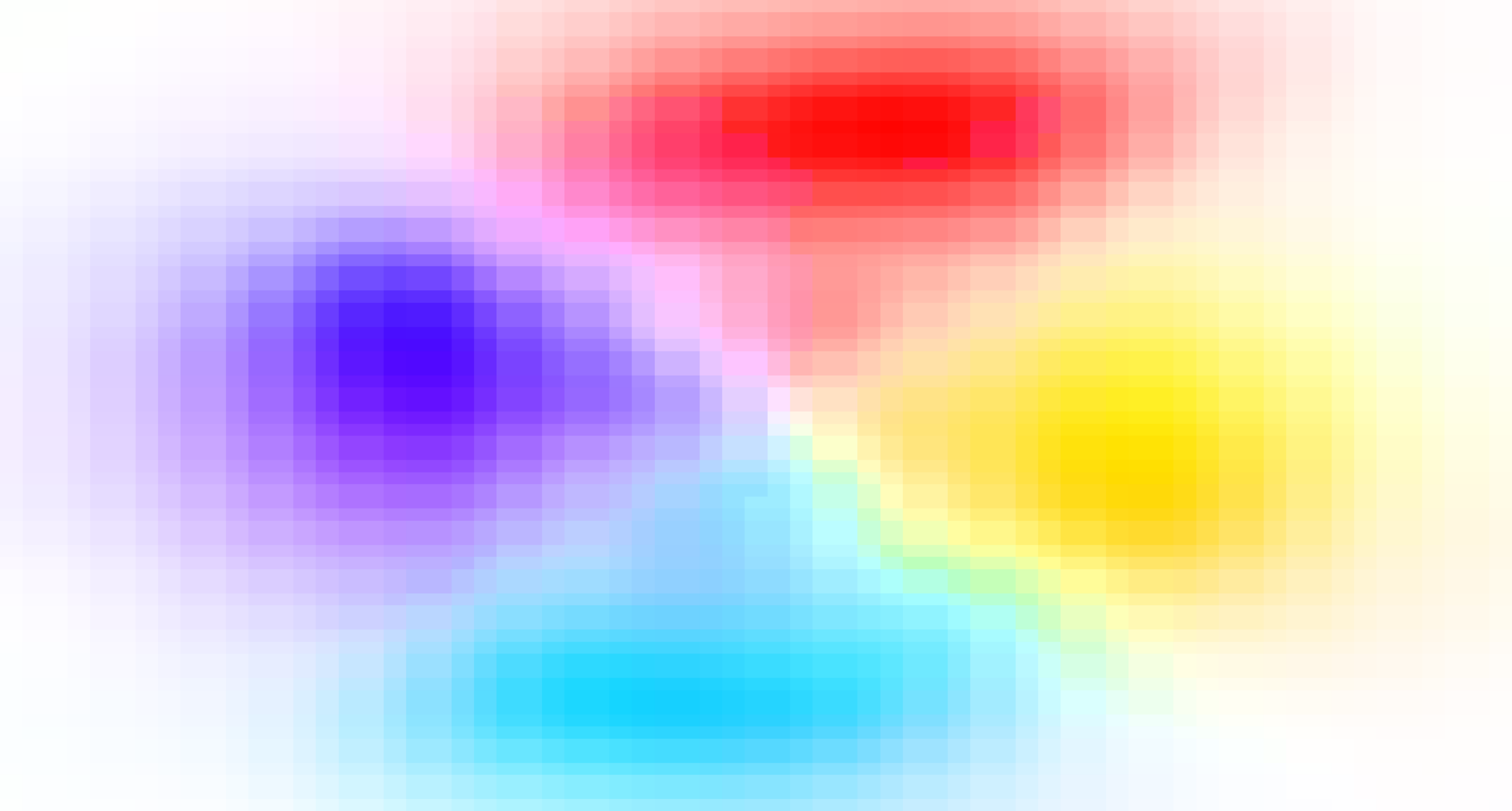} &
\includegraphics[width=0.065\textwidth]{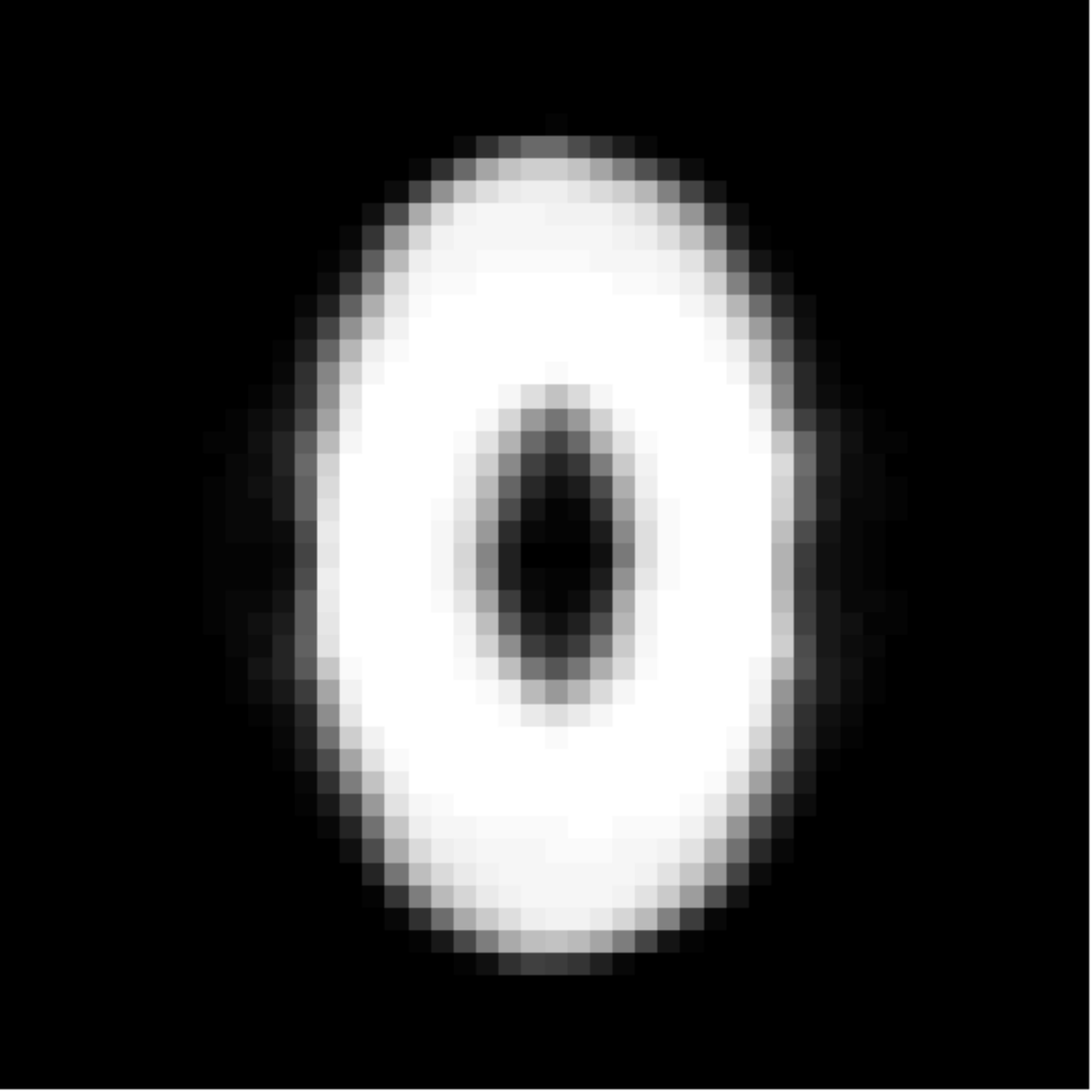} &
\includegraphics[width=0.75cm, height = 0.75cm]{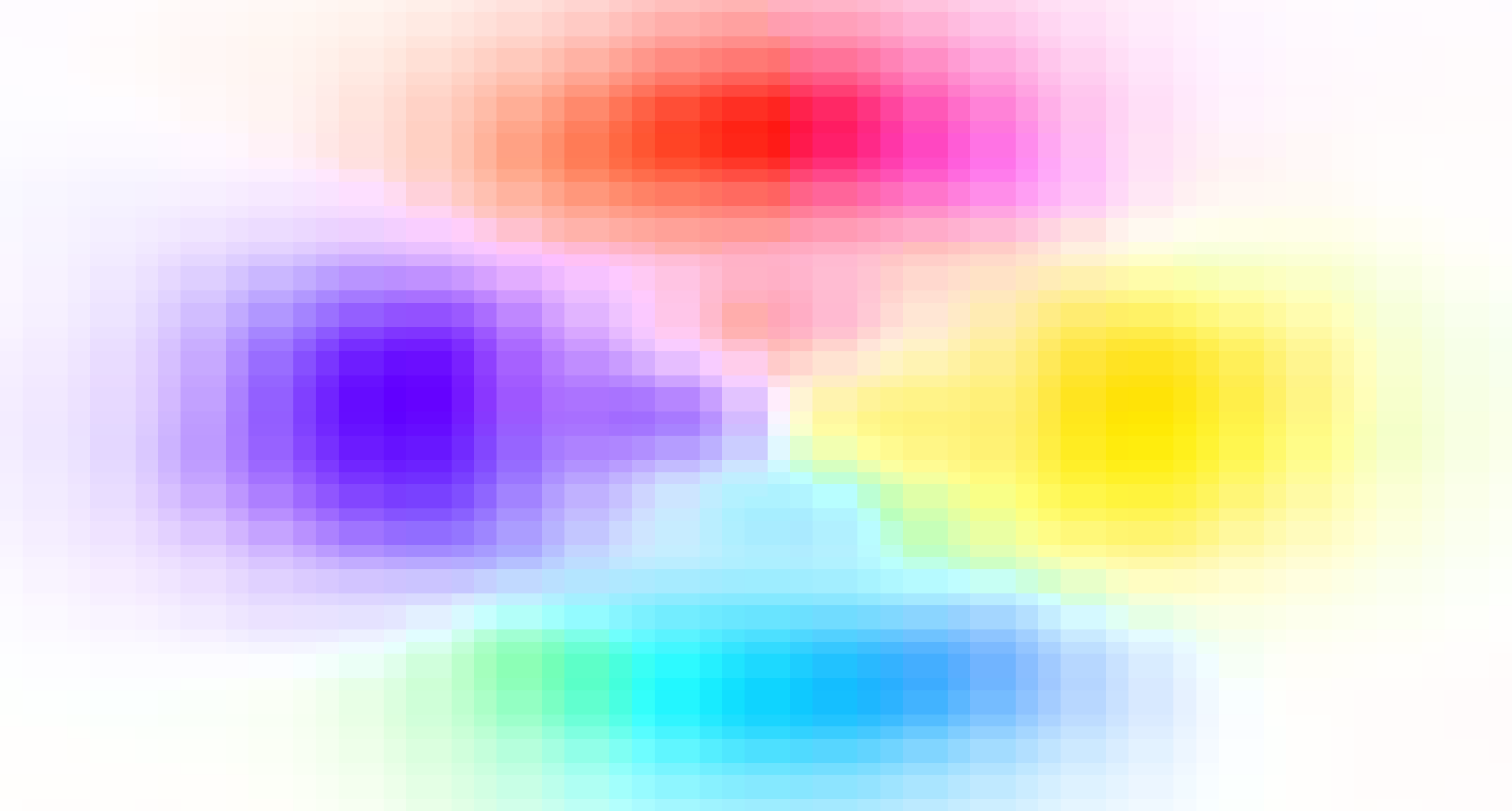} 
\\ 
\end{tabular}
\caption{\small Example of simulated 2D registration results. Up: source and target images of five selected experiments.
Down, left to right: deformed images and velocity fields computed from diffeomorphic Demons (DD), stationary LDDMM (St. LDDMM), 
Flash, and our proposed SVF-GAN and EPDiff-GAN.
SVF stands for a stationary velocity field and $V_0$ for the initial velocity field of a geodesic shooting approach, respectively.
}
\label{fig:2D}
\end{figure}

Figure~\ref{fig:2D} shows the deformed images and the velocity fields obtained in the 2D simulated dataset
by diffeomorphic Demons~\cite{Vercauteren_09}, a stationary version of LDDMM (St. LDDMM)~\cite{Hernandez_14}, 
the spatial version of Flash~\cite{Zhang_18}, and our proposed SVF and EPDiff GANs.
Apart from diffeomorphic Demons that uses Gaussian smoothing for regularization, all the considered methods 
use the same parameters for operator $L$. Therefore, St. LDDMM and SVF-GAN can be seen as a model-based and
a data-based approach for the minimization of the same variational problem. 
The same happens with Flash and EPDiff-GAN.

From the figure, it can be appreciated that our proposed GANs are able to obtain accurate warps of the source
to the target images, similarly to model-based approaches. For SVF-GAN, the inferred velocity fields 
are visually similar to model-based approaches in three of five experiments. For EPDiff-GAN, the inferred initial
velocity fields are visually similar to model-based approaches in four of five experiments.

\subsection{Results in the 3D NIREP dataset}

\begin{figure}[!t]
\centering
\begin{tabular}{cc}
\hspace{-0.25cm}
\includegraphics[width=0.4\textwidth]{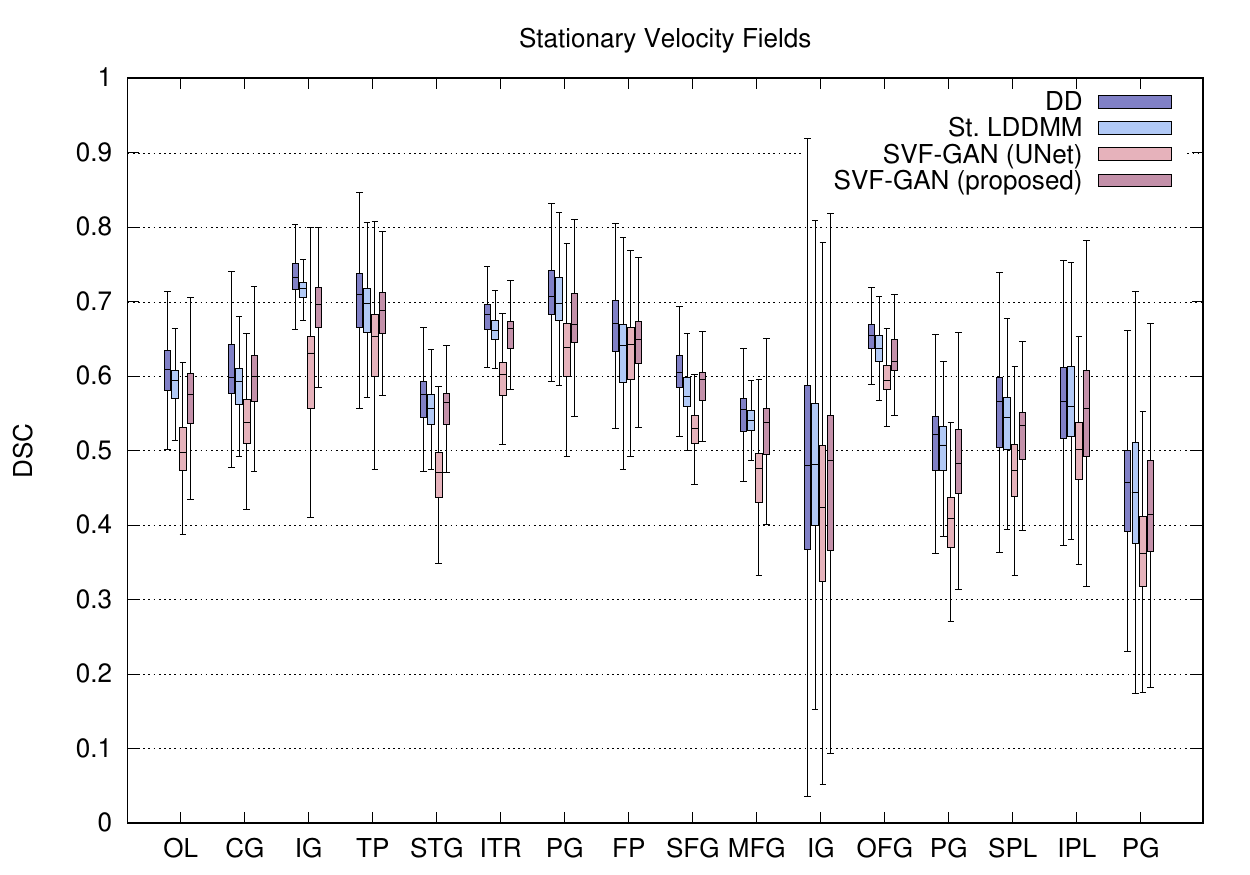} & 
\hspace{-0.25cm}
\includegraphics[width=0.4\textwidth]{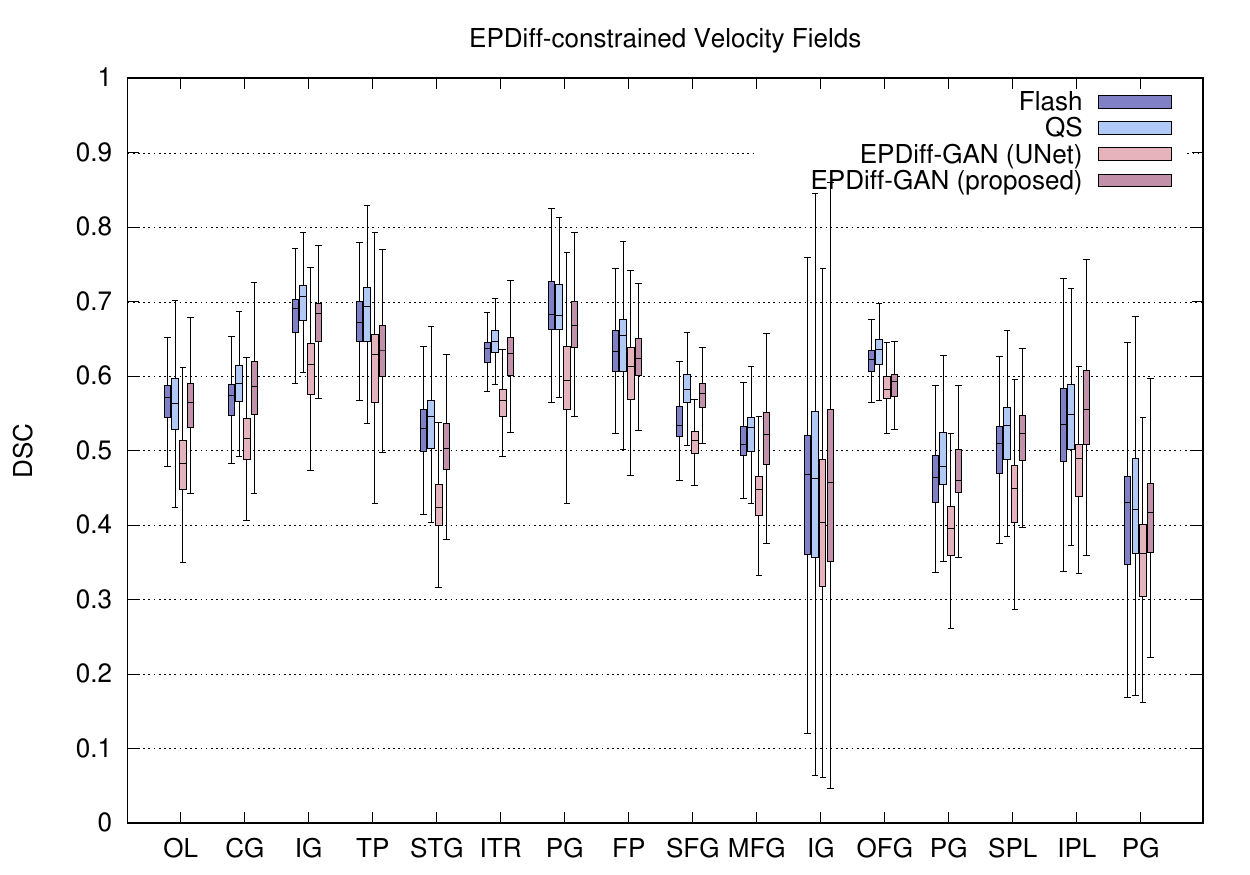} \\
\end{tabular}
\caption{\small Evaluation in NIREP. Dice scores obtained by propagating the diffeomorphisms to the segmentation labels on the
16 NIREP brain structures. Left, methods parameterized with stationary velocity fields: diffeomorphic Demons (DD), 
stationary LDDMM (St. LDDMM), Voxelmorph II, our SVF-GAN with U-Net architecture, and our proposed SFV-GAN with the two-stream architecture. 
Right, geodesic shooting methods: Flash, Quicksilver (QS), our EPDiff-GAN with U-Net architecture, and our proposed EPDiff-GAN.}
\label{fig:3D}
\end{figure}

\subsubsection{Quantitative assessment}

Figure~\ref{fig:3D} shows the Dice similarity coefficients obtained with 
diffeomorphic Demons~\cite{Vercauteren_09}, St. LDDMM~\cite{Hernandez_14}, Voxelmorph II~\cite{Balakrishnan_19}, the spatial version of Flash~\cite{Zhang_18}, 
Quicksilver~\cite{Yang_17} and our proposed SVF and EPDiff GANs.
From them, we show the results of our proposed two-stream architecture in contrast to a simpler U-Net based architecture.
The results have been split into stationary and geodesic shooting methods.

From the figure, it can be appreciated that our proposed two-stream architecture greatly improves the accuracy obtained by simple U-Net.
SVF-GAN shows an accuracy similar to St. LDDMM and competitive with diffeomorphic Demons.
Our proposed method tends to overpass Voxelmorph II in the great majority of the structures.
On the other hand, EPDiff-GAN shows an accuracy similar to Flash and Quicksilver in the great majority of regions, with the exception
of the temporal pole (TP) and the orbital frontal gyrus (OFG), two small localized and difficult to register regions.
It drives our attention that Flash underperformed in the superior frontal gyrus (SFG).

\subsubsection{Qualitative assessment}

\begin{figure}[!t]
\centering
\begin{tabular}{cc}
\begin{tabular}{ccc}
\includegraphics[width=0.13\textwidth]{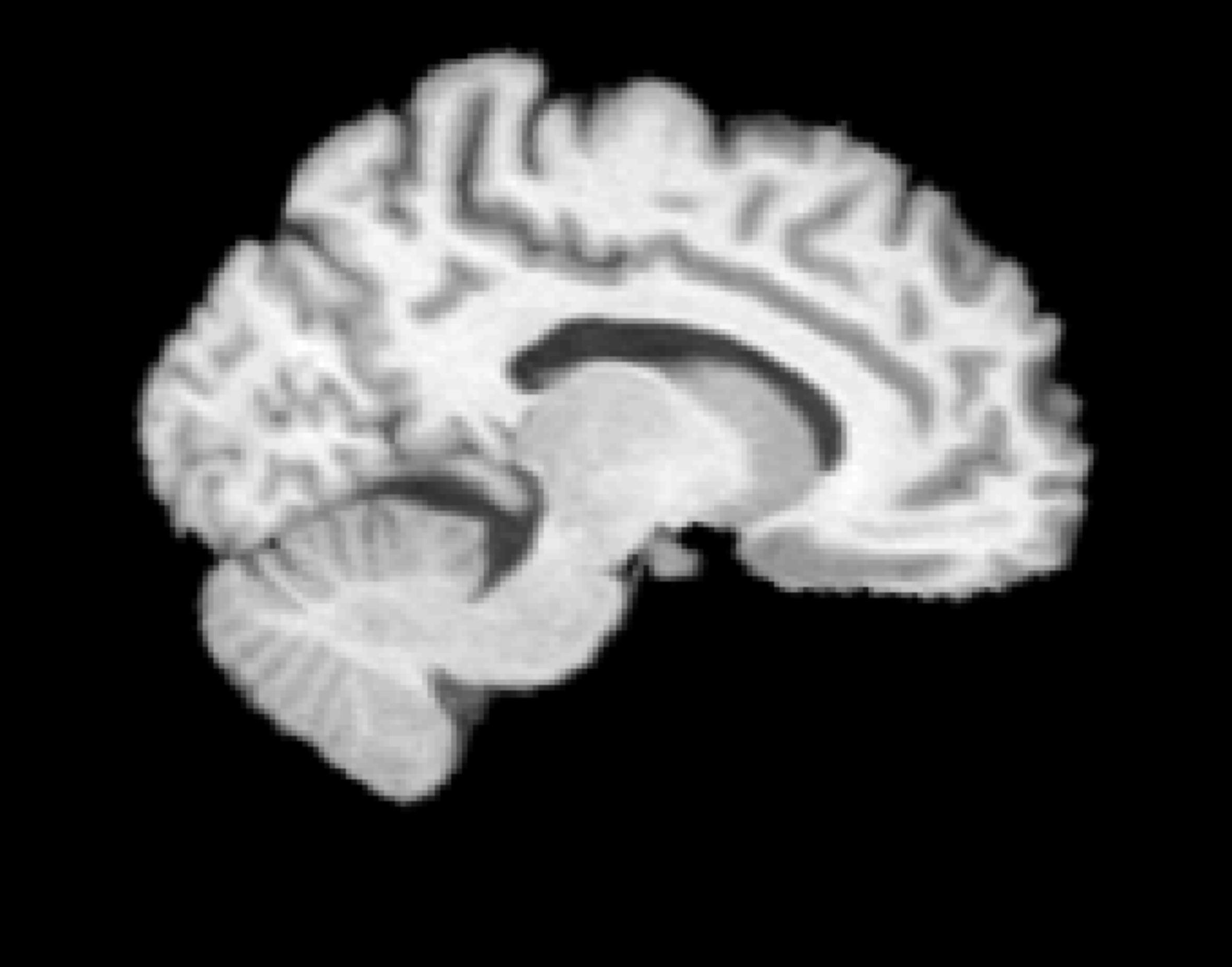} & 
\includegraphics[width=0.13\textwidth]{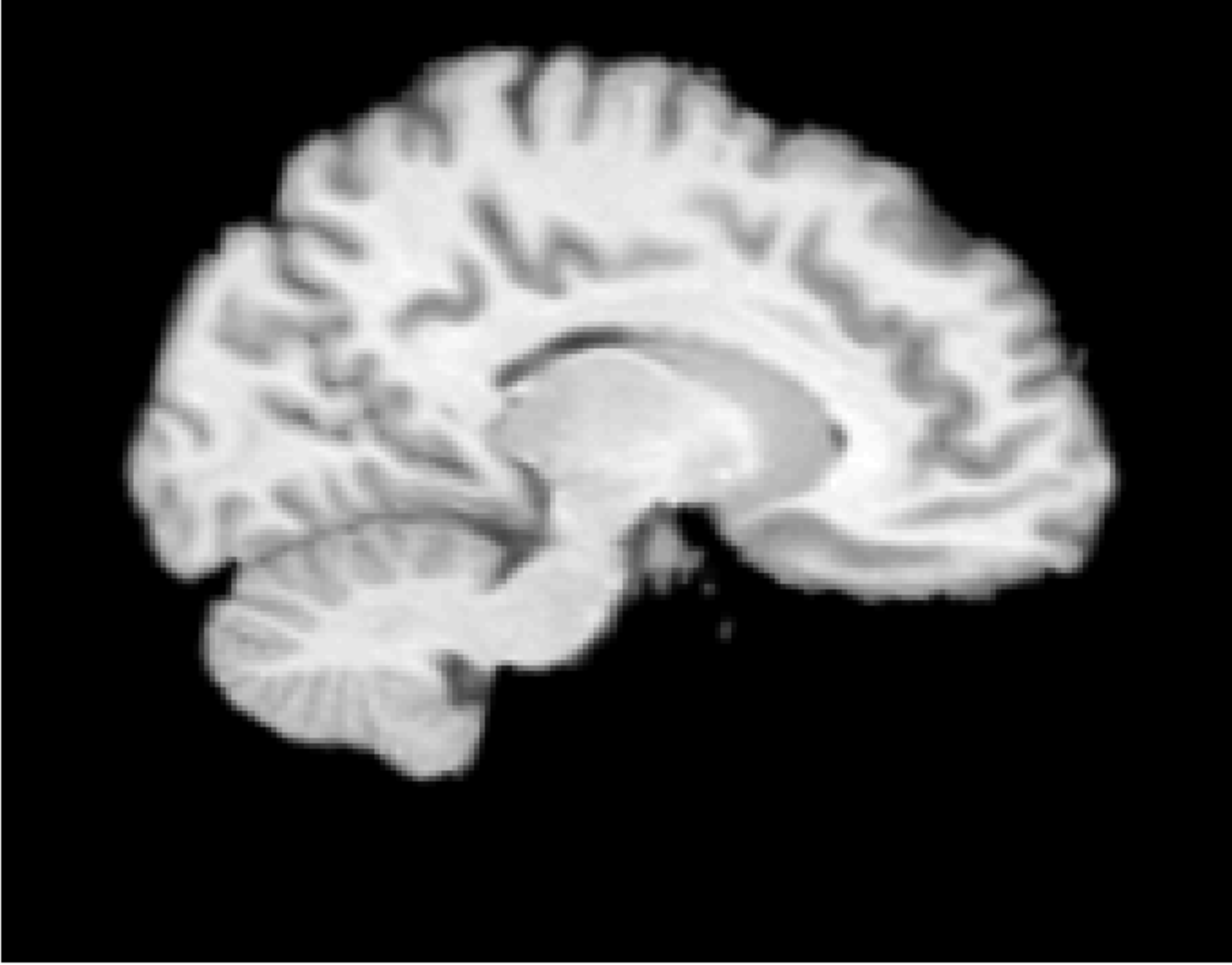} & 
\includegraphics[width=0.13\textwidth]{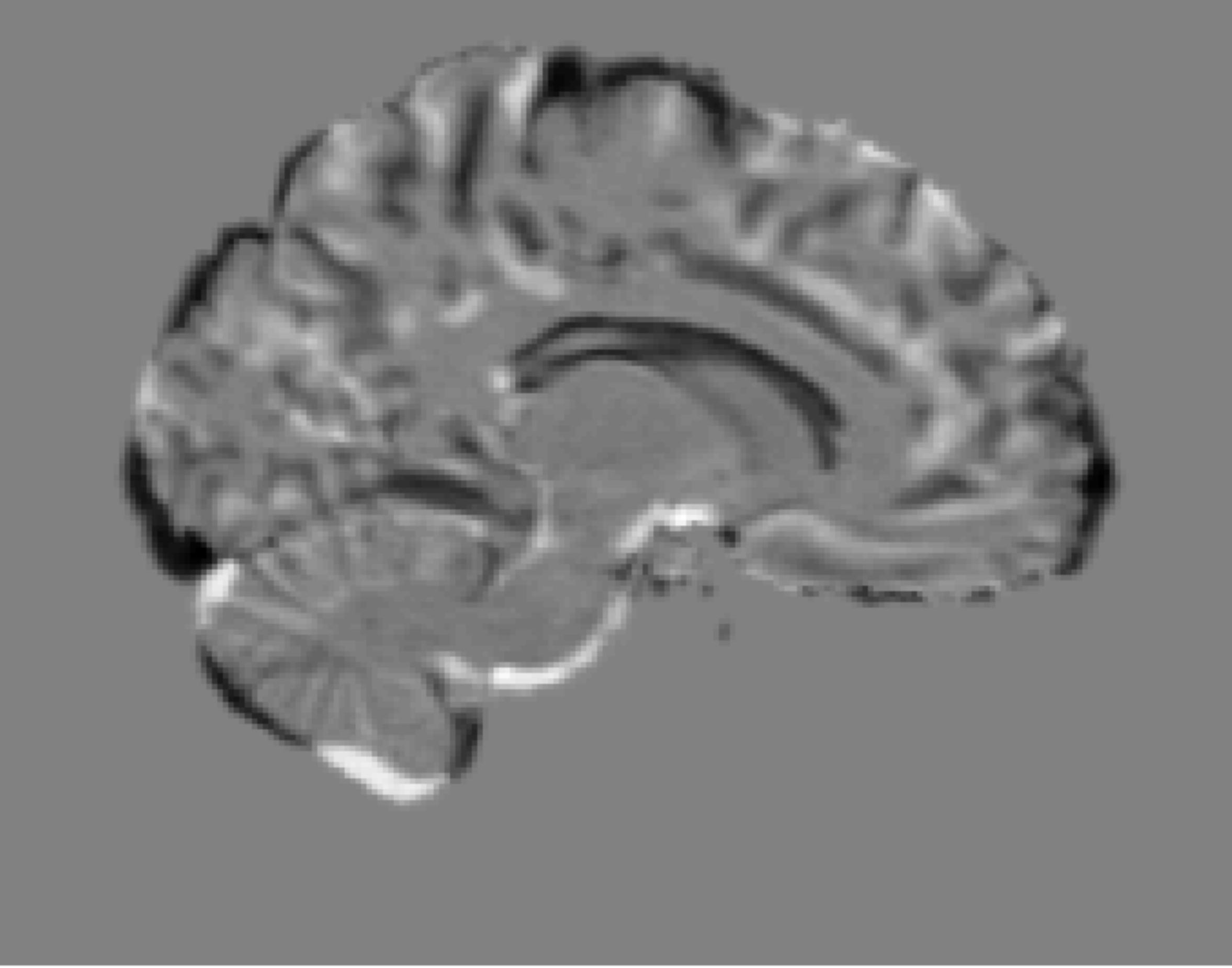} \\
\tiny $I_0$ & \tiny $I_1$ & \tiny $I_0-I_1$ \\
\includegraphics[width=0.13\textwidth]{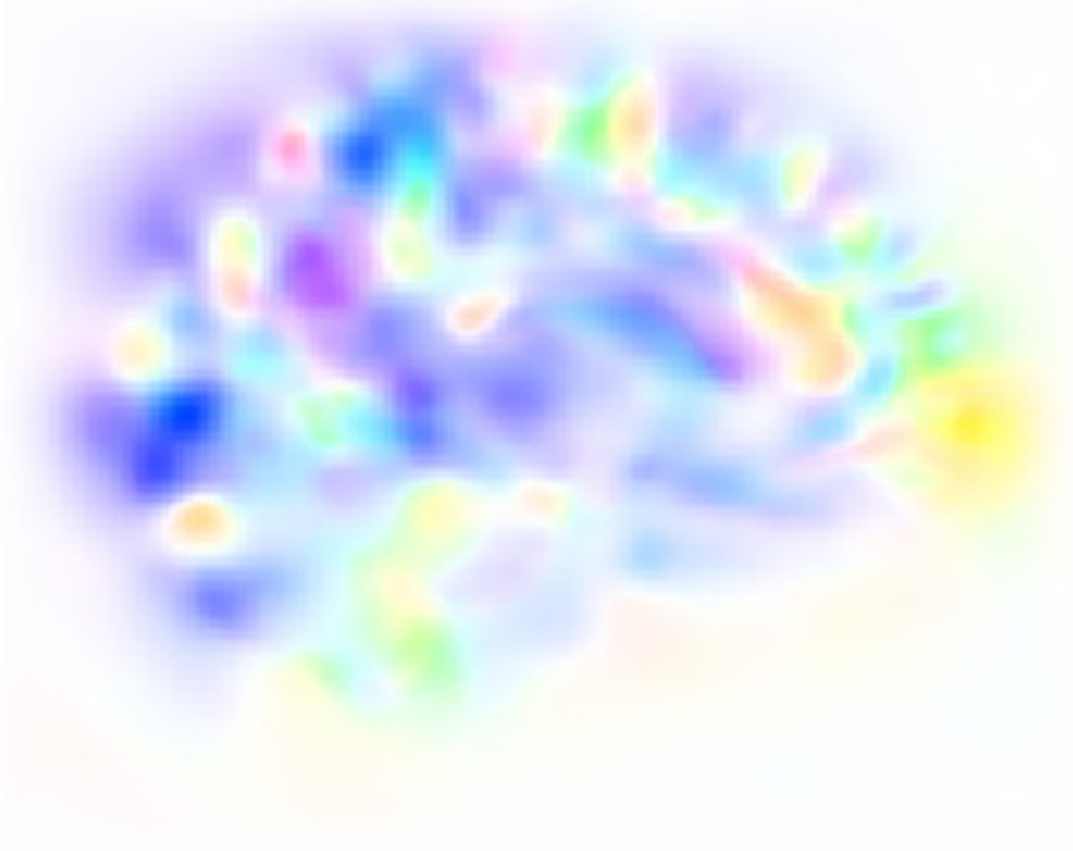} &
\includegraphics[width=0.13\textwidth]{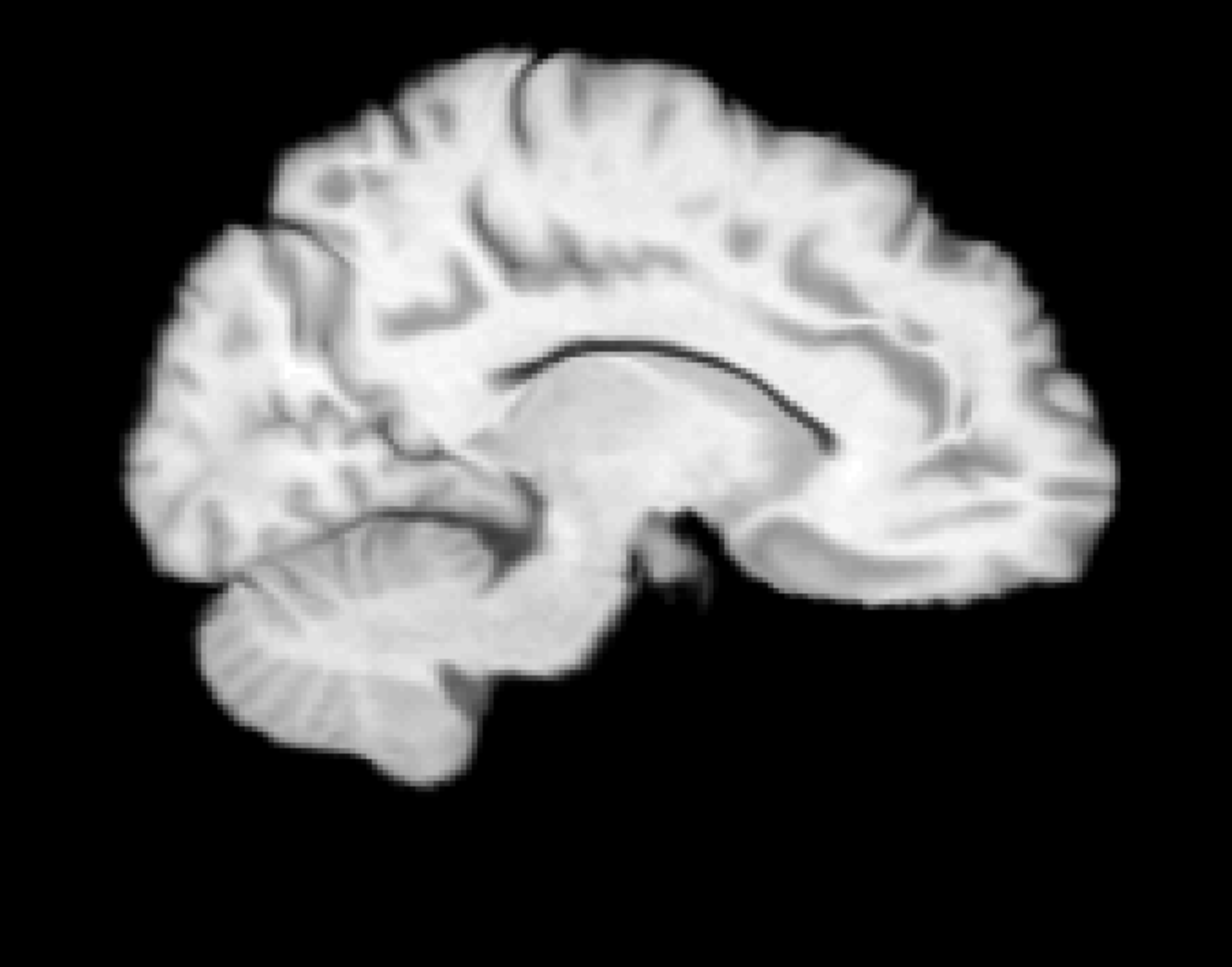} & 
\includegraphics[width=0.13\textwidth]{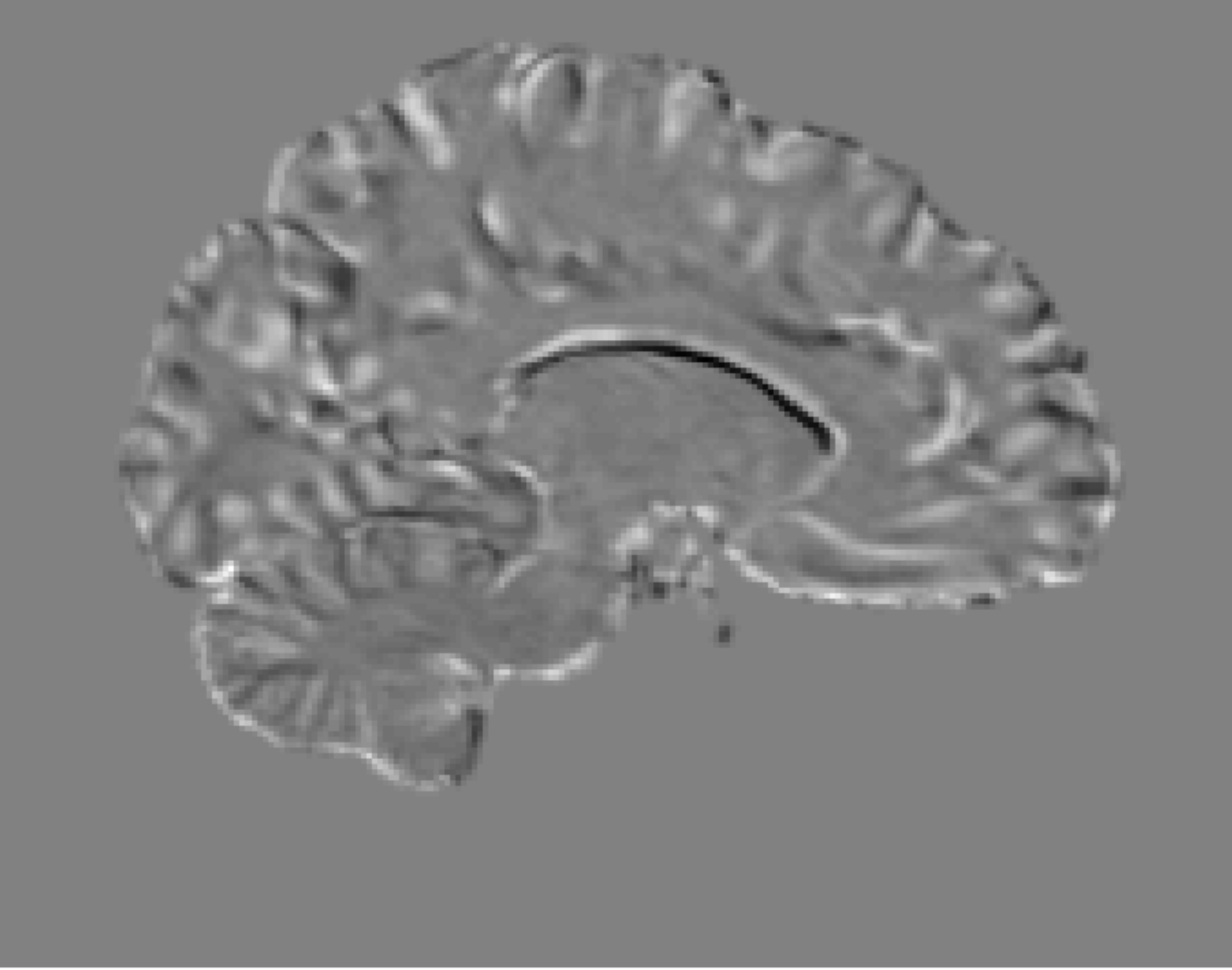} \\
\tiny SVF & \tiny $I_0 \circ (\phi_1^v)^{-1}$ & \tiny $I_0 \circ (\phi_1^v)^{-1} - I_1$
\\
\includegraphics[width=0.13\textwidth]{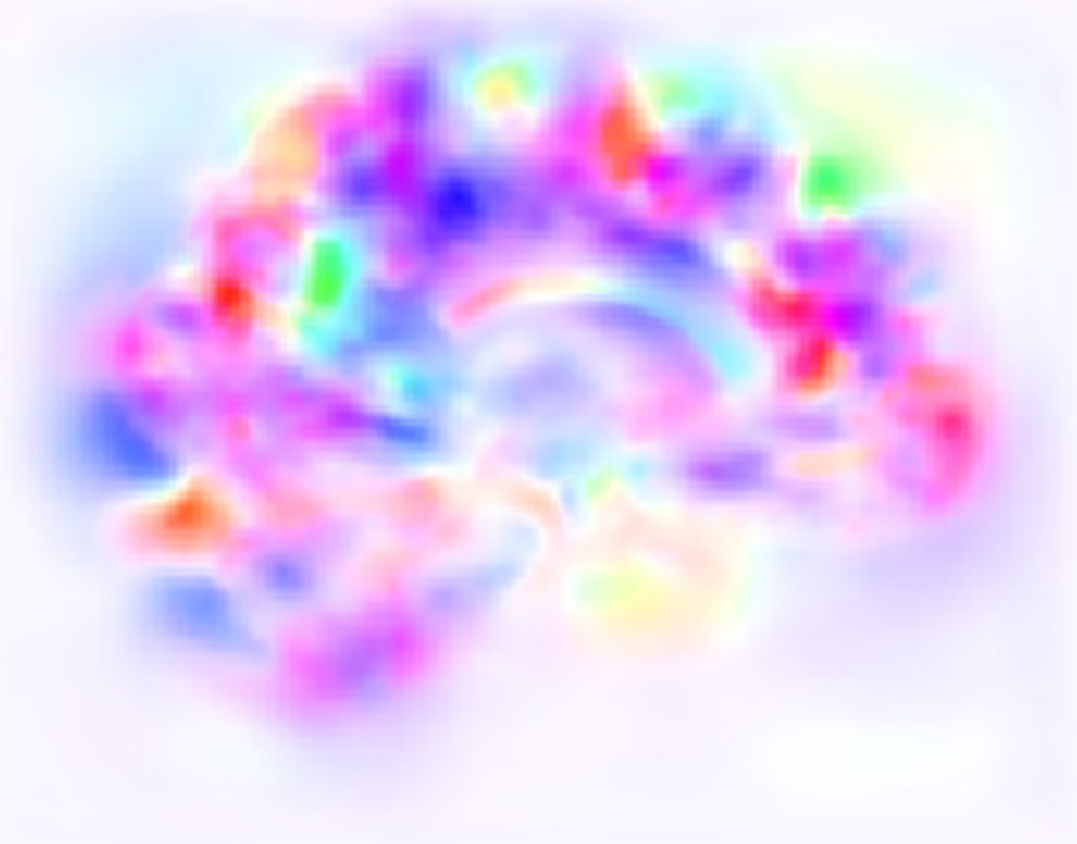} &
\includegraphics[width=0.13\textwidth]{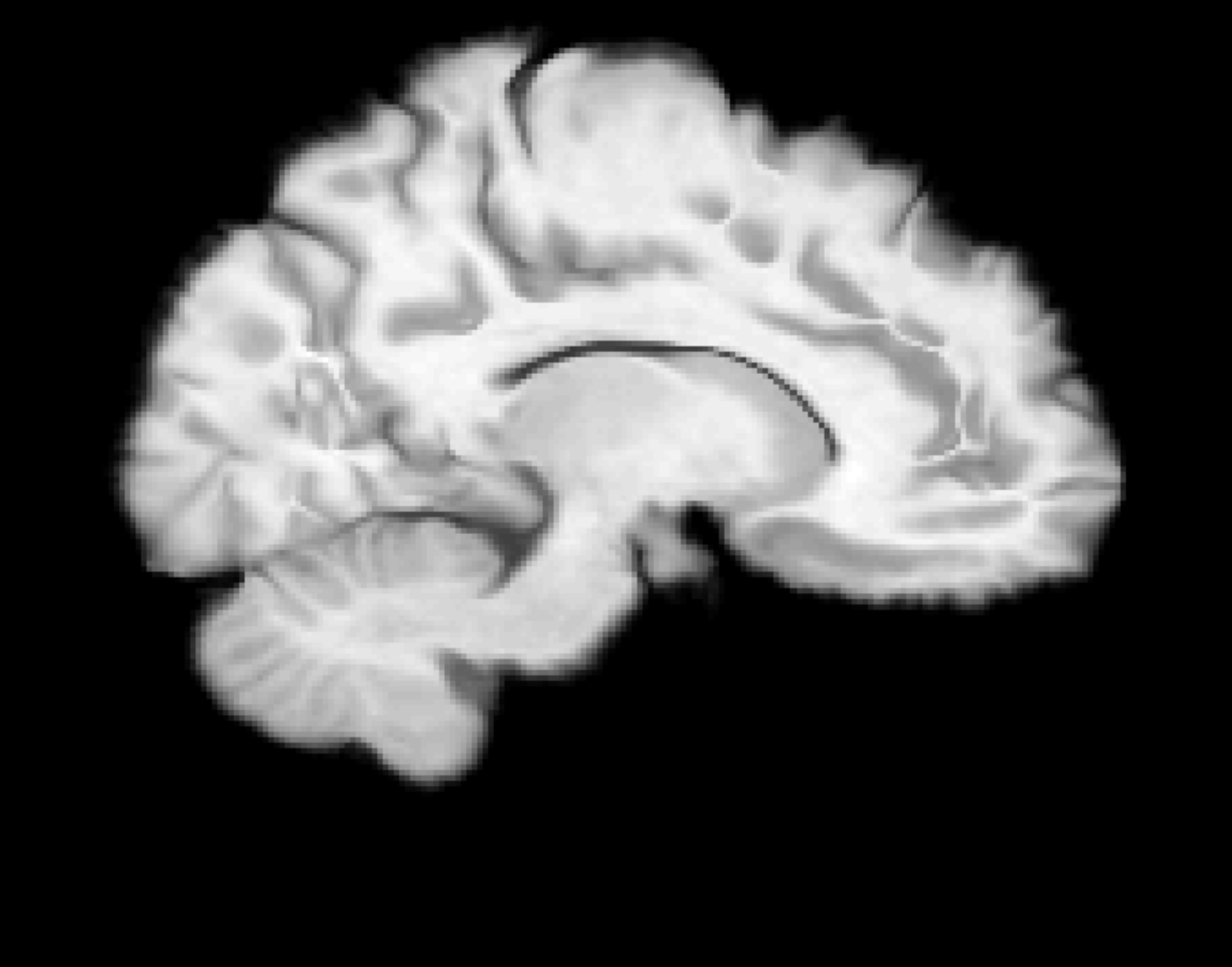} & 
\includegraphics[width=0.13\textwidth]{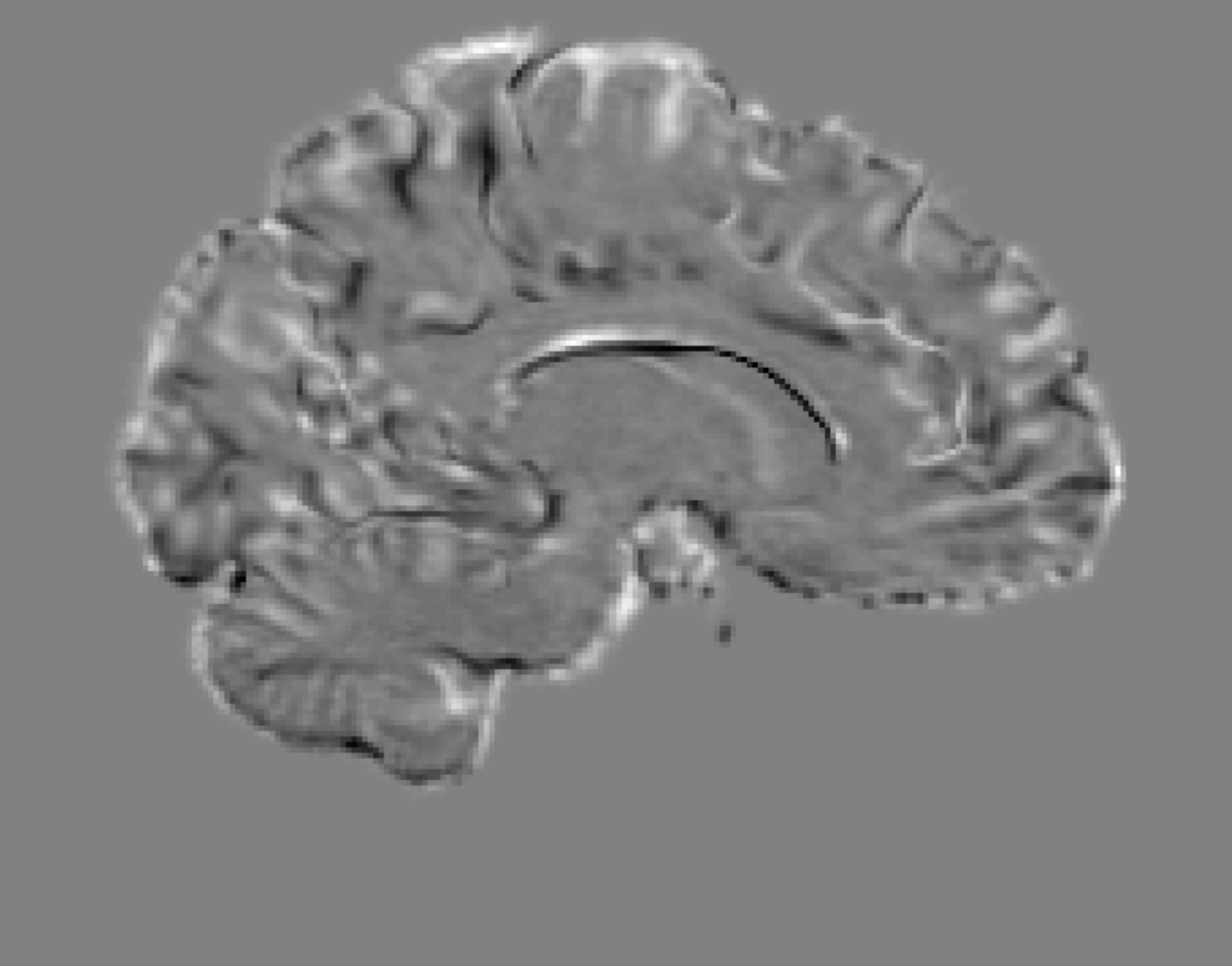} \\ 
\tiny $V_0$ & \tiny $I_0 \circ (\phi_1^v)^{-1}$ & \tiny $I_0 \circ (\phi_1^v)^{-1} - I_1$
\end{tabular} &
\begin{tabular}{ccc}
\includegraphics[width=0.09\textwidth]{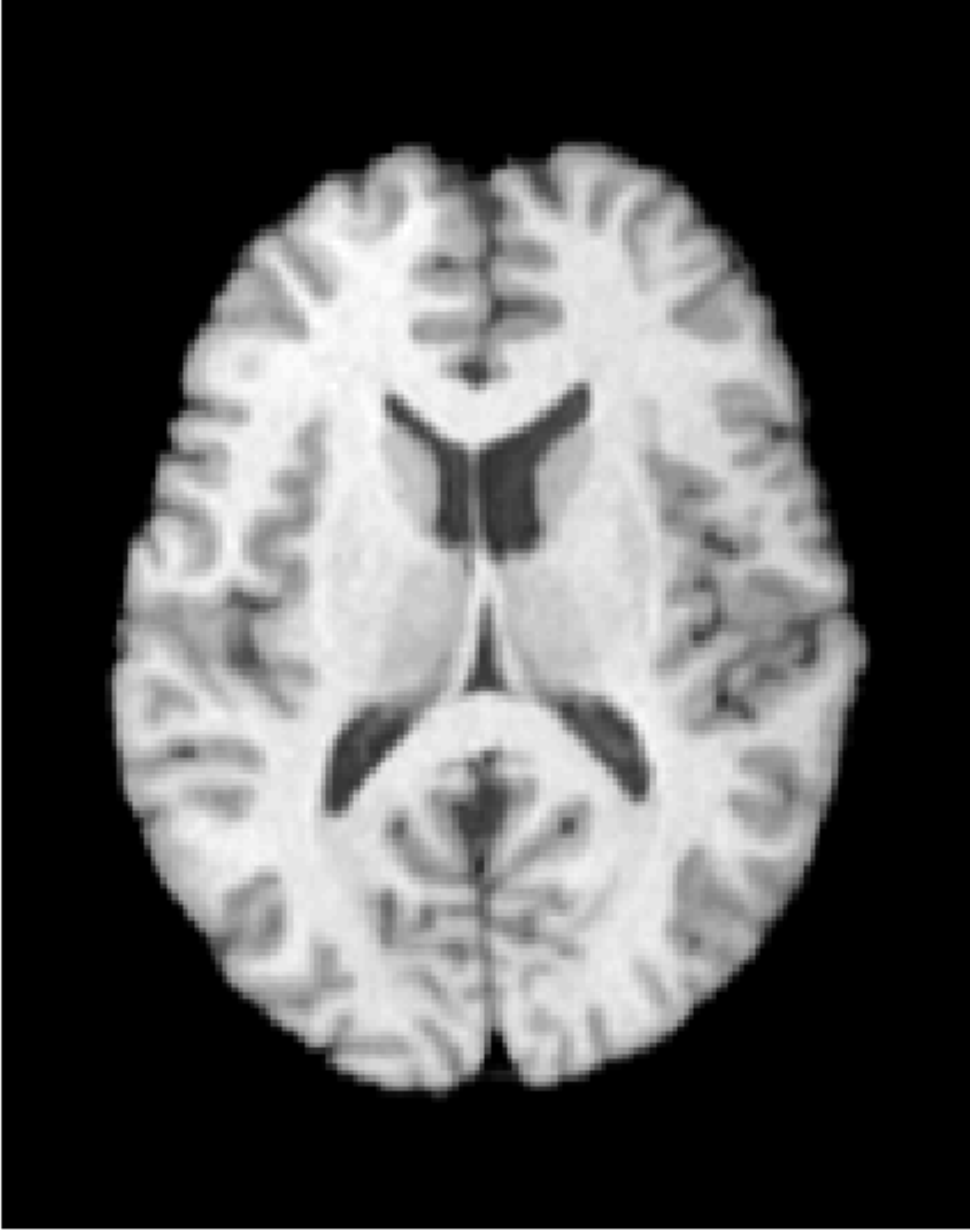} & 
\includegraphics[width=0.09\textwidth]{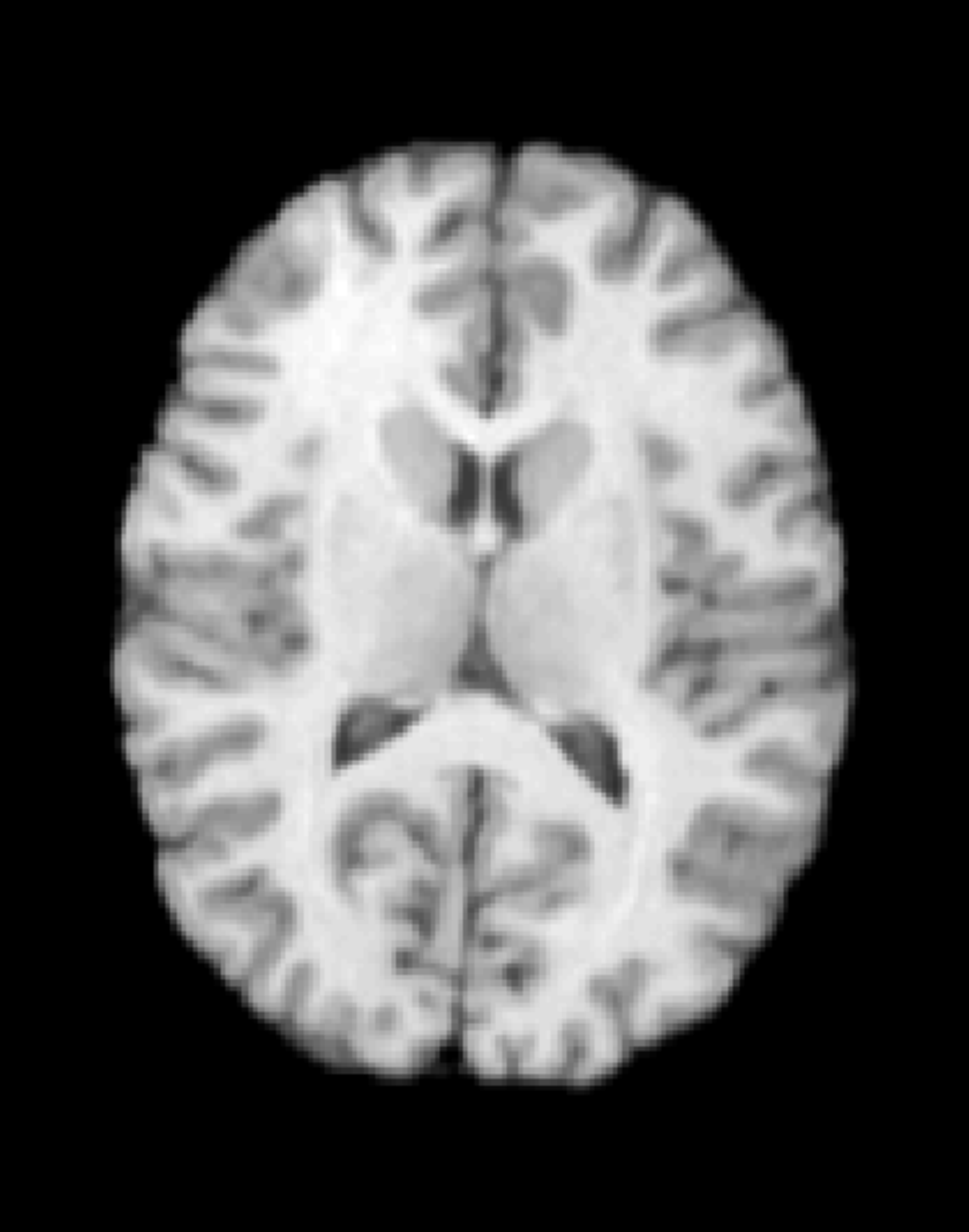} & 
\includegraphics[width=0.09\textwidth]{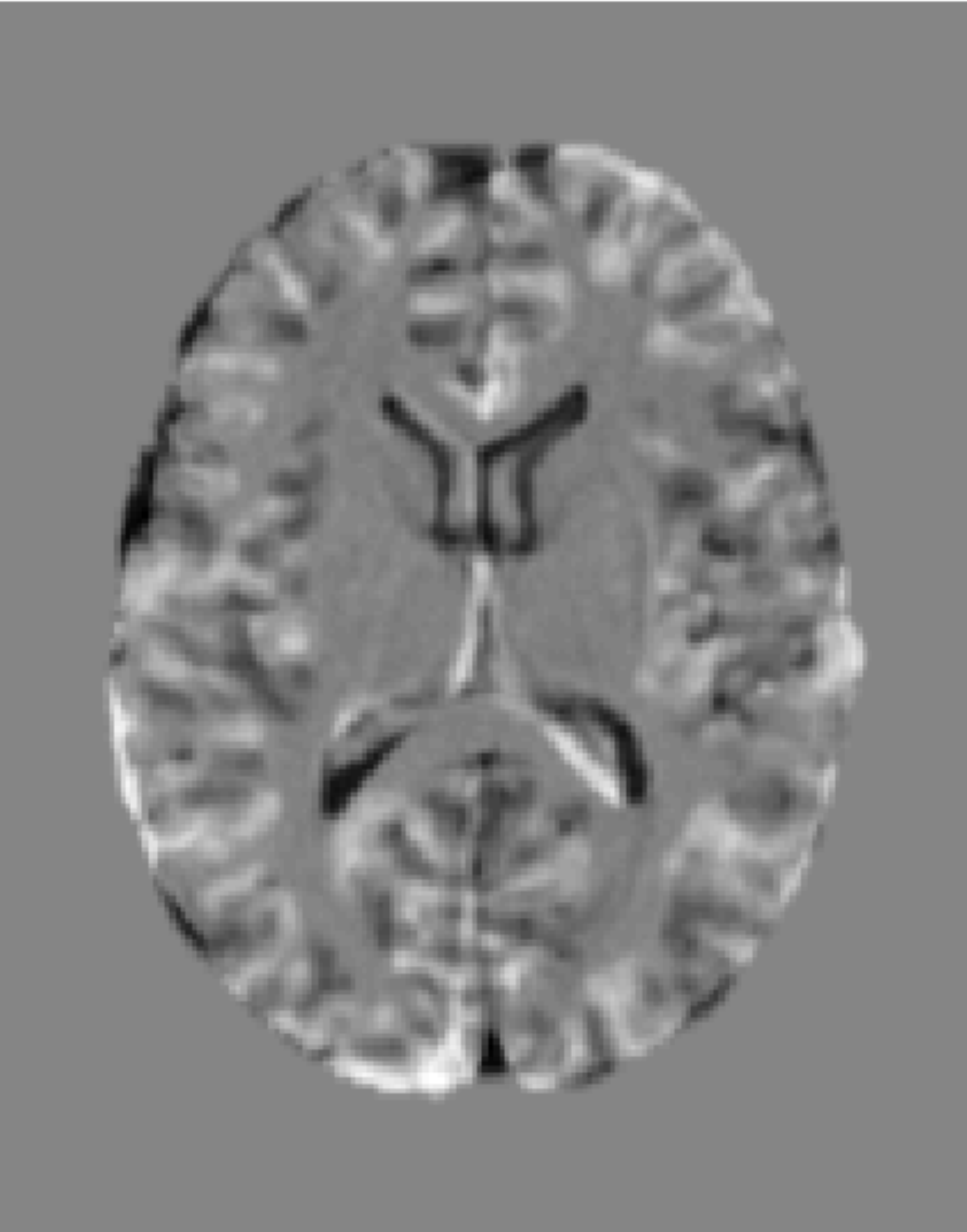} \\
\tiny $I_0$ & \tiny $I_1$ & \tiny $I_0-I_1$ \\
\includegraphics[width=0.09\textwidth]{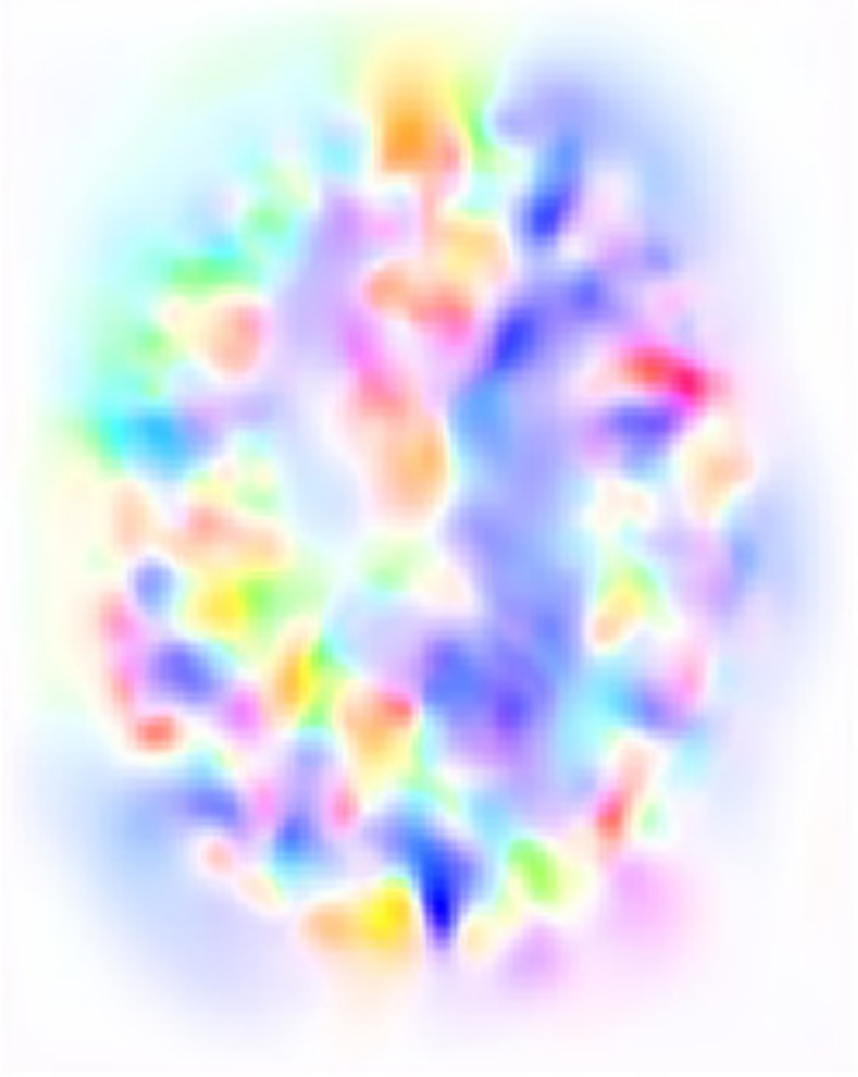} &
\includegraphics[width=0.09\textwidth]{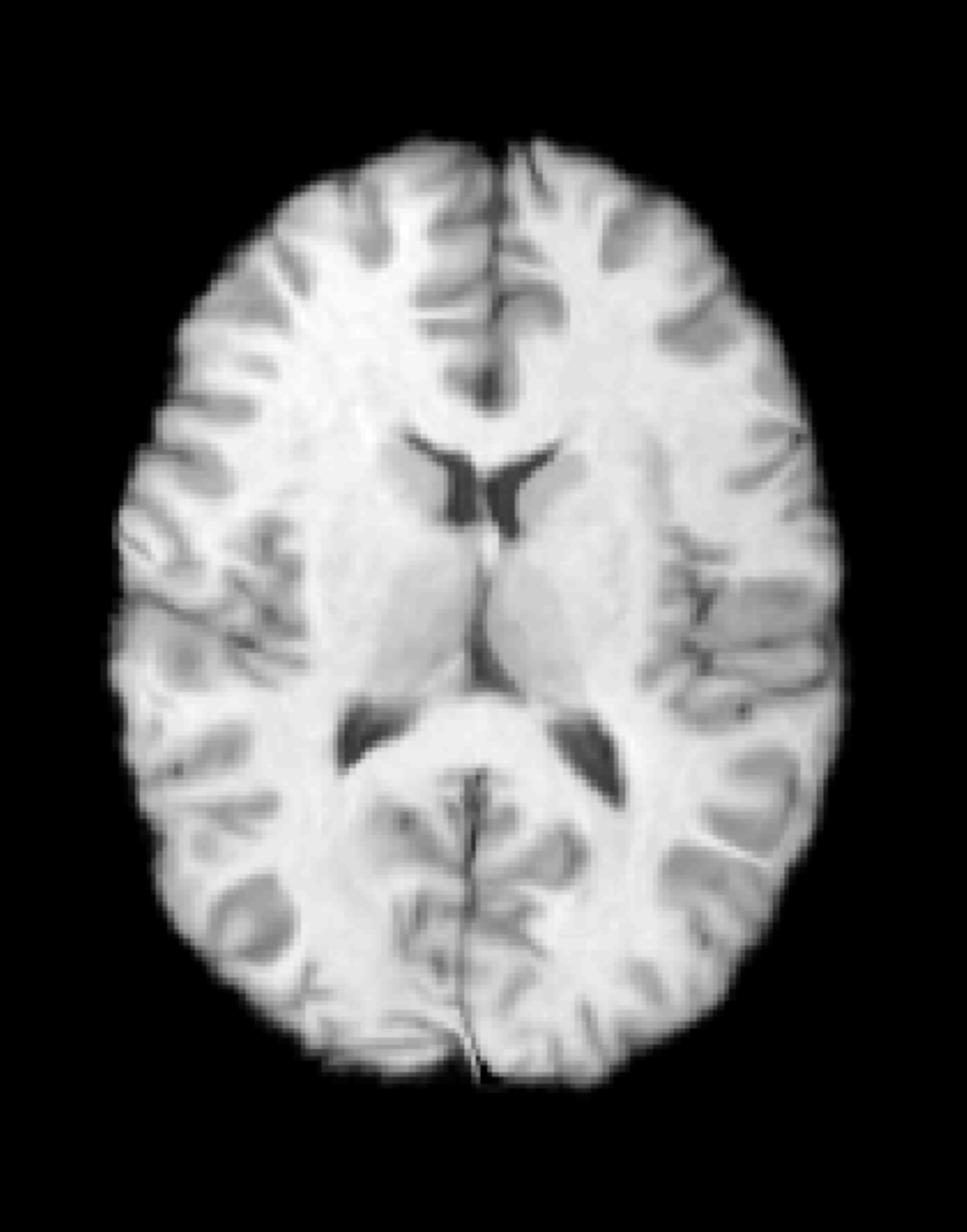} & 
\includegraphics[width=0.09\textwidth]{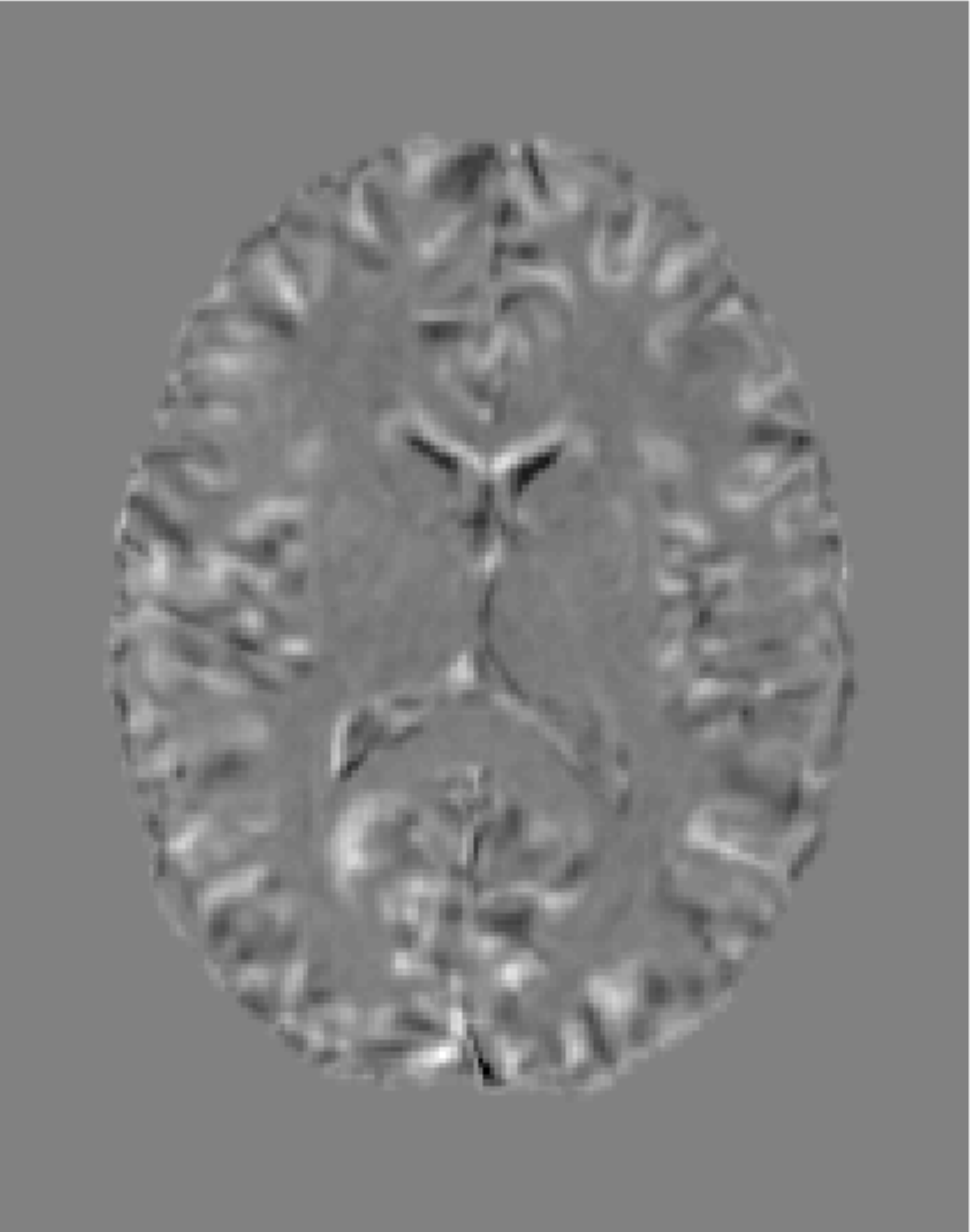} \\
\tiny SVF & \tiny $I_0 \circ (\phi_1^v)^{-1}$ & \tiny $I_0 \circ (\phi_1^v)^{-1} - I_1$
\\
\includegraphics[width=0.09\textwidth]{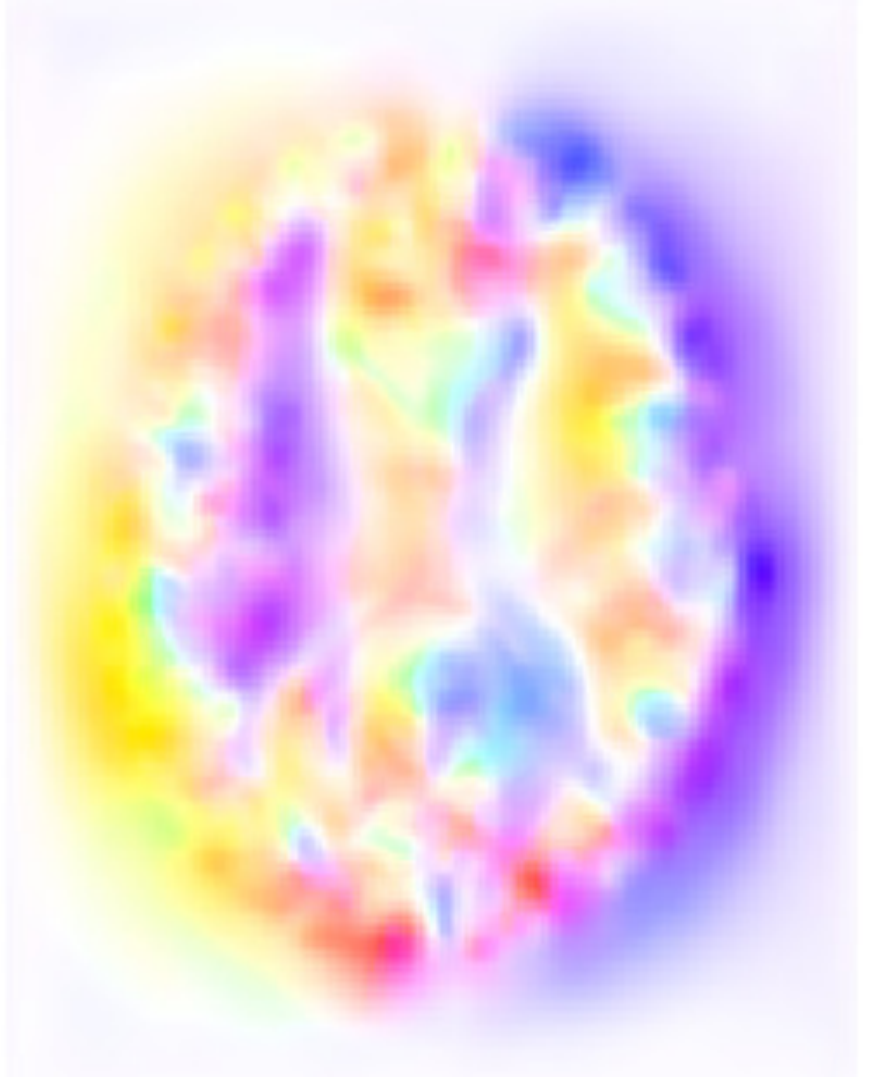} &
\includegraphics[width=0.09\textwidth]{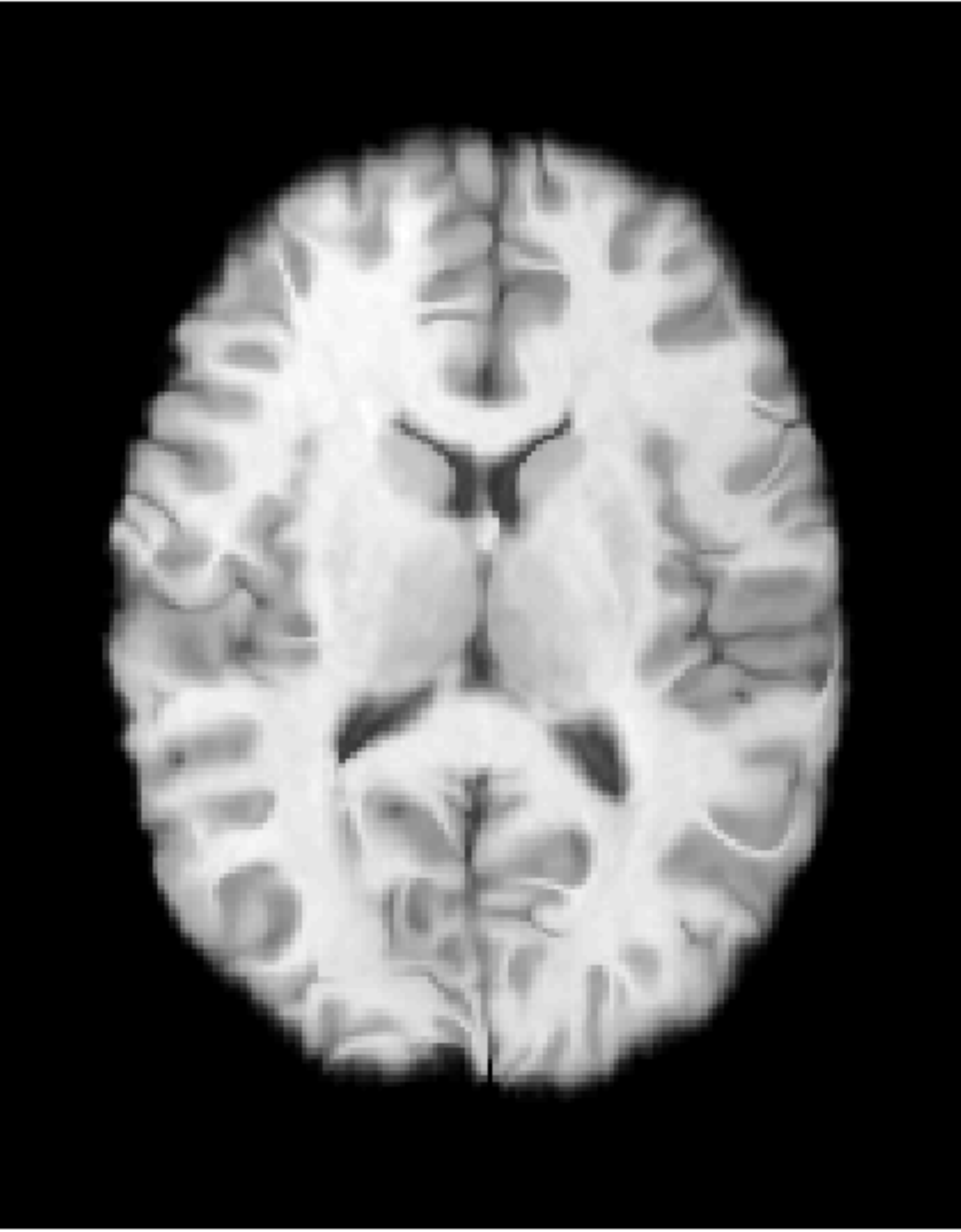} & 
\includegraphics[width=0.09\textwidth]{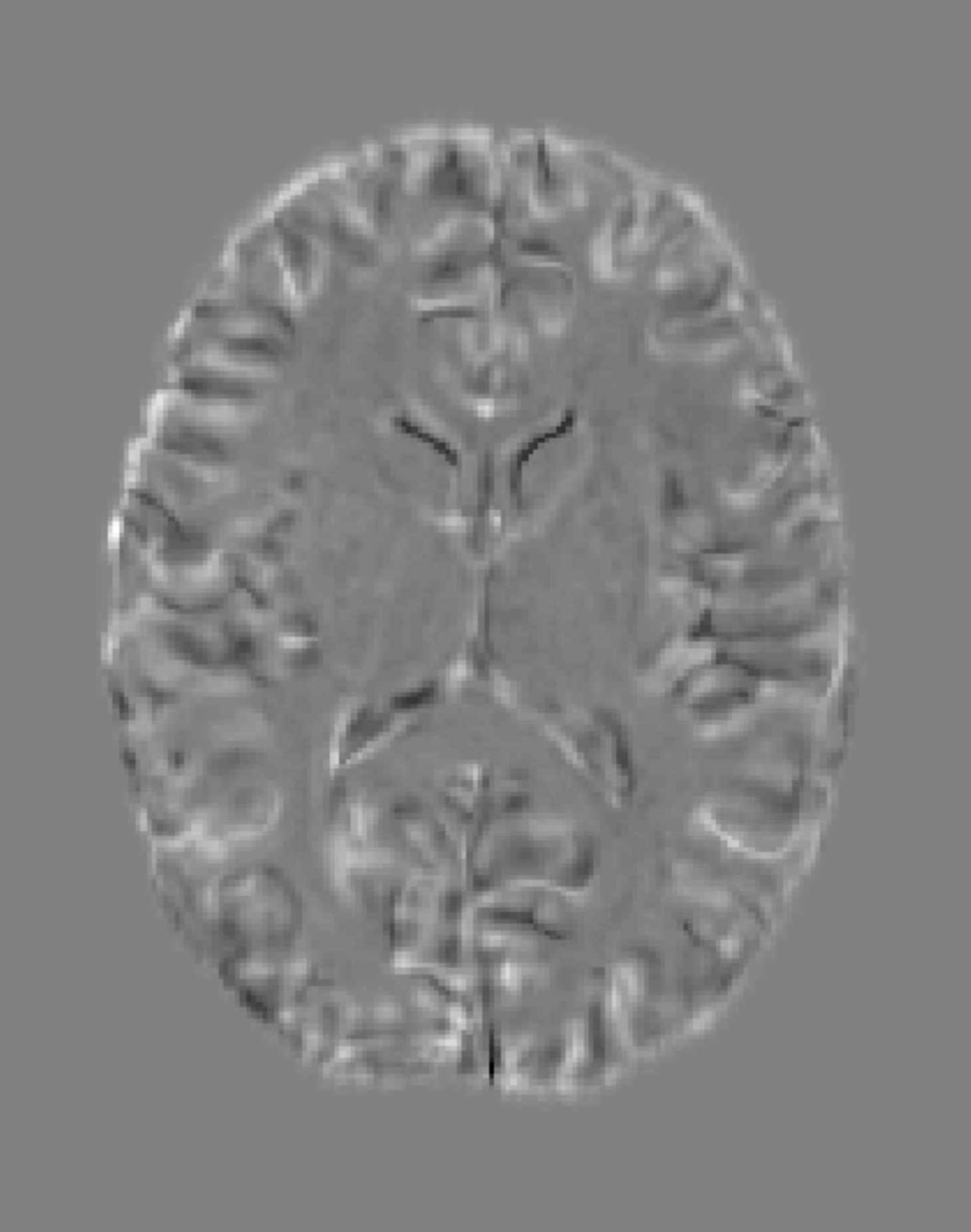} \\ 
\tiny $V_0$ & \tiny $I_0 \circ (\phi_1^v)^{-1}$ & \tiny $I_0 \circ (\phi_1^v)^{-1} - I_1$
\end{tabular}
\end{tabular}
\caption{\small Example of 3D registration results. 
First row, sagittal and axial views of the source and the target images and the differences before registration.
Second row, inferred stationary velocity field, warped image, and differences after registration for SVF-GAN. 
Third row, inferred initial velocity field, warped image, and differences after registration for EPDiff-GAN. 
}
\label{fig:3DQualitative}
\end{figure}

For a qualiative assessment of the quality of the registration results, Figure~\ref{fig:3DQualitative} shows 
the sagittal and axial views of one selected NIREP registration result. 
In the figure, it can be appreciated a high matching between 
the target and the warped ventricles, and more difficult to register regions like the cingulate gyrus 
(observable in the sagittal view) or the insular cortex (observable in the axial view). 
For those not familiar with brain anatomical regions, these regions are easily idenfified as the garnet and orange
regions in https://radiopaedia.org/cases/brain-lobes-annotated-mri-1.

\subsubsection{Computational complexity}

Our GANs models were trained during 2 days, 21 hours, 59 minutes and 48 seconds.
The VRAM memory load was equal to the whole GPU capacity.
The inference of a stationary or an initial velocity field took 1.3 seconds.

\section{Conclusions}
\label{sec:Conclusions}

We have proposed an adversarial learning LDDMM method for the registration of 3D mono-modal images.
Our method is inspired by the recent literature on deformable image registration with adversarial learning
and combines the best performing generative, discriminative, and adversarial ingredients from these works within 
the LDDMM paradigm.
We have successfully implemented two models: one for the stationary parameterization and the other for the 
EPDiff-constrained non-stationary parameterization (geodesic shooting). 

Our experiments have shown that the inferred velocity fields are comparable to the solutions of model-based approaches.
In addition, the evaluation study has shown the competitiveness of our approach with state of the art model- and data-
based methods. It should be remarked that our methods perform similarly to Quicksilver, a supervised method that uses
patches for training, and therefore, it learns in a rich-data environment.
In contrast, our method is unsupervised and uses the whole image for training in a data-hungry environment.
Despite the apparent disadvantages, the evaluation only reported the weakness of EPDiff-GAN in the registration 
of two regions that are located in challenging locations.
Indeed, our proposed methods outperform Voxelmorph II, a regular (not GAN) unsupervised method for diffeomorphic registration
usually selected as benchmark in the state of the art.

Once the training has been completed, our method shows a computational time of over a second for the inference 
of velocity fields. 
Therefore, our proposal may constitute a good candidate for the massive computation of diffeomorphisms in 
Computational Anatomy studies.

\section*{Acknowledgements}

*Data used in preparation of this article were obtained from the Alzheimer’s Disease Neuroimaging Initiative
(ADNI) database (adni.loni.usc.edu). As such, the investigators within the ADNI contributed to the design
and implementation of ADNI and/or provided data but did not participate in analysis or writing of this report.
A complete listing of ADNI investigators can be found at:
\url{http://adni.loni.usc.edu/wp-content/uploads/how_to_apply/ADNI_Acknowledgement_List.pdf}

Data collection and sharing for this project was funded by the Alzheimer's Disease Neuroimaging Initiative
(ADNI) (National Institutes of Health Grant U01 AG024904) and DOD ADNI (Department of Defense award
number W81XWH-12-2-0012). ADNI is funded by the National Institute on Aging, the National Institute of
Biomedical Imaging and Bioengineering, and through generous contributions from the following: AbbVie,
Alzheimer’s Association; Alzheimer’s Drug Discovery Foundation; Araclon Biotech; BioClinica, Inc.; Biogen;
Bristol-Myers Squibb Company; CereSpir, Inc.; Cogstate; Eisai Inc.; Elan Pharmaceuticals, Inc.; Eli Lilly and
Company; EuroImmun; F. Hoffmann-La Roche Ltd and its affiliated company Genentech, Inc.; Fujirebio; GE
Healthcare; IXICO Ltd.; Janssen Alzheimer Immunotherapy Research \& Development, LLC.; Johnson \&
Johnson Pharmaceutical Research \& Development LLC.; Lumosity; Lundbeck; Merck \& Co., Inc.; Meso
Scale Diagnostics, LLC.; NeuroRx Research; Neurotrack Technologies; Novartis Pharmaceuticals
Corporation; Pfizer Inc.; Piramal Imaging; Servier; Takeda Pharmaceutical Company; and Transition
Therapeutics. The Canadian Institutes of Health Research is providing funds to support ADNI clinical sites
in Canada. Private sector contributions are facilitated by the Foundation for the National Institutes of Health
(www.fnih.org). The grantee organization is the Northern California Institute for Research and Education,
and the study is coordinated by the Alzheimer’s Therapeutic Research Institute at the University of Southern
California. ADNI data are disseminated by the Laboratory for Neuro Imaging at the University of Southern
California.

\bibliographystyle{splncs}
\bibliography{abbsmallTMI.bib,ProyectoInvestigador.bib,Diffeo.bib,OpticalFlow.bib,Miccai2015.bib,CNNs.bib}

\end{document}